\documentclass[lettersize,journal]{IEEEtran}
\IEEEoverridecommandlockouts

\usepackage{booktabs}
\usepackage{graphicx}

\usepackage{amsfonts,amssymb}
\usepackage{xcolor}
\usepackage[ruled,vlined]{algorithm2e}
\usepackage{subfigure}
\usepackage{amsmath}
\usepackage{bm}
\usepackage{rotating, multirow}
\usepackage{epstopdf}
\usepackage{float}
\usepackage{enumitem}
\usepackage{makecell}
\usepackage{titlesec}
\usepackage{tcolorbox}
\usepackage{colortbl}
\definecolor{softteal}{HTML}{A2C9AE}
\usepackage[colorlinks=true]{hyperref}

\def\BibTeX{{\rm B\kern-.05em{\sc i\kern-.025em b}\kern-.08em
    T\kern-.1667em\lower.7ex\hbox{E}\kern-.125emX}}
\usepackage{balance}

\begin{document}
\begin{sloppypar}

\title{A Step Toward World Models: A Survey on Robotic Manipulation}
\author{Peng-Fei Zhang, Ying Cheng, Xiaofan Sun, Shijie Wang, Fengling Li, Lei Zhu, Heng Tao Shen
\thanks{School of Computer Science and Technology, Tongji University.}
}

\maketitle

\begin{abstract}
Autonomous agents are increasingly expected to operate in complex, dynamic, and uncertain environments, performing tasks such as manipulation, navigation, and decision-making. Achieving these capabilities requires agents to understand the underlying mechanisms and dynamics of the world, moving beyond reactive control or simple replication of observed states. This motivates the development of world models as internal representations that encode environmental states, capture dynamics, and support prediction, planning, and reasoning. Despite growing interest, the definition, scope, architectures, and essential capabilities of world models remain ambiguous. In this survey, we go beyond prescribing a fixed definition and limiting our scope to methods explicitly labeled as world models. Instead, we examine approaches that exhibit the core capabilities of world models through a review of methods in robotic manipulation. We analyze their roles across perception, prediction, and control, identify key challenges and solutions, and distill the core components, capabilities, and functions that a fully realized world model should possess. Building on this analysis, we aim to motivate further development toward generalizable and practical world models for robotics. \textit{This is an initial version of the survey. The content will be expanded and refined in future updates.}

\end{abstract}


\IEEEpeerreviewmaketitle

\begingroup
\hypersetup{linkcolor=blue} 
\tableofcontents
\endgroup

\section{Introduction}
\begin{quote}
``\textit{If I have seen further, it is by standing on the shoulders of giants}.''
\hfill --- Isaac Newton
\end{quote}

%------------------------------------------------------------------------
\begin{figure*}[tbh]
\center
\includegraphics[angle=0, width=1\textwidth]{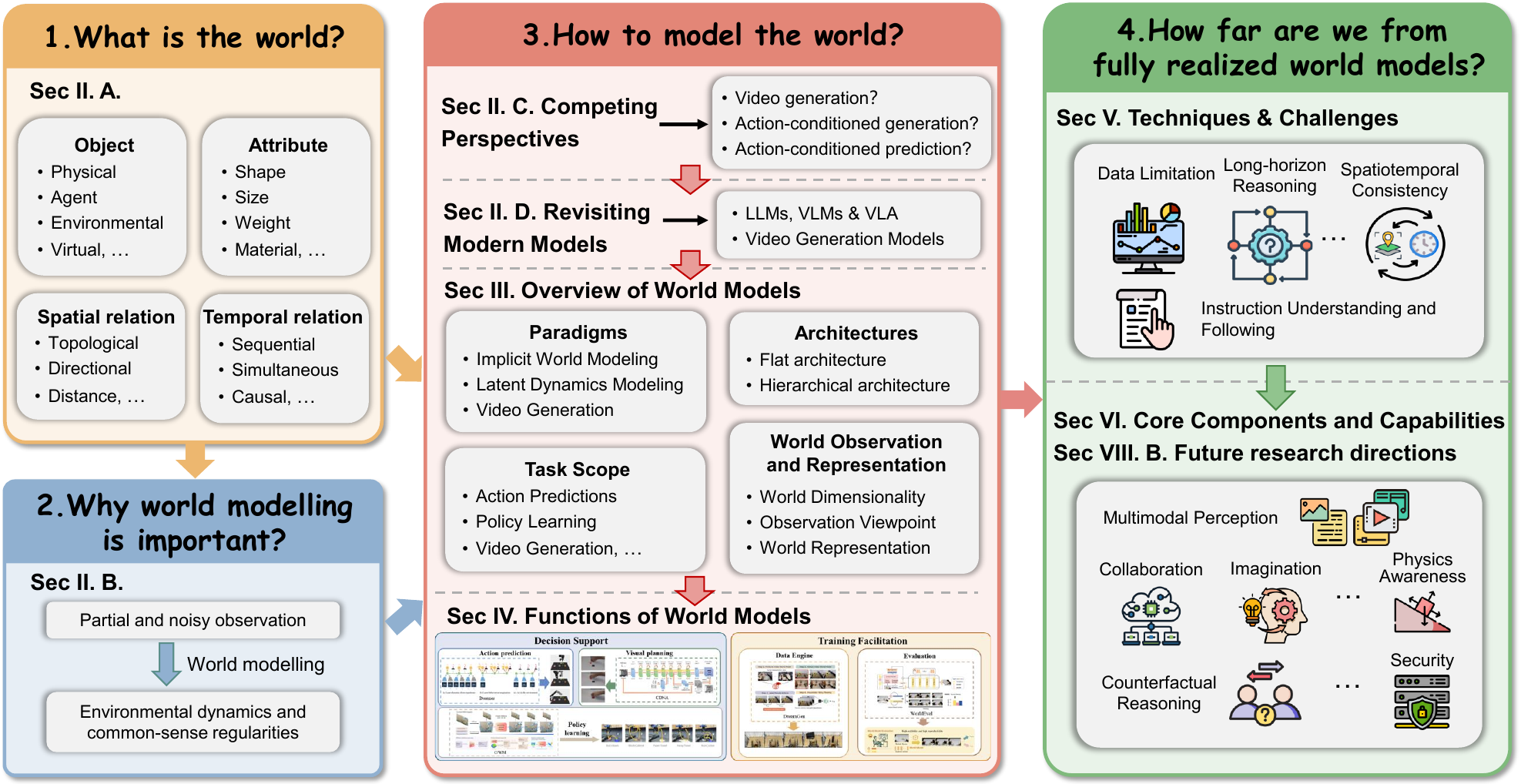}
\caption{Conceptual flow of the survey. The survey aims to clarify the motivation behind world modeling, explore its essential scope and development, and illuminate pathways toward more general and capable world models.}
\label{thinking}
\end{figure*}
%------------------------------------------------------------------------

Autonomous agents are designed to extend human capabilities, assisting in tasks that are dangerous, repetitive, or demand high precision, thereby enhancing productivity across diverse applications. Robots serve as their physically embodied form, aiming to realize seamless interaction with the physical world. Achieving such capabilities requires agents to move beyond reactive control and the mere imitation of observed behaviors, instead developing the ability to model, reason about, and predict environmental dynamics. In this context, world models have emerged as powerful internal representations that enable robots to anticipate future outcomes, plan effectively, and ultimately act intelligently in the real world. Richens \textit{et al.} \cite{richens2025general} argue that any agent capable of generalizing to solve multi-step tasks must implicitly learn a predictive model of its environment, e.g., a world model. 

The concept of ``world models'' in computer science dates back to the 1960s \cite{minsky1975framework}, and numerous methods have since been proposed toward more capable models \cite{schmidhuber2015learning,ha2018world,du2023video,wu2024ivideogpt,zhang2025combo,huang2025enerverse,wang2025language,yang2025roboenvision}, though not all explicitly identify as world models. For example, Wang \textit{et al.} \cite{wang2025language,yang2025roboenvision,du2023video,wu2024ivideogpt,zhang2025combo,huang2025enerverse} leverage video generation models as a form of world models, which encode extensive world knowledge from large-scale training data and can predict future states based current observations and/or actions. LeCun \textit{et al.} \cite{lecun2022path,hafnerdream,hafner2021mastering,wu2023daydreamer,hafner2023mastering} emphasizes modeling abstract world state representations, while Zitkovich \textit{et al.} \cite{zitkovich2023rt,huang2024embodied,hong2024multiply} utilize vision-language-action models (VLA) models that do not explicitly generate future states. The scope of existing methods varies from 2D scene prediction to 4D world modeling \cite{yang2023learning,bruce2024genie,guo2025flowdreamer,bu2024closed,team2025aether,zhen2025tesseract}, reflecting different understandings of what it means to model the world. The observation viewpoint of the world includes both third-person (exocentric) \cite{guo2025flowdreamer,ferraro2025focus,villar2025playslot} and first-person (egocentric) \cite{chen2025egoagent,grauman2024ego} perspectives. 

World models play a critical role in robotic learning in two ways. They allow robots to improve autonomous policies by the internal prediction and simulation of environmental dynamics, future states, and/or action outcomes \cite{hafner2019learning,hafnerdream,hafner2021mastering,wu2023daydreamer,ebert2018robustness,guo2025flowdreamer}. They also support scalable policy training and evaluation by generating realistic rollouts and physical interactions \cite{zitkovich2023rt,black2024pi_0,team2025gemini,barcellona2025dream,lu2025gwm,liu2025rdt}. From a functional standpoint, current approaches range from single-purpose models, such as those for visual planning \cite{ebert2018robustness,finn2016unsupervised}, future-scene generation \cite{sudhakar2024controlling,barcellona2025dream,jang2025dreamgen}, or action prediction \cite{sekar2020planning}, to more integrated systems that couple multiple abilities within a unified framework \cite{cen2025worldvla,chen2025egoagent,zhen20243d,song2025physical,zhu2025irasim}. 

\textit{These variations indicate that the notion of a world model remains unsettled, with its conceptual, architectural, and functional boundaries not yet clearly defined.}

Addressing these questions requires standing on the shoulders of prior contributions, carefully analyzing existing methodologies to gain inspiration for elucidating the boundaries of world models. In this survey, rather than hastily defining what constitutes a world model, we provide a comprehensive review of the literature, highlighting their core principles, architectures, and functional roles in enabling intelligent robotic systems. We extend the scope beyond works explicitly labeled as world models, examining their core principles and outlining pathways for constructing practical models that can drive the development of general and adaptive robotic agents. 
 
This survey is organized around a set of guiding questions designed to provoke thought and provide inspiration. Readers can explore the survey with these questions in mind, using them to provoke thought, gain inspiration, and reflect on the challenges and opportunities in developing world models for robotic manipulation.
\begin{itemize}
    \item \textit{What is the world model and its conceptual, architectural, and functional boundaries?}
    \item \textit{How should the world be sensed and presented?}
    \item \textit{What level of model fidelity and coverage is required to reliably support robotic tasks?}
    \item \textit{Is it necessary to learn a world model, given the complexity, resource demands, and potential challenges involved?}
    \item \textit{How far are current world models from fully realized world models?}
    \item \textit{Is human cognition \cite{bjorck2025gr00t,wang2025dmwm} the ultimate goal for world models?}
\end{itemize}

The main contributions of this survey are as follows:
\begin{itemize}
\item \textbf{Comprehensive taxonomy of world model architectures.} We provide a systematic categorization of existing designs, including latent space modeling methods, video generation-based models, direct projection based (implicit modeling) methods and other emerging structures.
\item \textbf{Functional analysis.} We discuss diverse roles of world models in robotic manipulation, including action prediction and planning, robotic learning, and evaluation, highlighting their contribution to intelligent manipulation.
\item \textbf{Capability framework.} We analyze the essential abilities that a world model should possess, such as perception, prediction, imagination, and interaction, aiming to clarify what constitutes a generalizable and capable world model.
\item \textbf{Challenges and future directions.} We summarize key challenges, including long-horizon reasoning, physical awareness, and generalization, and discuss potential research directions and solutions toward building practical, real-world models.
\end{itemize}

\noindent\textbf{Related Surveys.} Our survey differs substantially from existing reviews. Several prior surveys have examined world models in robotics, but most focus on specific aspects and provide limited conceptual analysis. For example, Yu \textit{et al.} \cite{yu2025survey} emphasize video generation, Kong \textit{et al.} \cite{kong20253d} cover 3D/4D world modeling, Ai \textit{et al.} \cite{ai2025review} study dynamics learning, and Lin \textit{et al.} \cite{lin2025exploring} address physics cognition. Long \textit{et al.} \cite{long2025survey} review architectures and functional roles of world models, whereas Zhu \textit{et al.} \cite{zhu2024sora,liang2025large,ding2025understanding} primarily compile representative works. While these surveys provide valuable overviews, they offer limited discussion of the essential characteristics and functional requirements of comprehensive world models. In contrast, our survey presents a holistic, problem-centered perspective, highlighting key challenges, solution strategies, and future directions for world modeling in robotics.

\noindent\textbf{Paper Organization.} The remainder of this paper is organized as follows. Section \ref{Prelim} introduces the conceptual foundations of world models. Section \ref{overview} provides an overview of current world models, including their learning paradigms, structural designs, representations of the world, and task scopes. Sections \ref{fwm} and \ref{ktnc} describe the key functions of existing world models and summarize the principal techniques and challenges, respectively. Section \ref{datasets} reviews the major training resources used in world-model research. Section \ref{def_wm} then summarizes the fundamental components and capabilities of world models based on this review, followed by Section \ref{cfrd}, which presents conclusions and outlines future research directions. Although this may occasionally lead to some repetition, certain key ideas are revisited throughout the paper to aid understanding and reinforce their conceptual connections.

\begin{table*}[t]
\centering
\caption{A Summary of Representative World Models.}
% \vspace{-0.3cm}
\label{tab_rwm}
% \small
\begin{itemize}[leftmargin=*,noitemsep,topsep=0pt]
    \item \textbf{Prediction Tasks:} \underline{\textit{AP}}: Action Prediction, \underline{\textit{PL}}: Policy Learning, \underline{\textit{VP}}: Visual Planning, \underline{\textit{Static}}: Static Visual Prediction, \underline{\textit{Action-cond.}}: Action-conditioned Visual Prediction.
    \item \textbf{Input \& Output:} \underline{\textit{L}}: Language, \underline{\textit{V}}: Video, \underline{\textit{A}}: Action, \underline{\textit{S}}: State, \underline{\textit{I}}: Image, \underline{\textit{P}}: Point Cloud, \underline{\textit{Tr}}: Trajectory.
    \item \textbf{Core Components:} \underline{\textit{CLIP}}: Contrastive Language-Image Pre-training, \underline{\textit{DiT}}: Diffusion Transformer, \underline{\textit{IDM}}: Inverse Dynamics Model, \underline{\textit{GPT}}: Generative Pre-trained Transformer, \ 
    \underline{\textit{LLM}}: Large Language Model, \underline{\textit{LSTM}}: Long Short-Term Memory,  \underline{\textit{RSSM}}: Recurrent State-Space Model, \
    \underline{\textit{U-Net}}: U-shaped Convolutional Neural Network, \underline{\textit{VAE}}: Variational AutoEncoder, \underline{\textit{VDM}}: Video Diffusion Model, \underline{\textit{ViT}}: Vision Transformer, \underline{\textit{VLM}}: Vision-Language Model, \underline{\textit{VQ}}: Vector Quantization, \underline{\textit{Ar}}: Autoregressive.
\end{itemize}
\vspace{0.1cm}
\resizebox{1\linewidth}{!}{
\begin{tabular}{lr|ccc|cc|c|c|c|c|c|c}
\toprule
 & \multicolumn{1}{c|}{\multirow{2}{*}{\textbf{Model}}} & \multicolumn{5}{c|}{\textbf{Decision Support}} & \multicolumn{2}{c|}{\textbf{Training Facilitation}}   & \multirow{2}{*}{\textbf{Training Data}} & \multirow{2}{*}{\textbf{Input}} & \multirow{2}{*}{\textbf{Output}} & \multirow{2}{*}{\textbf{Core Components}} \\
\cline{3-9}
 & & \textbf{AP} & \textbf{PL} & \textbf{VP} & \textbf{Static} & \textbf{Action-cond.} & \multirow{1}{*}{\textbf{Data Engine}} & \multirow{1}{*}{\textbf{Evaluation}} & & & & \\
\hline
\hline  
\rowcolor{blue!5}
 & PlaNet~\cite{hafner2019learning} & \checkmark & \checkmark  &  &  &  & & & V+A & V+A &A  &RSSM\\
 & DreamerV1,V2,V3~\cite{hafnerdream,hafner2021mastering,wu2023daydreamer,hafner2023mastering} & \checkmark & \checkmark  &  &  &  & & & V+A & V+A &A  &RSSM\\
 \rowcolor{blue!5}
 & PaLM-E~\cite{driess2023palm} & \checkmark &  &  &  &  & & & V+L & V+L &A  &LLMs, ViTs\\
 & OpenVLA~\cite{kim2025openvla} & \checkmark &  &  &  &  & & & V+L+A & V+L &A  &VLA\\
  \rowcolor{blue!5}
 & Plan2Explore~\cite{sekar2020planning} & \checkmark & \checkmark  &  &  &  & & & V+A & V+A &A  &RSSM\\
 & FOCUS~\cite{ferraro2025focus} & \checkmark & \checkmark  &  &  &  & & & V+A & V+A &A  &RSSM\\
  \rowcolor{blue!5}
 & EgoAgent~\cite{chen2025egoagent} & \checkmark & \checkmark  &  &  &  & & & V & V+A &A  &JEPA\\
 & GR00T ~\cite{bjorck2025gr00t} & \checkmark & \checkmark  &  &  &  & & &V+L+A+S & V+L+A+S &A  &VLA, VLM\\
   \rowcolor{blue!5}
 & THICK~\cite{gumbsch2023learning} & \checkmark & \checkmark  &  &  &  & & & V+A & V+A &A  &Hierarchical, RSSM\\
 & DayDreamer~\cite{wu2023daydreamer} & \checkmark & \checkmark  &  &  &  & & & V+A & V+A &A  &RSSM\\
 \rowcolor{blue!5}
 & RetryingVisualMPC~\cite{ebert2018robustness} & \checkmark &  &  &  &\checkmark  & & & V & I  &A  & Registration network \\
 & Genie~\cite{bruce2024genie} &  & \checkmark &  &  & \checkmark&  &  & V & I/V  &V  & VQ-VAE, VQ-VAE  \\
\rowcolor{blue!5}
 & GE~\cite{liao2025genie} & &\checkmark  &\checkmark  & \checkmark  &\checkmark  &\checkmark  &  & V + L  & L+I  &V, Ar  & VAE, DiT, CLIP\\
 & GR-2~\cite{cheang2024gr} & \checkmark&  & \checkmark & \checkmark &  & \checkmark &  & V  &  V+L+S & V+A+S, Ar & CLIP, VQGAN, GPT, condition VAE\\
\rowcolor{blue!5}
 & UniSim~\cite{yang2023learning} &  & \checkmark  & \checkmark &\checkmark &  &  \checkmark  &  &  V+L+A &  V+A & V & Stable Diffusion  \\
 & UniPi~\cite{du2023learning} &\checkmark  & &\checkmark  &\checkmark  &  &  &  &  V+L & L+I & V  & IDM, VDM \\
\rowcolor{blue!5}
 & DiVA~\cite{wang2025language} & &\checkmark &\checkmark  &\checkmark &  &  &  & V+L &V+L& V+A &VDM, CLIP, VAE \\
 & WorldVLA~\cite{cen2025worldvla} & &\checkmark&\checkmark & \checkmark& &\checkmark& &V+L & V+L+A & V+A, Ar & VLA, VQ-GAN \\
\rowcolor{blue!5}
 & iVideoGPT~\cite{wu2024ivideogpt} &  & &\checkmark & &\checkmark &  &\checkmark  &  V+A & V+A &V, Ar  & VQ-GAN, GPT\\
 & DreamVLA~\cite{zhang2025dreamvla} &\checkmark &\checkmark  &  &  & \checkmark &  &  &  V+L+A+S &V+L+S  &  A, Ar  & CLIP, Masked Autoencoders, LLMs (GPT-2), DiT\\
\rowcolor{blue!5}
 & Seer~\cite{tian2025predictive}  &\checkmark  & \checkmark  &  &  &\checkmark&  &  &V+L+A+S  &V+L+S  &  V+A, Ar  & CLIP, ViT, GPT\\
 & EnerVerse~\cite{huang2025enerverse} &\checkmark &  &  &  &\checkmark &  &  &V+L+A  &V+L  &  V, Ar  & VAE, VDM\\
\rowcolor{blue!5}
 & SayCan~\cite{ahn2022can} &\checkmark  & & & & &  &  & V+L+A &V+L  &A  & LLMs \\
 & FOCUS~\cite{ferraro2025focus} &\checkmark  &\checkmark  &  &  & \checkmark  & \checkmark  & \checkmark  & I+A+S & I+A+S & I+A  & RSSM\\
\rowcolor{blue!5}
 & Visual MPC~\cite{finn2017deep} &\checkmark & &\checkmark &  & \checkmark  &  &  & V+A+S  &V+A+S  &V+A  &LSTM\\
 & CDNA~\cite{finn2016unsupervised} &  & &  &  &\checkmark &  &  &  V+A+S &V+A+S  &V  &LSTM\\
\rowcolor{blue!5}
 & VisualForesight~\cite{ebert2018visual} &\checkmark  & &\checkmark &  &\checkmark  &\checkmark & \checkmark &  I & V+goal(I/L) & A  &LSTM \\
% visual planning
 & VLP~\cite{du2023video} &  & & \checkmark &  &\checkmark  &\checkmark &  &  I+V+L+A &I+L  &V+A  &  VLM, VDM \\
\rowcolor{blue!5}
 & FLIP~\cite{gao2024flip} &  & & \checkmark &  &\checkmark  &  & \checkmark &  V+L &V+L  &V  &CVAE, DiT, LIV \\
 & RoboDreamer~\cite{zhou2024robodreamer}  &  & &\checkmark &  &\checkmark  &\checkmark &  &  V+L &V+L  &V  &Text Parser, VDM, IDM \\
\rowcolor{blue!5}
 & COMBO~\cite{zhang2025combo}  &  & &\checkmark &  &\checkmark  &\checkmark & \checkmark &  V+L+A &V+A  &V+A  & VDM, VLM, Tree Search  \\
 & AVID~\cite{rigter2025avid}  &  &\checkmark &  &  &\checkmark  & & \checkmark&  V+A &V+A  &V  &VDM, Action Adapter \\
\rowcolor{blue!5}
 & SuSIE~\cite{black2024zero} &\checkmark  & & \checkmark &  &\checkmark  &  &  &  V+L+A &V+L  &V+A  &  Image-Editing Diffusion Model\\
 & 3D-VLA~\cite{zhen20243d} &\checkmark  & &\checkmark  & \checkmark &\checkmark &  &  &  V+L+A+S &V+L  &V+A+P  & 3D-LLM, Stable Diffusion \\
\rowcolor{blue!5}
 & AETHER~\cite{team2025aether} &  &\checkmark &\checkmark  & \checkmark  & \checkmark &  &  &  V &V+A  &V  &  Geometric Encoding, VDM \\
 & TesserAct~\cite{zhen2025tesseract} &\checkmark  &\checkmark  &  &  & \checkmark & \checkmark &  &  V+L+A &V+L  &V+A  &  VDM \\
\rowcolor{blue!5}
% trajectory based
 & IRASim~\cite{zhu2025irasim}&  &\checkmark &\checkmark  &  &\checkmark  &\checkmark &  &V+A  &V+Tr  &  V, Ar  &DiT\\
 & 3DFlowAction~\cite{zhi20253dflowaction}& \checkmark & &\checkmark &  &\checkmark  &  & &V+L  &V+L+P  &Tr, Ar &VDM, CLIP \\
\rowcolor{blue!5}
% #dynamic/action prediction, RSSM, RL-based
 & THICK~\cite{gumbsch2023learning}  &\checkmark  &\checkmark &  &  &\checkmark  & \checkmark & \checkmark &  V+A+S &V+A  & V+A+S  &Context-Specific RSSM \\
 & FlowDreamer~\cite{guo2025flowdreamer}  &\checkmark  &\checkmark &\checkmark  &  &\checkmark  &  &  &  V+A &V+A  &V  &U-Net, Stable Diffusion\\
\rowcolor{blue!5}
 & DREMA~\cite{barcellona2025dream}  & & &\checkmark  &  & \checkmark & \checkmark &  &  V+A+S &V+S &V+A  &  Gaussian Splatting \\
 & AVDC~\cite{ko2024learning}  & & & \checkmark &  &\checkmark  &\checkmark  &  &  V+L &V+L &V+Optical Flow+A  &U-Net, CLIP-Text, Optical Flow Model\\
\rowcolor{blue!5}
 & GWM~\cite{lu2025gwm}  &  &\checkmark &  &\checkmark  &\checkmark  & \checkmark & \checkmark &  V+A &V+A &V  &3D Gaussian VAE, DiT\\
 & Plan2Explore~\cite{sekar2020planning} &\checkmark &\checkmark &  &  &  & &  & V &V &A  & DM, Dreamer\\
\rowcolor{blue!5}
 & DreamGen~\cite{jang2025dreamgen}  & & &  &\checkmark  &  & &  & V &V+L &V  & DiT \\
 & UWM~\cite{zhu2025unified} & &\checkmark &  & \checkmark  &  & &  & V &V+A &V+P  &IDM, DiT \\
\rowcolor{blue!5}
 & CLOVER\cite{bu2024closed}  &\checkmark &\checkmark &\checkmark  &\checkmark & & & &V+L &V+L &V+A  & ViTs, CLIP,  VDM \\
 & Pandora~\cite{xiang2024pandora}  & & & & & \checkmark & &  &V+L &V+L &V, Ar  & LLM, DynamiCrafter Model \\
\rowcolor{blue!5}
 & PAR~\cite{song2025physical}  & & &  &  & \checkmark & &  &V+L &V+L+A &V+A, Ar  & 3D-VAE, Phi Model, DiT, Causal Transformer \\
 & RoboEnvision~\cite{yang2025roboenvision}  &\checkmark & &  &\checkmark  &  & &  &V+L &V+L &V+A  & VLM, DiT\\
\rowcolor{blue!5}
 & HMA~\cite{wang2025learning}  & & &  &\checkmark  &  &\checkmark &  &V+L &V+L &V+A  &  VLM, DiT \\
 & WorldGym~\cite{quevedo2025evaluating}  & & &  &  &  & &\checkmark  &V+A &V+A &V  &VLM, DiT \\
\rowcolor{blue!5}
 & VideoWorld~\cite{ren2025videoworld}  &\checkmark & &  &  &\checkmark  & &  &V &V &V+A, Ar  & IDM, VQ-VAE, Ar Transformer \\
 & PlaySlot~\cite{villar2025playslot}  & & &  &  &\checkmark &  &  &V &V &V, Ar  & IDM, Recursive Encoder-Decoder, Ar Transformer \\
 \rowcolor{blue!5}
&LOReL \cite{nair2022learning} & & &\checkmark  &  &\checkmark &  &  &V+L &V+A &V+A  & LSTM \\
& V-JEPA 2 \cite{assran2025v} &\checkmark & &\checkmark  &  &\checkmark &  &  &V+A &V+A &V+A  & JEPA, LLM  \\
 \rowcolor{blue!5}
&UP-VLA \cite{zhang2025up} &\checkmark & &  &\checkmark  & &  &  &V+L &V+L &V+A+L  &VLA, Decoder  \\
&CoT-VLA \cite{zhao2025cot} &\checkmark & &  &\checkmark  & &  &  &V+L &V+L &V+A  &VLA, Decoder  \\
\toprule
\end{tabular}}
\end{table*}
%-------------------------------------------------------------------------

\section{Preliminaries} \label{Prelim}
\subsection{What Is the “World” to Be Modeled?}
Despite the debate among philosophers about the ultimate nature of the world, the world can be roughly described as a set of entities, each with its own attributes or properties, along with the relationships and interactions that connect them. These attributes, such as shape, size, material, or state, and the connections, can be spatial, causal, functional, or temporal. As a result, objects, agents, and features are not only statically arranged but also evolve and influence one another over time. In order to interact effectively with such a world, an intelligent agent must capture critical information about entities, their properties, and their interactions across spatial and temporal dimensions. Collectively, these entities and interactions form a rich and dynamic environment in which an agent must actively explore, interact, and learn to achieve its goals. This naturally raises the question of what fundamental capabilities underpin an agent’s ability to capture and reason about such complex dynamics, as well as what forms of representation, learning, and interaction are required to model and act within an uncertain and evolving world.

\subsection{World Models Empowering Robot Intelligence} 
Embodied intelligence refers to a system's ability to perceive, reason, and learn through direct interaction with its environment. Unlike traditional AI confined to abstract or symbolic domains, embodied intelligence integrates a physical body, sensors, actuators, and computational processes that jointly enable situated perception, reasoning, and action. Intelligent robotic agents serve as the primary physical instantiation of embodied intelligence. They inherently combine perception (via sensors), cognition (for learning and reasoning), and motor control (via actuators) to operate autonomously and acquire knowledge from real-world experience, much like biological organisms.

However, because intelligent agents perceive only a partial and noisy projection of reality through their sensors, many underlying relationships and causal dependencies remain latent. This limitation makes structured internal representations essential for prediction, planning, and multi-step reasoning. To achieve robust and efficient embodied intelligence, recent research introduces the notion of the “world model”, which serve as internal representations that capture environmental dynamics and common-sense regularities of how the world operates. By internally simulating potential outcomes, world models empower embodied agents to understand their context, anticipate the consequences of actions, and plan complex behaviors before execution, thereby reducing reliance on costly or unsafe real-world trial and error.

\subsection{Competing Perspectives on World Models}
Although the concept of the “world model” is prevalent in computer science, its definition remains unsettled, with ongoing debate in the research community regarding its fundamental nature and role in intelligence \cite{xing2025critiques}. A central point of contention concerns its generative capability, as illustrated by NVIDIA \cite{nvidia_wm2025}, who define world models as systems that learn environmental dynamics from multimodal data and generate videos capturing spatial and physical properties. Emphasizing action dependence, Sudhakar \textit{et al.} \cite{sudhakar2024controlling,cen2025worldvla} characterize world models specifically as action-conditioned video generation models, distinguishing them from conventional video prediction. Similarly, Hafner \textit{et al.} \cite{wu2023daydreamer,hafnerdream,hafner2021mastering,hafner2023mastering,chen2025vlwm} identify action-conditioned prediction as a core feature of world models, emphasizing the prediction of latent representations rather than raw observations. Despite these differing perspectives, a common consensus emerges: world models aim to construct internal representations that capture environmental dynamics and action consequences, thereby enabling the prediction of future states.

%------------------------------------------------------------------------
\begin{figure}[tbh]
\begin{subfigure}
{\includegraphics[angle=0, width=0.45\textwidth]{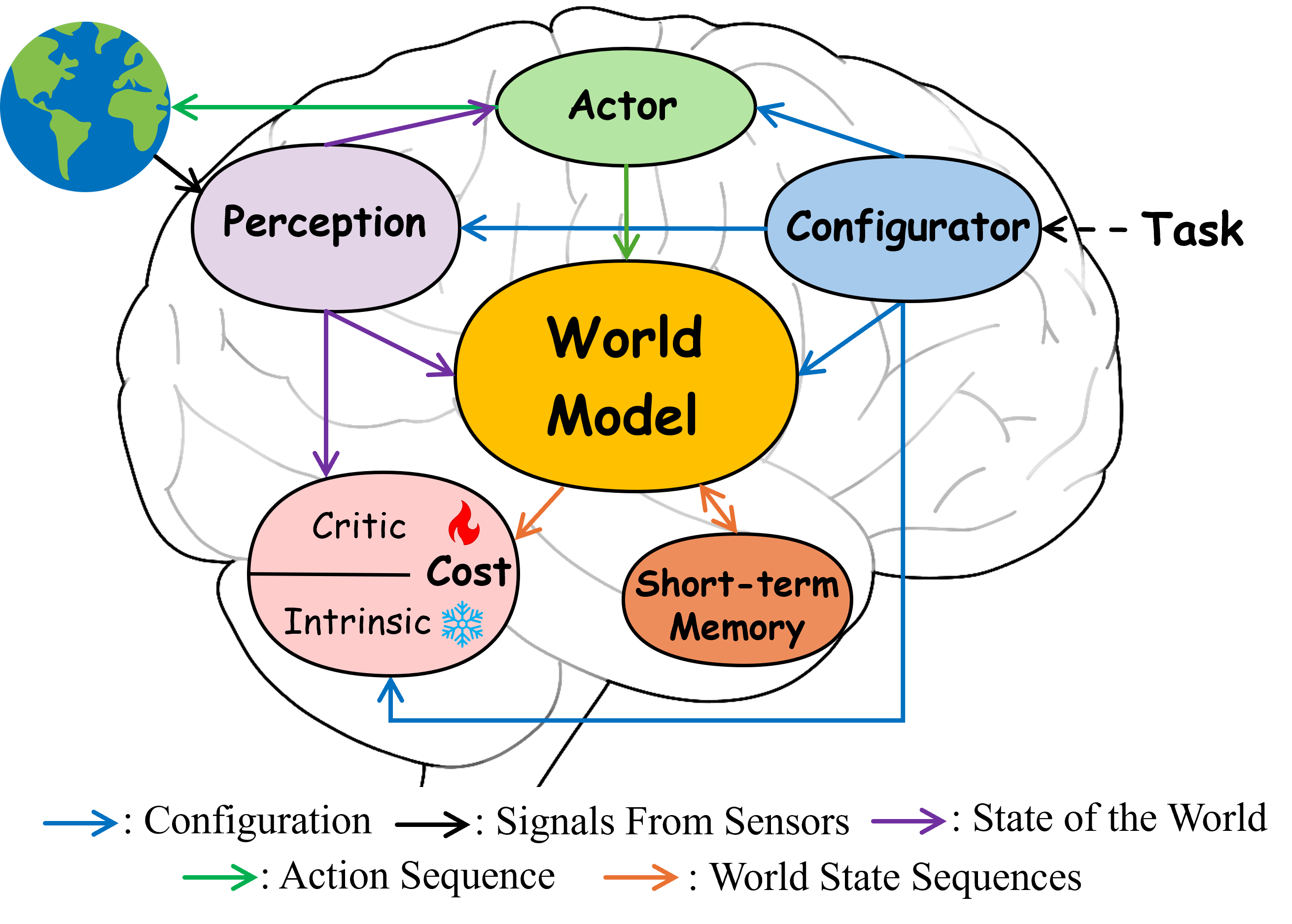}}
\end{subfigure}
\caption{A visualization of an agent \cite{lecun2022path}, where the world model predicts possible future world states as a function of imagined actions sequences proposed by the actor.}
\label{lecun_worldmodel}
\end{figure}
%------------------------------------------------------------------------

\subsection{Revisiting Modern AI Models Through the Lens of World Modeling}

The rapid progress of large-scale artificial intelligence models has blurred the boundaries between traditional task-specific learning and general world modeling. Although many of these models are not explicitly designed as world models, they exhibit key characteristics of world modelimodelling as learning structured representations of reality, reasoning about causality, and predicting or generating plausible future states. Revisiting these modern models through the lens of world modeling provides valuable insights into how intelligence emerges from data, embodiment, and multimodal integration. This perspective helps clarify which components of contemporary architectures, such as large language models (LLMs), vision-language models (VLMs), vision-language-action models (VLAs) and video generation models, implicitly capture aspects of the world and how they contribute to the broader goal of general-purpose world understanding.

\subsubsection{LLM, VLM \& VLA} \label{LLM_VLMA}
The strong reasoning capabilities and next-token prediction mechanism of LLMs make them natural foundations for constructing world models, as they capture sequential dependencies, causal relationships, and abstract dynamics. When equipped with auxiliary modules such as value functions \cite{ahn2022can} or modality-specific encoders \cite{xiang2024pandora,driess2023palm,zhang2025dreamvla}, LLMs can achieve a more comprehensive understanding of the environment. Moving beyond the language-centric paradigm, VLMs focus on the joint modeling of multiple modalities, providing a perceptually grounded understanding of the world \cite{hu2023look,zhao2024vlmpc}. Furthermore, an increasing number of studies have explored augmenting VLMs with low-level action-generation capabilities, thereby transforming them into VLA \cite{zitkovich2023rt,hong20233d,bjorck2025gr00t} that bridge perception, reasoning, and control. In addition to  action generation, there are also many work that enable additional visual prediction \cite{zhang2025up,zhao2025cot}.

From the above discussion, the designs and functions of LLMs, VLMs, and VLAs align with the spirit of world models, as they aim to represent and reason about world dynamics. Therefore, these models should not be excluded from the broader conceptual scope of world modeling. However, solely relying on LLMs, VLMs, or VLAs often constrains a system’s capacity for long-horizon prediction, reasoning, and imagination, all of which are essential for modeling dynamic and interactive environments. Recent studies have thus begun to integrate these models into architectures that explicitly function as world models, such as the JEPA framework \cite{chen2025egoagent}, Dreamer-style frameworks \cite{wang2025founder}, positioning them as core mechanisms for capturing temporal and causal dynamics.

\subsubsection{Video Generation Models} 
Video generation models primarily aim to produce visually realistic and temporally coherent sequences, which implicitly rely on learning the underlying dynamics of the environment. They can operate on diverse modalities, including language, visual data, and action inputs, allowing them to access environmental context and imagine future scenes. These characteristics position video generation models as a form of world modeling. Indeed, many recent world models adopt video generation as their core mechanism \cite{du2023video,wu2024ivideogpt,zhang2025combo}, enabling the prediction of future states encompassing observations, actions, and environmental changes. However, most video generation models focus on observation-level prediction and may suffer from issues such as a lack of interpretable internal representations and limited causal understanding of the world.

\begin{tcolorbox}[colback=blue!5!white, colframe=blue!70!white,title=Implications for Modern AI Models]
Viewing LLMs, VLMs, VLAs, and video generation models through the lens of world modeling reveals the shared objective across modern AI paradigms: constructing internal representations that capture the structure and dynamics of the world. This perspective reframes world modeling not as a separate task, but as the underlying principle driving the integration of perception, reasoning, and action in intelligent systems.
\end{tcolorbox}

\section{Overview of World Models}  \label{overview}

\begin{quote}
\textit{``What we observe is not nature itself, but nature exposed to our method of questioning.''}

\hfill --- Werner Heisenberg
\end{quote}

\subsection{Paradigms}

Building on the previous review of current models, contemporary architectures for capturing world dynamics can be broadly stratified along a methodological spectrum: implicit world modeling (e.g., LLMs, VLMs, and VLAs) \cite{chen2025egoagent,team2023internlm,zhen20243d,hong20233d}, latent dynamics modeling \cite{wu2023daydreamer,hafnerdream,hafner2021mastering,hafner2023mastering}, and video generation paradigms \cite{wu2024ivideogpt,du2023video,zhang2025combo,rigter2025avid}, each targeting distinct representational granularities and predictive mechanisms.

%------------------------------------------------------------------------
\begin{figure*}[tbh]
\center
\begin{subfigure}
{\includegraphics[angle=0, width=1\textwidth]{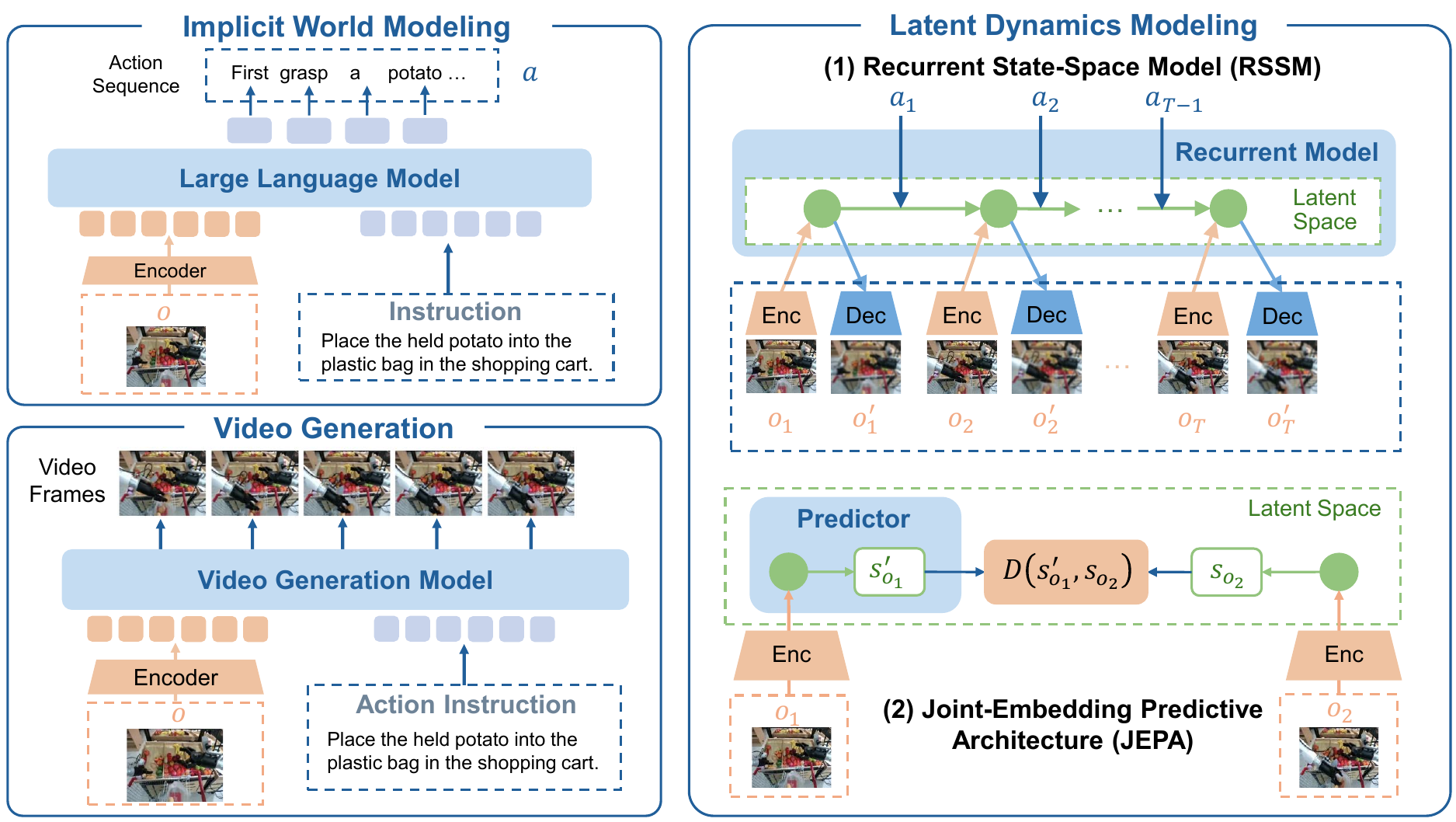}}
\end{subfigure}
\caption{An overview of world models. Implicit world models map observations and instructions directly to actions without explicitly modeling environmental dynamics. Latent-dynamics world models capture the evolution of environment dynamics within a latent space, while video-generation-based world models predict future visual scenes.}
\label{fig_llm}
\end{figure*}
%------------------------------------------------------------------------

\subsubsection{Implicit World Modeling}

Implicit World Modeling refers to the paradigm in which an agent learns to map sensory inputs (e.g., vision and language) directly to actions without constructing an explicit, disentangled representation of the environment’s states or dynamics. Representative models include LLMs, VLMs, and VLAs, which offer distinct advantages in semantic grounding, generalization, and interpretability \cite{ahn2022can,xiang2024pandora,driess2023palm,zhang2025dreamvla,zhao2024vlmpc}. An illustration of these models is shown in Fig.~\ref{fig_llm}. At the same time, these models can be integrated into broader world-modeling architectures to capture temporal dependencies and enable long-horizon prediction \cite{zitkovich2023rt,hong20233d,bjorck2025gr00t}. Detailed discussions are provided in Sections \ref{LLM_VLMA} and \ref{IWM}.

\subsubsection{Latent Dynamics Modeling}

Latent dynamics models typically encode high-dimensional observations into compact latent states through a variational autoencoder or encoder network, and employ recurrent or transformation modules (e.g., RNNs or Transformers) to predict the temporal evolution of these latent representations \cite{hafnerdream,hafner2021mastering,wu2023daydreamer,hafner2023mastering}. This architecture is characterized by latent-space imagination and task-oriented optimization over visual granularity, facilitating long-horizon learning by forecasting future states without the need for pixel-level reconstruction.

\noindent\textbf{Recurrent State-Space Model (RSSM)} \cite{hafner2019learning} resembles the structure of a partially observable Markov decision process. Its learning framework consists of three main components: an encoder, a decoder, and a dynamics model. The encoder network fuses sensory inputs (observations) $o$ together into the stochastic representations $z$. The dynamics model learns to predict the sequence of stochastic representations by using its recurrent state $s$. The decoder reconstructs sensory inputs to provide a rich signal for learning representations and enables human inspection of model predictions, but is not needed while learning behaviors from latent rollouts. Specifically, at time step $t$, let the image observation be $o_t$, the action vectors $a_t$ and the reward $r_t$. RSSM can be formulated as the generative process of the images and rewards conditioned a hidden state sequence $s_t$:
\begin{equation}\label{eq1-1}
\begin{split}
\begin{array}{llll}
& \text{Encoder/representation model:} &s_t \sim p_\theta \left(s_t \mid s_{t-1}, a_{t-1}, o_t\right) \\
& \text{Decoder/observation model:} &o_t \sim p_\theta \left(o_t \mid s_t\right) \\
& \text{Dynamics/Transition model:} &s_t \sim p_\theta \left(s_t \mid s_{t-1}, a_{t-1}\right)  \\
& \text{Reward model:} &r_t \sim p_\theta \left(r_t \mid s_{t}\right) \\
\end{array}
\end{split}
\end{equation}

\textbf{PlaNet} \cite{hafner2019learning} first demonstrates the effectiveness of learning dynamics in a latent space. The \textbf{Dreamer} family of models (a visualization is shown in Fig.~\ref{fig_llm}) \cite{hafnerdream,hafner2021mastering,wu2023daydreamer,hafner2023mastering} further verify this paradigm and establish a representative framework that reduces reliance on real-world data by performing imagination directly in latent space. Dreamer enables policy learning through imagined trajectories without continuous interaction with the environment, allowing agents to simulate multi-step consequences of actions and generalize to new states, objects, and environments.

While sharing the objective of learning predictive world-state representations, \textbf{Joint-Embedding Predictive Architecture (JEPA)} \cite{lecun2022path,chen2025egoagent} and RSSM diverge fundamentally in their learning mechanisms. RSSM relies on generative reconstruction of observations to model latent dynamics, whereas JEPA (a visualization is shown in Fig.~\ref{fig_llm}) employs self-supervised predictive coding in embedding spaces, directly forecasting future state representations without decoding to raw sensory inputs. This paradigm eliminates the computational cost of pixel-level reconstruction but necessitates powerful hierarchical encoders to compress sufficient environmental information into abstract embeddings. As a result, it introduces an implicit information bottleneck that demands careful architectural balancing to preserve task-relevant features. Under the JEPA framework, Assran \textit{et al.} \cite{assran2025v} combine pre-trained video models with an action-conditioned predictor to autoregressively predict future states and actions.

The \textbf{MuZero} series \cite{schrittwieser2020mastering,ye2021mastering,wang2024efficientzero} represent another form of latent-dynamics-based world modeling. Instead of modeling the complete environment dynamics, MuZero predicts only future quantities directly relevant to planning, such as rewards, values, and policies, given the complexity of real-world environments, and employs a tree-based search algorithm \cite{silver2018general} to select optimal actions.

\subsubsection{Video Generation}

Video-based generative models are powerful tools for capturing environmental dynamics and predicting future scenes. These models operate directly on high-dimensional raw observations, such as RGB images, depth maps, or force fields, treating the environment as a sequence of frames \cite{wu2024ivideogpt,team2025aether,yang2023learning,bruce2024genie,brooks2024video,xiang2024pandora,zheng2024open,zhou2024robodreamer,ali2025humanoid}. By generating future scenes, they can support a wide range of applications, including simulation, action prediction, and visual planning \cite{finn2017deep,ebert2018visual,wu2024ivideogpt,zhang2025combo,rigter2025avid}. Moreover, they can leverage large-scale pre-training to enhance generalization and improve sample efficiency \cite{rigter2025avid,team2025aether,rigter2025avid,wang2025language,jang2025dreamgen}. Depending on the input modality, world models can be constructed using action-conditioned video prediction models \cite{wu2024ivideogpt}, text-to-video models \cite{du2023video,yang2023learning,jang2025dreamgen,zhou2024robodreamer}, or trajectory-to-video models \cite{zhu2025irasim,cheang2024gr}. 

There are several architectural families of video-based world models. Diffusion-based world models generate videos by progressively denoising random noise through multiple iterative steps. Representative examples include U-Net-based models \cite{ho2022video,ko2024learning} and diffusion transformer (DiT)-based architectures \cite{ferraro2025focus,agarwal2025cosmos,zhu2025irasim,wan2025wan,yang2025roboenvision}. Autoregressive world models, in contrast, predict the next token or frame conditioned on previously generated ones, effectively modeling temporal dependencies in the sequence \cite{liao2025genie,wu2024ivideogpt,xiang2024pandora,huang2025enerverse,villar2025playslot,bruce2024genie,cheang2024gr}. Other architectures include variational autoencoder-based models \cite{bruce2024genie} and convolutional LSTMs \cite{finn2017deep,ebert2018visual}.

Autoregressive-based world models generate each step conditioned on previous outputs, allowing them to predict sequences of arbitrary length and making them well-suited for long-horizon predictions. However, they often suffer from error accumulation over extended sequences \cite{yang2025roboenvision} and may struggle to represent highly multi-modal distributions. In contrast, diffusion-based models generate samples through an iterative denoising process, enabling them to model complex, multi-modal distributions and produce globally coherent sequences. This iterative refinement also makes diffusion models more robust to individual prediction errors, resulting in better performance on tasks requiring long-horizon consistency or high-quality generative outputs. On the downside, diffusion models are computationally intensive and slower during inference, and adapting them to sequential prediction requires careful conditioning. Overall, autoregressive world models tend to excel in scenarios demanding speed and accurate short-term predictions. In contrast, diffusion models are more suitable for tasks such as long-horizon prediction, where maintaining global coherence is crucial.

Compared with implicit and latent-space world models, video generation models provide more detailed visual predictions but at a higher computational cost, lower generation speed and sample efficiency. In addition, action predictions are only proved to be align with visual future generation \cite{wang2025learning}, as visual data contains relevant information to actions.

\subsection{Architectural Design} 

\subsubsection{Flat Architecture} 
Most current methods adopt flat architectures \cite{guo2025flowdreamer,ferraro2025focus,villar2025playslot,bruce2024genie,brooks2024video,xiang2024pandora,zheng2024open,zhou2024robodreamer}, which face critical limitations. They lack structured representations of the environment, resulting in poor handling of multi-scale dynamics, limited long-horizon prediction, error accumulation, and reduced generalization. Specifically, in robotic manipulation, placing fragile objects requires the robot to react instantly to unexpected slips while simultaneously planning the sequence of pick-and-place actions to achieve the overall goal. Many tasks further involve long-term objectives that must be completed through sequential subgoals and temporally extended actions. For example, assembling a piece of furniture requires picking up components, aligning and attaching them correctly, and tightening screws for each part. Moreover, operating at a single level of abstraction causes small prediction errors to compound over time, degrading performance in long-horizon tasks. Finally, flat architectures fail to extract high-level abstractions, limiting transferability across tasks and environments.

\subsubsection{Hierarchical Architecture} 
Several studies have begun to explore and design hierarchical world models. In this scheme, lower-level modules handle intermediate reactions and short-term predictions, while higher-level components are responsible for long-term planning and abstraction. Lecun \textit{et al.} \cite{lecun2022path} hypothesize a hierarchical JEPA architecture, where low-level and high-level representations are learned for short- and long-term predictions, respectively. Gumbsch \textit{et al.} \cite{gumbsch2023learning} propose an RSSM-based hierarchical world model, where the low-level module captures immediate dynamics for reactive control, and the high-level module models abstract temporal patterns for strategic planning. Björck \textit{et al.} \cite{bjorck2025gr00t} introduce a dual-system architecture in which System 2 interprets the environment and task goals, while System 1 generates continuous motor commands in real time. Similarly, Wang \textit{et al.} \cite{wang2025dmwm} design a dual-level world model consisting of an RSSM-based System 1 (RSSM-S1) and a logic-integrated neural network System 2 (LINN-S2). The inter-system feedback mechanism ensures that predicted sequences comply with domain-specific logical rules: LINN-S2 constrains RSSM-S1’s predictions, while RSSM-S1 updates LINN-S2 based on new observations, enabling dynamic adaptation. Wang \textit{et al.} \cite{song2025hume} further employ System 2 for value-guided high-level planning by estimating state-action values and selecting optimal actions, while System 1 executes real-time motions via cascaded action denoising.

Despite their advantages, hierarchical architectures introduce greater model complexity, higher computational cost, and increased training difficulty. Determining which goals or sub-tasks should be handled by high-level versus low-level modules remains challenging, as does designing appropriate architectures and preparing suitable training datasets. Moreover, maintaining effective information flow and coordination between layers is essential for stable and coherent performance. Consequently, developing hierarchical world models requires substantial effort in architecture design, goal decomposition, dataset construction, and inter-layer coordination.

\begin{tcolorbox}[colback=blue!5!white, colframe=blue!70!white,title=Implications for World Model Paradigms and Architectures]
World models can take diverse forms depending on the specific approach and task, but their fundamental objective remains the same: to model environmental dynamics and predict future states. Their design must balance efficiency, long-horizon reasoning, generalization across tasks, and the integration of multi-modal inputs. To capture both short-term reactions and long-term planning, hierarchical architectures are often adopted. However, such designs introduce additional challenges, including greater complexity, higher computational cost, difficult goal decomposition, demanding dataset preparation, and the need for effective inter-layer coordination.
\end{tcolorbox} 

\subsection{World Observation and Representation}
\subsubsection{Dimensionality of the World} 
In designing world models, the dimensionality of the environment is critical in shaping a model’s ability to capture spatial structures, temporal evolution, and causal dynamics.

Some works operate purely in 2D pixel space \cite{yang2023learning,bruce2024genie,brooks2024video,xiang2024pandora,zheng2024open,zhou2024robodreamer}, capturing visual appearance and short-term dynamics but ignoring the real-world geometry.
While 2D pixel-space models \cite{yang2023learning} capturing visual appearance and short-term dynamics but lacking geometric awareness of real-world structure. This limitation motivates the development of 3D-aware architectures.
To incorporate geometric understanding of the 3D world, Bu \textit{et al.} \cite{bu2024closed,ko2024learning,zhang2025combo} construct world models based on RGB-D data, while others extract richer 3D cues such as scene flow \cite{guo2025flowdreamer}, motion fields \cite{zhi20253dflowaction} and 3D point clouds with associated language descriptions \cite{zhen20243d}, enabling more comprehensive modeling of 3D world dynamics. Additionally, Lu \textit{et al.} \cite{lu2025gwm} leverage 3D Gaussian Splatting, Diffusion Transformers, and 3D Gaussian Variational Autoencoder to extract 3D representations from RGB observations. Zhang \textit{et al.} \cite{zhang2025dreamvla} incorporate depth estimation to enhance the understanding of 3D worlds. In addition to geometric structure, temporal dynamics are incorporated to construct 4D world models that jointly capture spatial and temporal evolution. For example, Zhu \textit{et al.} \cite{team2025aether} synthesize 4D data from RGB-D videos by estimating depth and camera pose.
Zhen \textit{et al.} \cite{zhen2025tesseract} leverage the pre-trained 3D VAE \cite{kingma2013auto} to encode RGB, depth, and normal videos and sum them together. Huang \textit{et al.} \cite{huang2025enerverse} employ 4D Gaussian splatting to model spatiotemporal dynamics in robotic environments.

\subsubsection{Observation Viewpoint of the World}
Robots acquire skills by observing and imitating humans or other agents in their environment. Depending on the observation viewpoint, world models for robot learning can be categorized into third-person (exocentric) \cite{guo2025flowdreamer,ferraro2025focus,villar2025playslot} and first-person (egocentric) \cite{chen2025egoagent,grauman2024ego} perspectives. Many existing methods learn from exocentric perspectives, capturing skills from an external viewpoint \cite{guo2025flowdreamer,ferraro2025focus,villar2025playslot}. However, exocentric observations do not fully align with how humans perceive the world. This has motivated the development of egocentric world models. For example, Chen \textit{et al.} \cite{chen2025egoagent} observe a continuous loop of human interactions, in which humans perceive egocentric observations and take 3D actions repeatedly. They model interactions as sequences of ``state-action-state-action'' tokens, processed using a causal attention mechanism. Zhang \textit{et al.} \cite{zhang2025combo} focus on multi-agent planning, inferring other agents' actions from world states estimated via partial egocentric observations.

Grauman \textit{et al.} \cite{grauman2024ego} argue that egocentric and exocentric viewpoints are complementary. Learning through egocentric viewpoints allows robots to better understand hand-object interactions and the attention mechanism of the camera wearer, while exocentric perspectives provide information about the surrounding environment and whole-body poses.

\subsubsection{Representation of the World}
A central aspect of world models lies in how the environment is represented, which directly influences their ability to reason about dynamics, predict future states, and generalize across tasks. World representations can be broadly categorized into scene-centric, object-centric, and flow-centric approaches. In scene-centric representations, the environment is encoded as a single holistic latent, typically learned directly from pixels or raw sensory inputs \cite{hafner2019learning,hafnerdream,hafner2021mastering,hafner2023mastering,hafner2025mastering}. While video generation tasks aim to maximize the visual fidelity of predicted sequences, robotic manipulation often does not require the full visual detail. Irrelevant elements such as the background or parts of the robot arm can be ignored. This motivates the use of object-centric representations, which focus on task-relevant entities and their interactions \cite{ferraro2025focus,barcellona2025dream,villar2025playslot,zhi20253dflowaction,hong2024multiply}. Flow-centric representations, in contrast, are designed to capture the motion dynamics of the environment, emphasizing temporal change and spatial displacement \cite{gao2024flip}.

\begin{tcolorbox}[colback=blue!5!white, colframe=blue!70!white,title=Implications for World Observation and Representation]
The world is inherently structured, requiring models to consider multiple dimensions in order to capture spatial and temporal dynamics, and to select appropriate observation viewpoints and representations according to task requirements. Each strategy provides distinct advantages, and effectively combining their complementary strengths while maintaining computational efficiency remains a central challenge.
\end{tcolorbox}

\subsection{Task Scope}

World models can also be categorized based on their task coverage. Some studies focus on single-task objectives, such as future-scene prediction \cite{sudhakar2024controlling,finn2016unsupervised,barcellona2025dream,jang2025dreamgen,ebert2018robustness}, or planning and action prediction \cite{sekar2020planning}.

In contrast, an increasing number of studies aim to support multiple tasks simultaneously, thereby enhancing the generality and applicability of world models.
For instance, Cheang \textit{et al.} \cite{cheang2024gr,zhou2024robodreamer,du2023video,gao2024flip} generate videos for future-scene prediction and accordingly infer corresponding actions. Other works pursue simultaneous action prediction and world-scene forecasting \cite{cen2025worldvla,chen2025egoagent,zhen20243d,song2025physical}. Beyond dual-task integration, several approaches extend world models to even broader capabilities. For instance, Bruce \textit{et al.} \cite{bruce2024genie} propose interactive video generation that supports environment prediction and imitation learning, and utilize a latent action model to infer policies from unseen, action-free videos. Liao \textit{et al.} \cite{liao2025genie} introduce a unified framework for embodied video generation, policy learning, and simulation. Lu \textit{et al.} \cite{lu2025gwm} learn 3D world representations for future-state prediction, imitation learning, and simulator through video generation. Zhu \textit{et al.} \cite{zhu2025irasim} develop an action-conditioned world model supporting trajectory-conditioned video generation, policy evaluation, and planning. Similarly, Huang \textit{et al.} \cite{huang2025enerverse} achieve multi-view video generation, robotic action prediction, and a data flywheel mechanism for sim-to-real adaptation.

\noindent\textbf{Would Foundation Models.} 
When discussing task scope, the notion of ``foundation world models'' becomes essential. These approaches aim to generalize across diverse tasks through large-scale training, paving the way for world models that act as universal backbones for robotics. One line of research achieves this through large-scale pretraining followed by task-specific fine-tuning \cite{lu2025gwm,wu2024ivideogpt,agarwal2025cosmos,cheang2024gr,mazzaglia2024genrl}.
In particular, Mazzaglia \textit{et al.} \cite{mazzaglia2024genrl} integrate a foundation VLM with a generative world model to enhance multimodal generalization. Other works directly pursue large-scale end-to-end training to build general-purpose world models \cite{bruce2024genie,cen2025worldvla}.

\begin{tcolorbox}[colback=blue!5!white, colframe=blue!70!white,title=Implications for the Task Scope]
The task scope directly shapes the capabilities and practical functions of a world model. Single-task models can achieve high performance but offer limited generalization, whereas multi-task models support diverse tasks at the cost of efficiency and potential task-specific performance. The choice between single-task and multi-task world models depends on task complexity, generalization requirements, available data, computational resources, and modularity, with multi-task models often preferred for flexible and reusable robotic systems.
\end{tcolorbox}

\section{Functions of World Models} \label{fwm}
World models play a central role in modern robotics by providing an internal predictive understanding of the environment. They enable robots to reason about future states, anticipate the consequences of actions, and perform evaluations, which are particularly valuable in real-world settings where interactions are costly, risky, or time-consuming. By modeling environmental dynamics, world models form the foundation for autonomous, adaptable, and efficient robotic systems. In robotic manipulation, world models serve two complementary functions: decision support, by predicting future scenes, actions and planning, and training facilitation, by generating data or acting as learned simulators. These roles are often closely related. For example, a world model used as a simulator can simultaneously generate training data and assist decision making \cite{lu2025gwm,liao2025genie}. By combining these functionalities, world models provide a comprehensive framework that enables robots to act intelligently, learn efficiently, and adapt to complex and dynamic environments. Additional details are provided in Table~\ref{tab_rwm}, which complements the following discussion.

\subsection{Decision Support}

\subsubsection{Implicit World Models for Action Prediction and Planning} \label{IWM}
This line of work explores world models that enable action prediction and planning without explicitly modeling state transitions or world dynamics. These approaches typically leverage the strong reasoning and next-token prediction capabilities of Large Language Models (LLMs), Vision-Language Models (VLMs), and Vision-Language-Action (VLA) models. Since LLMs lack direct access to environmental or robotic states, auxiliary components are often incorporated to provide grounding. For example, Ahn \textit{et al.} \cite{ahn2022can} introduce affordance functions to evaluate the feasibility of skills for completing a target task. Xiang \textit{et al.} \cite{xiang2024pandora,driess2023palm} employ encoders to process environmental information, while Zhang \textit{et al.} \cite{zhang2025dreamvla} integrate multimodal tokens including states, images, and text to enhance reasoning and generalization. Zhang \textit{et al.} \cite{huang2024embodied} further combine 2D and 3D encoders to process RGB images and 3D point clouds, capturing complementary spatial cues for richer world understanding. Hong \textit{et al.} \cite{hong2024multiply} extend this paradigm by incorporating additional sensory modalities such as vision, audio, tactile, and thermal inputs to achieve a more comprehensive understanding of the environment.

Conventional LLMs are language-centric and typically treat visual and other sensory information as auxiliary inputs. VLMs extend this paradigm by jointly learning aligned visual and linguistic representations, enabling grounded perceptual understanding of the world \cite{hu2023look}. Zhang \textit{et al.} \cite{zhao2024vlmpc} further leverage VLMs to generate candidate action sequences, which are evaluated using a lightweight action-conditioned video prediction model to forecast future scenes. The predicted outcomes are then assessed by the VLM to select the final action. An increasing number of studies extend VLMs to VLA by equipping them with low-level action generation capabilities. 
For instance, Zitkovich \textit{et al.} \cite{zitkovich2023rt} represent robot actions as a form of language, effectively bridging perception and control through textual grounding. Zhen \textit{et al.} \cite{zhen20243d} employ a 3D-based LLM \cite{hong20233d} to represent and predict 3D world states and generate actions, incorporating a diffusion model to synthesize future scenes. Inspired by the dual-process theory of human cognition \cite{kahneman2011thinking}, Björck \textit{et al.} \cite{bjorck2025gr00t} design a dual-system architecture in which a VLM serves as the reasoning module (System 2) and a Diffusion Transformer functions as the action module (System 1), with both components jointly optimized for coordinated reasoning and actuation. Zhou \textit{et al.} \cite{zhou2025vision} preserve the reasoning capability of VLMs while introducing a Mixture-of-Experts to alleviate conflicts between multimodal understanding and robotic manipulation in the parameter space. Kim \textit{et al.} \cite{kim2025openvla} train their model on a large corpus of real-world robot demonstrations, enabling efficient adaptation to new robotic platforms through parameter-efficient fine-tuning.

To further enhance long-horizon prediction, reasoning, and imagination, several methods integrate large language or multimodal models into other world model architectures, where they serve as core components. For instance, Chen \textit{et al.} \cite{chen2025egoagent} employ the open-source LLM, i.e., InternLM \cite{team2023internlm}, to predict future states from egocentric observations as a fundamental element of the JEPA framework. Similarly, Vision-Language Models \cite{mazzaglia2024genrl} and Video-Language Models \cite{wang2025founder} have been incorporated into Dreamer-style architectures for low-level dynamics modeling, where they extract high-level semantic knowledge of the world to guide prediction.

Notably, LLMs, VLMs, and VLAs can also act as explicit world models that predict future scenes \cite{zhang2025up,zhang2025dreamvla,zhao2025cot} or world knowledge \cite{zhang2025dreamvla}. We will elaborate them in Section~\ref{vbapp}.

\subsubsection{Latent Dynamics Modeling for Action Prediction and Planning}

This line of research focuses on modeling the temporal evolution of environment dynamics within a latent space, facilitating efficient action prediction, planning and future imagination. Operating in a compact latent space requires fewer environment interactions and reduces computational cost compared to pixel-based modeling. Hafner \textit{et al.} \cite{hafner2019learning,hafnerdream,hafner2021mastering,wu2023daydreamer,hafner2023mastering} introduce online planning in latent space through the Recurrent State-Space Model (RSSM), which learns to reconstruct input observations \cite{hafner2019learning}. The Dreamer series \cite{hafnerdream,hafner2021mastering,wu2023daydreamer,hafner2023mastering} introduces latent imagination, allowing agents to predict and plan over latent trajectories instead of pixels for more efficient and stable policy learning. Specifically, DreamerV1 \cite{hafnerdream} learns long-horizon behaviors from images by jointly predicting actions and state values, greatly improving sample efficiency. DreamerV2 \cite{hafner2021mastering} extends this framework to discrete environments by introducing binary latent variables, achieving human-level performance on the Atari benchmark. DreamerV3 \cite{hafner2023mastering,hafner2025mastering} further improves scalability and generative capacity through techniques such as symlog normalization for reward stabilization, refined KL balancing, and enhanced replay buffers. Sekar \textit{et al.} \cite{sekar2020planning} enhance generalization to downstream tasks through self-supervised learning without task-specific rewards, while Wu \textit{et al.} \cite{wu2023daydreamer} deploy Dreamer in the real world without simulators. Gumbsch \textit{et al.} \cite{gumbsch2023learning} introduce context-sensitive dynamics via a context-specific RSSM and hierarchical architecture to improve scalability and long-horizon prediction. Ferraro \textit{et al.} \cite{ferraro2025focus} develop object-centric world models for improved interaction reasoning.

Under the JEPA framework, Chen \textit{et al.} \cite{chen2025egoagent} capture causal and temporal dependencies by organizing states and actions into an interleaved sequence, integrating future state prediction and action generation within a unified transformer architecture. Building on this, Assran \textit{et al.} \cite{assran2025v} leverage pre-trained video encoders optimized with a masked denoising objective as the core of JEPA, enabling self-supervised learning through an action-conditioned predictor that autoregressively forecasts future states and actions. Incorporating other potential world models, such as LLMs and VLMs, have been introduced in Section~\ref{IWM}.

There are also approaches that couple Model Predictive Control (MPC) with learned world models, where the predictive model is used to simulate future trajectories and select optimal actions in a receding-horizon manner. For example, Hansen \textit{et al.}~\cite{hansen2022temporal} learn task-specific latent dynamics models using temporal-difference objectives and apply them for efficient online Model Predictive Control. Hansen \textit{et al.}~\cite{hansen2024td} further improve generalization across diverse embodiments and action spaces by learning an implicit, control-centric dynamics model.

\subsubsection{Vision-based Action Prediction and Planning} \label{vbapp}

Vision-based methods enable robots to predict future observations from sensory inputs, allowing them to plan actions in complex and unstructured environments. By simulating sequences of visual outcomes, robots can evaluate long-horizon behaviors, integrate multiple modalities (e.g., vision, language, and control), and generalize to novel tasks without task-specific retraining. This predictive capability makes visual imagination a key component of goal-directed and adaptive robotic decision-making. In particular, action-conditioned multi-frame prediction serves as a crucial element of prediction and planning, allowing robots to mentally simulate the outcomes of different actions before selecting the optimal one for a given task. According to the task formulation, existing approaches can be broadly classified into vision-conditioned and language-conditioned goal representations. 

\noindent\textbf{Vision-Conditioned Goals.} 
Finn \textit{et al.}~\cite{finn2016unsupervised} learn to predict motion dynamics that remain consistent across visual appearances, aiming to enable long-range, action-conditioned video prediction and generalization to unseen objects. Ebert \textit{et al.}~\cite{ebert2018robustness,ebert2018visual} improve long-horizon prediction by using an image registration–based cost function that continuously corrects errors during execution, achieving closed-loop visual planning. Bu \textit{et al.}~\cite{bu2024closed} further extend this idea with text-conditioned video generation to synthesize depth- and flow-consistent sub-goal images. A feedback mechanism then selects sub-goals and generates corresponding actions based on visual error evaluation, bridging visual planning and policy learning.

Imagining the future does not inherently produce actions. To enable action predictions, Finn \textit{et al.} \cite{finn2017deep,ebert2018robustness,ebert2018visual} incorporate visual prediction models with model-predictive control (MPC) to select the best action (sequence). Bu \textit{et al.} \cite{bu2024closed} use an error-measurement strategy to select the best sub-goal images and an MLP that is train with an Inverse Dynamics objective to decode the corresponding actions.

% #text-to-video
\noindent\textbf{Language-Conditioned Goals.} In \cite{finn2016unsupervised,finn2017deep,ebert2018robustness,ebert2018visual}, task specifications are provided as goal images, which are often difficult to obtain and prone to over- or under-specification. To address this limitation, a growing line of research leverages language as a more flexible, compact, and general medium for specifying tasks. However, translating language instructions into precise, actionable representations grounded in the robot’s observations remains challenging due to the misalignment between linguistic descriptions and visual perception. To bridge this gap, Nair \textit{et al.}~\cite{nair2022learning} use action-conditioned video prediction to simulate future scenes under different action sequences and learn a language-conditioned reward function from crowd-sourced descriptions to measure task completion. The best sequence is selected to maximize the reward. Zhang \textit{et al.} \cite{zhang2025up} take advantage of the semantic knowledge and reasoning
abilities of VLA and incorporate a decoder into VLA to enable future scene predictions and action generation. Zhou \textit{et al.}~\cite{zhou2024robodreamer} parse language instructions into compositional primitives to capture spatial object relations and generalize to novel commands, while also supporting multimodal task inputs such as goal images and sketches. Zhang \textit{et al.}~\cite{zhang2025dreamvla} enhance reasoning and generalization by introducing additional dream queries that capture historical information and predict dynamic regions, depth, and semantic maps using foundation models such as DINOv2~\cite{oquab2024dinov2} and SAM~\cite{kirillov2023segment}.

\noindent\textbf{Diverse Goals.} Some works leverage diverse goal conditions to improve task understanding and completion. For instance, Wang \textit{et al.}~\cite{wang2025language} develop a language–gesture-conditioned video generation model to disambiguate task specifications and integrate a behavior-cloning policy that unifies visual plan generation and manipulation. Du \textit{et al.}~\cite{du2023learning} incorporate observed images as additional context in each frame-denoising step to synthesize video plans and employ an inverse-dynamics model to infer the corresponding action sequences. Zhao \textit{et al.}~\cite{zhao2025cot} introduce visual chain-of-thought reasoning into VLA models by autoregressively generating sub-goal images alongside language instructions, enabling temporal planning and improving reasoning capability. 

\noindent\textbf{Action Inference.} A key advantage of vision–based action prediction is that it does not rely on large-scale action-labeled data. They can be pre-trained on large-scale video data and infer actions by training a simple action extractor using small amounts of action data. Techniques for action extraction include the inverse dynamics model \cite{liang2025video,du2023learning,zhou2024robodreamer}, a transformer encoder-decoder architecture \cite{wang2025language}. Zhang \textit{et al.} \cite{zhang2025combo} use vision language models to propose actions, and a tree search to find the best plan. However, video predictions would contain irrelevant information to the target tasks or actions to execute such as background and robot arm. To handle this, Zhi \textit{et al.} \cite{zhi20253dflowaction} extract 3D Flow from video data and learn 3D optical flow as a representation of object motions to guide action planning. Zhang \textit{et al.} \cite{zhang2025dreamvla} propose dynamic region-based forecasting, which leverages optical flow prediction model \cite{karaev2024cotracker,karaev2024cotracker3} to identify dynamic regions within the scene, enabling the model to concentrate on areas of motion that are critical for task execution instead of redundant frame reconstruction. Agarwal \textit{et al.} \cite{agarwal2025cosmos} leverage large-scale pre-training on both image and post-training for robotic manipulation, including instruction-based video prediction and action-based next-frame prediction. 3D positional embeddings, including 3D factorized Rotary Position Embedding and absolute positional embedding for relative positions and absolute coordinates respectively, are adopted to capture spatial and temporal information. Actions are predicted through an action embedder MLP. Tian \textit{et al.} \cite{tian2025predictive} propose an end-to-end Predictive Inverse Dynamics Models, which learn actions and visual futures synergistically to enhance the simulation and action predictions ability. \cite{guo2024prediction} predict both future frames and robot actions within joint latent denoising process, which support planning and acting in a closed-loop manner. 

\noindent\textbf{Visual Fidelity vs. Action Prediction.}
Guo \textit{et al.}~\cite{guo2025flowdreamer} hypothesize that models trained solely with frame-prediction losses tend to emphasize visual appearance fidelity while underestimating accurate dynamics modeling. This highlights the need for approaches that explicitly separate dynamics learning from visual rendering. To address this, FlowDreamer adopts a two-stage framework that first predicts environment dynamics and then renders corresponding visual observations.

\subsection{Training Facilitation} 
World models can act both as data engines, generating synthetic trajectories that support imitation learning and reinforcement learning, and as evaluation modules that provide internal reward estimation or predictive feedback. Because many models combine these roles, it is difficult to assign them to a single category. Accordingly, when discussing each role, we introduce their complementary functions in parallel to highlight this overlap.

\subsubsection{Data Engine} 
Large-scale human teleoperation datasets have greatly advanced robot learning \cite{zitkovich2023rt,black2024pi_0,team2025gemini,bu2025agibot,bjorck2025gr00t,liu2025rdt}.
However, collecting such data is labor-intensive and limits coverage across diverse environments and tasks.
Vision-based world models, particularly video world models, offer an alternative by learning environment dynamics and generating synthetic data. These models can be broadly divided, according to whether they are conditioned on actions, into static video generation models \cite{jang2025dreamgen}, which predict general future scenes, and action-conditioned video generation models, which simulate how actions change the environment. Beyond data generation, video-based world models increasingly support diverse tasks such as planning, policy learning, and action prediction, which will be reflected in the following content. 

Specifically, Du \textit{et al.} \cite{du2023learning} target to enable visual world imagination, action planning and generating video demonstration for training by learning a text-conditioned video generation model. Wu \textit{et al.}~\cite{wu2024ivideogpt} train a large-scale video world model to generate accurate and realistic simulated experiences, enabling video prediction, visual planning, and policy training. Jang \textit{et al.} \cite{jang2025dreamgen} propose to leverage video world models \cite{wan2025wan} to generate robot video data. They first fine-tune video world models on a target robot to capture the embodiment-specific dynamics and kinematics and prompt the model with to initial frames and language instruction to generate corresponding data. Pseudo-action labels are generated by means of either a latent action model \cite{ye2025latent} or an inverse dynamics model \cite{baker2022video}. Lu \textit{et al.} \cite{lu2025gwm} leverage 3D-GS reconstruction with Diffusion Transformers to effectively model 3D dynamics, which can promote future scenes generation to support imitation learning and reinforcement learning. Ye \textit{et al.} \cite{Ye2025GigaBrain} synthesize data from diverse perspectives to introduce variations in texture, illumination, viewpoints, physical properties, task diversity, and interaction patterns. Their approach includes: (i) re-rendering real trajectories with diverse visual content, (ii) generating viewpoint-consistent multi-camera scenes with pose adjustments, and (iii) synthesizing embodied interaction sequences, such as converting first-person human videos into robot-centric demonstrations. To ensure realism and avoid hallucination artifacts, the authors further leverage a set of quality assessment metrics that evaluate geometric consistency \cite{liu2025robotransfer}, multiview consistency \cite{liu2025robotransfer}, text–scene alignment \cite{azzolini2025cosmos}, and physical plausibility \cite{azzolini2025cosmos}. When constructing world models for training data generation, it is unrealistic to expect any training distribution to encompass all possible configurations of the world. To handle this, Barcellona \textit{et al.} \cite{barcellona2025dream} construct a compositional world model to generate novel demonstration data for training by combining Gaussian Splatting \cite{kerbl20233d} and physics simulators. Equivariant transformation is leveraged to augment data, which modify both observations and the corresponding action sequences to ensure semantical consistency. Wang \textit{et al.} \cite{xiao2025world} present a video-based world model capable of predicting future visual observations conditioned on VLA-generated actions. A VLM-guided instant reflector serves as a reward function that quantifies task completion through the semantic alignment between the predicted trajectory and the textual instruction. 

% \noindent\textbf{Support reinforcement learning based Robotics.} 
Despite recent progress, existing methods continue to face challenges in generating diverse and counterfactual data that remain physically plausible, thereby limiting the quality and diversity of synthetic datasets \cite{bjorck2025gr00t}.

\subsubsection{Evaluation} 
Traditionally, robot control policies have been developed and evaluated using handcrafted physics simulators \cite{todorov2012mujoco,erez2015simulation,tedrake2019drake}. However, such simulators rely on simplified or manually engineered dynamics, which struggle to capture complex real-world phenomena, particularly high-DoF interactions, deformable objects, and other non-rigid or contact-rich scenarios \cite{sunderhauf2018limits,afzal2020study,choi2021use}. Consequently, the resulting discrepancies between simulated and real environments, commonly referred to as the sim-to-real gap, have significantly hindered the deployment and generalization of robotic policies in practice \cite{dulac2019challenges,zhao2020sim}. To handle this, world models potentially emerge as a scalable, reproducible, and informative tool, which reduce reliance on trial-and-error in the real world. Compared to other filed such as autonomous driving \cite{dosovitskiy2017carla} and navigation \cite{deitke2020robothor}, simulated evaluation of robotic manipulation remains difficult because of the highly varied and dynamic interactions that arise between the agent and its environment. Li \textit{et al.}~\cite{li2025worldeval} leverage a video generative world model~\cite{wan2025wan} to produce videos based on action representations from a policy network. A success detector \cite{team2023gemini} is then used to evaluate task completion from the generated videos and corresponding text prompts. Quevedo \textit{et al.} \cite{quevedo2025evaluating} evaluate robot polices by means of Monte Carlo rollouts in the world model and take a vision-language model, i.e., GPT-4o \cite{hurst2024gpt}, as the reward model. He \textit{et al.} \cite{he2025pre} introduce a frame-level control and a motion-reinforced training to improve action-following ability and temporal, dynamic consistency, enhancing the dynamic prediction and action responsiveness of world simulator. More valuable transitions are discovered for policy learning. Zhu \textit{et al.} \cite{zhu2025irasim} construct a frame-level, action-conditioned video world model based on a Diffusion Transformer, enabling scalable policy evaluation, planning, and future-scene generation. Liao \textit{et al.} \cite{liao2025genie} take an action-conditioned video generator as the core to model the spatial, temporal, and semantic regularities of real-world interactions that are fundamental to robotic manipulation. The base world model can support future scene generation, action predictions, data engine and closed-loop policy evaluation. Wang \textit{et al.} \cite{wang2025learning} promote the versatility of video world models for policy evaluation, visual simulation, synthetic data generation by perform training on heterogeneous actions data with a shared spatial-temporal transformer.

Escontrela \textit{et al.} \cite{escontrela2023video} train an autoregressive transformer-based video prediction model and use the next-token likelihoods of the frozen model as a general \textbf{reward} function across diverse tasks.

% \noindent\textbf{World model as reward model.}

\begin{tcolorbox}[colback=blue!5!white, colframe=blue!70!white,title=Implications for Functions of World models]
World models advance robotic learning by providing a unified predictive core that supports both decision-making and training. This highlights their growing importance and motivates efforts to build foundational world models capable of supporting diverse downstream tasks. However, differing objectives, such as pixel-level video generation and action-centric prediction, impose competing requirements on representations, suggesting that a single model must carefully balance fidelity, controllability, and task relevance.
\end{tcolorbox}

\section{Key Techniques and Notable Challenges}    \label{ktnc}
This section summarizes the key techniques that drive the development of world models and discusses the major challenges that remain in achieving general, scalable, and robust modeling. Some techniques and concepts are revisited across subsections to emphasize their central importance.

\subsection{Data Limitations}
World models require large amounts of data and supervision to learn generalizable representations of world dynamics and support diverse tasks. However, collecting real-world robotic data is labor-intensive and costly, and the available data are often heterogeneous in format and modality. To overcome these limitations, a variety of strategies have been proposed.

\subsubsection{Training Data Scarcity}

\paragraph{Leveraging Pre-trained Models} 
Given the limited availability of training data, many approaches leverage existing pre-trained models. For example, Xiang \textit{et al.}~\cite{xiang2024pandora} bypass the need for training from scratch by integrating a pre-trained LLM and a pre-trained video model, requiring only lightweight fine-tuning. Zhu \textit{et al.}~\cite{zhu2025irasim} initialize IRASim with the pre-trained weights of OpenSora~\cite{zheng2024open} to expedite training. Similarly, Sudhakar \textit{et al.}~\cite{sudhakar2024controlling} leverage a pre-trained diffusion model, while Wang \textit{et al.}~\cite{wang2025language} utilize Stable Video Diffusion, fine-tuned with robotic videos to adapt to the robotics domain. Song \textit{et al.}~\cite{song2025physical} further exploit the world knowledge embedded in pre-trained autoregressive video generation models such as NOVA~\cite{deng2025autoregressive}.

\paragraph{Incorporating Auxiliary Data Sources} 
Some works tackle the shortage of robot data by using other available sources, such as human manipulation datasets. For instance, Zhi \textit{et al.} \cite{zhi20253dflowaction} use both human and robot manipulation videos for training. However, these datasets often contain cluttered backgrounds and similar-looking objects. To address this, they apply optical flow constraints to make the learned representation embodiment-agnostic. Sudhakar \textit{et al.} \cite{sudhakar2024controlling} leverage an automatic hand segmentation method to obtain agent-agnostic data for robot learning. Others resort to more diverse data. For example, Yang \textit{et al.} \cite{yang2023learning} leverage diverse kinds of data, including objects, scenes, actions, motions, language, and motor control, and convert all actions into a common format.

\paragraph{Synthetic Data Generation} 
Instead of relying on real-world data, Deng \textit{et al.}~\cite{deng2025graspvla} synthesize large-scale action data to train their model. To address the scarcity of 4D data, the Aether team~\cite{team2025aether} generate RGB-D synthetic videos and develop a robust camera-pose annotation pipeline to reconstruct full 4D dynamics. Similarly, Zhen \textit{et al.}~\cite{zhen2025tesseract} build a 4D embodied video dataset that combines synthetic data with ground-truth depth, normal information and real-world data with estimated depth and normal maps obtained from off-the-shelf estimators.

\subsubsection{Heterogeneous Action Data}

World models should be able to understand different forms of actions and embodiments to ensure their real-world applications. A basic strategy is to utilize diverse datasets for training. However, the inherent cross-domain and cross-embodiment nature of datasets lead to heterogeneous actions data, including action spaces, action frequencies, and action horizon. For example, diverse embodiment (e.g., different degrees of freedom across robotic arms) and control interface (end effector (EEF) position for arms) would lead to actions of different forms. To handle this, Zheng \textit{et al.} \cite{zheng2025universal} learn to capture their shared structural features to obtain the generic atomic behaviors by means of vision language models. Similarly, Zheng \textit{et al.} \cite{wang2025learning} lean a share latent space for actions by decoupling observation and actions. More strategies can borrow from relevant fields \cite{doshi2025scaling,team2024octo,wang2024scaling}.

\subsubsection{Action Label Missing}
Action-labeled data, which are essential for learning action-conditioned future predictions~\cite{yang2023learning}, are particularly scarce in real-world settings.

\paragraph{Self-supervised Learning} 
Finn \textit{et al.}~\cite{finn2016unsupervised,finn2017deep} propose to learn pixel-level motion in a self-supervised manner, while Ebert \textit{et al.}~\cite{ebert2018robustness,ebert2018visual} leverage image-to-image registration between consecutive video frames to capture dynamics without explicit action labels. However, goal image-based learning presents several drawbacks: such goals are inconvenient for humans to specify, may over-constrain the desired behavior (leading to sparse rewards), or under-specify task-relevant information for non-goal-reaching tasks.

\paragraph{Action Label Extraction} 
Another approach to handling missing action labels is to infer them directly from unlabeled videos. More specifically, 
Bruce \textit{et al.}~\cite{bruce2024genie,gao2025adaworld} employ latent action autoencoders to extract latent actions in a self-supervised manner. In their studies, Bruce \textit{et al.}~\cite{bruce2024genie} sample actions uniformly, while Gao \textit{et al.}~\cite{gao2025adaworld} introduce biased action sampling to encourage broader exploration and enable action reuse across contexts.
Jiang \textit{et al.}~\cite{jang2025dreamgen} extract pseudo-actions using either a latent action model~\cite{ye2025latent} or an inverse dynamics model (IDM)~\cite{baker2022video}. Du \textit{et al.}~\cite{du2023learning,ren2025videoworld,villar2025playslot,ko2024learning} learn from unlabeled videos by training inverse dynamics models to infer actions or their embeddings. Ren \textit{et al.}~\cite{ren2025videoworld} further integrate an inverse dynamics module into a latent dynamics model to leverage rich temporal representations, improving the temporal consistency of predicted actions. Villar \textit{et al.}~\cite{villar2025playslot} predict latent actions from object-centric representations.

\paragraph{Other Strategies} 
Some works aim to leverage \textbf{pre-trained video models}. For instance, Rigter \textit{et al.}~\cite{rigter2025avid} adapt a pre-trained video diffusion model for action-conditioned world modeling by training a lightweight adapter, which is then fine-tuned on a small set of domain-specific, action-labeled videos. Black \textit{et al.}~\cite{black2024zero} similarly employ a pre-trained image-editing diffusion model to support video-based world modeling. In addition, Zhu \textit{et al.} \cite{zhu2025unified} design a \textbf{unified world model} that integrates the action and video diffusion processes within a unified transformer architecture using separate diffusion timesteps. This can enable learning from action-free video data. Ko \textit{et al.} \cite{ko2024learning} utilize \textbf{optical flow} extracted from videos, thereby circumventing the need for explicit action labels. 

\begin{tcolorbox}[colback=blue!5!white, colframe=blue!70!white, title=Implications for Data Limitations]
The scarcity of data can be alleviated by leveraging external datasets or pre-trained base models from other sources. The key challenge lies in bridging the gap between the source and target domains. Furthermore, it is desirable to uncover the underlying knowledge through supervised learning.
\end{tcolorbox}

\subsection{Perception and Representation}

Perception lies at the heart of robotic world models, enabling systems to interpret task instructions and transform raw sensory inputs into meaningful representations. These representations allow robots to understand structured environments and, in turn, predict, react, and plan effectively.

\subsubsection{Inputs}
\noindent\textbf{Language.} Task instructions are usually given in language. Many methods use pretrained models such as CLIP~\cite{bu2024closed,ko2024learning,radford2021learning,tian2025predictive}, Phi~\cite{javaheripi2023phi,song2025physical}, or conditional VAEs~\cite{song2025physical} to extract semantic representations from the instructions.

\noindent\textbf{Visual data.} Similarly, visual inputs are often processed using pre-trained visual encoders. For example, Tian \textit{et al.}~\cite{tian2025predictive} leverage pre-trained Vision Transformers (ViTs)~\cite{he2022masked} to process image observations.
Wu \textit{et al.}~\cite{wu2024ivideogpt} employ a conditional VQGAN that encodes only task-relevant dynamic information, such as the position and pose of moving objects, to reduce temporal redundancy across frames. An autoregressive, GPT-like transformer is then used to generate the next tokens, which are decoded into future frames.

\noindent\textbf{Action data.} Actions are sometimes represented as integer values, which lack the contextual richness. This limitation can prevent world models from accurately capturing the intended meaning behind actions. To address this, He \textit{et al.}~\cite{he2025pre} propose representing actions through language templates that explicitly encode their semantic meaning. In many cases, actions are instead expressed in natural language, as noted above. While this enables richer semantic representations, it also introduces challenges, such as instruction-following ambiguity, which are discussed in Section~\ref{IUF}.

\noindent\textbf{Diverse data inputs.} Robots need to gain a structured understanding of the world by jointly considering diverse sensory inputs. To achieve this, Song \textit{et al.}~\cite{song2025physical} embed images and robot actions into a unified physical space, enabling the model to capture the sequential evolution of both the robot and its environment. Hong \textit{et al.}~\cite{hong2024multiply} incorporate visual, auditory, tactile, and thermal modalities, projecting them into a shared feature space where a language model generates subsequent states and action tokens.

\subsubsection{Challenges}

\paragraph{Instruction Understanding and Following} \label{IUF}
Instructions convey task goals and can take various forms, including linguistic directives (natural language or structured text), visual cues (sketches, images, or demonstration videos), and others. Compared to image-based goals, textual descriptions provide a more abstract, compositional, and flexible way of specifying objectives, enabling better generalization, clearer intent communication, and more efficient human–robot interaction. Many recent works express target goals through text descriptions~\cite{du2023learning}. Ideally, language instructions should clearly describe the task and remain easily interpretable by the model. However, real-world scenarios often involve ambiguous or novel instructions, making effective interpretation and grounding critical for successful task execution.

\noindent\textbf{Ambiguous Instructions.}
In real-world scenarios, language instructions are often ambiguous (e.g., ``put this near here''~\cite{wang2025language}).
To resolve such ambiguity, Wang \textit{et al.}~\cite{wang2025language} use pointing gestures, interpreted through 2D gripper and object tracking, as an additional instruction modality.

\noindent\textbf{New Instructions.}
World models are constrained to make predictions based on language instructions similar to those encountered during training, limiting their ability to generalize to novel commands. To solve this problem, Xiang \textit{et al.}~\cite{xiang2024pandora} curate a large and diverse set of action-state sequences from re-captioned videos and simulations, and fine-tune world models on this data to improve instruction interpretation and generalize to novel commands and tasks. Li~\textit{et al.} \cite{zhou2024robodreamer} employ a text parser to decompose language instructions into primitives, separating actions and spatial relationships. This decomposition allows the model to flexibly recombine these components and generalize to previously unseen combinations of instructions. However, decomposing instructions into primitives can ignore their interrelationships. To address this, Li~\textit{et al.} \cite{li2025manipdreamer} represent each instruction as an action tree, capturing the hierarchical structure among primitives to better model task organization.

\paragraph{Raw Pixels Modeling vs. Concept Abstraction} Some studies suggest that humans make predictions based on abstract concepts rather than raw pixels~\cite{chen2025egoagent}.
Instead of converting images into discrete tokens~\cite{yang2023learning,wu2024ivideogpt}, Chen \textit{et al.}~\cite{chen2025egoagent} use learnable convolutional layers to project images into continuous semantic embeddings.
Song \textit{et al.}~\cite{song2025physical} adopt an open-source 3D variational autoencoder (Open-Sora~\cite{zheng2024open}) to obtain video representations. In contrast, another line of work operates directly in pixel space. For instance, Ko \textit{et al.}~\cite{ko2024learning} adapt a U-Net-based image diffusion model with factorized spatial–temporal convolutions~\cite{dhariwal2021diffusion} to jointly capture spatial and temporal information.

\paragraph{Task-irrelevant Issues}
Visual data often contain information irrelevant to the task, and models such as Vision Transformers (ViTs) may produce hundreds of features per image, affecting both efficiency and effectiveness. To address this, Tian \textit{et al.}~\cite{tian2025predictive} extract task-relevant features using a perceiver resampler~\cite{alayrac2022flamingo}. Ren \textit{et al.}~\cite{ren2025videoworld} learn compact visual representations that preserve fine-grained temporal dynamics through a causal encoder–decoder structure and quantization with a discrete codebook~\cite{mentzer2024finite}.

\paragraph{Spatiotemporal Awareness}
Understanding the world requires modeling how spatial structures evolve over time. To this end, several works design architectures that explicitly capture spatial and temporal dependencies. Tian \textit{et al.}~\cite{tian2025predictive} enhance token representations with learnable positional embeddings at each timestep to capture temporal information. Bruce \textit{et al.}~\cite{bruce2024genie} develop a spatiotemporal transformer composed of multiple spatiotemporal blocks to model spatial–temporal relationships in dynamic scenes. Ko \textit{et al.}~\cite{ko2024learning} adopt factorized spatiotemporal convolutions following the design of~\cite{ho2022video}. Zhang \textit{et al.}~\cite{zhang2025dreamvla} extract spatiotemporal patch representations using a masked autoencoder~\cite{he2022masked}. Other studies incorporate additional cues to better understand the three-dimensional structure of the environment. For example, Zhang \textit{et al.}~\cite{zhang2025dreamvla} estimate depth information using depth estimation techniques~\cite{yang2024depth} to enhance 3D spatial understanding. When encoding multi-view inputs, Liao \textit{et al.}~\cite{liao2025genie} augment each token with 2D rotary positional embeddings, view-specific learnable embeddings, and timestep encodings to promote spatiotemporal alignment while preserving viewpoint-specific distinctions.

\begin{tcolorbox}[colback=blue!5!white, colframe=blue!70!white, title=Implications for Perception]
World models should process and integrate diverse sensory inputs to build a coherent understanding of real-world dynamics.
While current models primarily rely on vision and language, incorporating additional modalities such as tactile and proprioceptive sensing is crucial for achieving comprehensive perception in complex environments. It is also important to consider which information to perceive and how to model its spatial and temporal structure.
\end{tcolorbox}

\subsection{Long-horizon Reasoning}

Many robotic tasks require coherent long-horizon reasoning, where achieving the final objective depends on executing a temporally consistent sequence of actions over extended time scales. Existing methods are limited in long-horizon predictions \cite{nair2022learning,ha2018world,hafner2019learning,hafner2021mastering,hafner2023mastering}. For example, Ha \textit{et al.} \cite{ha2018world,hafner2019learning,hafner2021mastering,hafner2023mastering} predefine temporal horizons to guide planning in their world models. In terms of video generation, existing methods still suffer from limited length (short-horizon future video) \cite{gao2024flip}. For example, Ko \textit{et al.} \cite{ko2024learning} predicts a fixed number (eight) of future frames with U-Net based diffusion model \cite{dhariwal2021diffusion}. Bruce \textit{et al.} \cite{bruce2024genie} can only memorize 16 frames and cannot produce consistent predictions. For autoregressive models, small prediction errors compound sequentially, leading to substantial inaccuracies in long-horizon forecasts.

\subsubsection{Closed-loop Learning}
A line of work enabling long-term planning/predictions by learning through interaction with feedback and adjusting their behaviour accordingly \cite{du2023video,bu2024closed} . For example, Ebert \textit{et al.} \cite{ebert2018robustness,ebert2018visual} utilize image-to-image registration between predicted video frames and both the start and the goal images with the average length of the warping vectors as a cost function. The model would continue to retry until the task is completed. Du \textit{et al.} \cite{du2023video} proposes a recursive planning framework comprising action proposal, video rollout generation, and evaluation. Vision–language models (VLMs) are used to propose potential next actions, while video generation models simulate multiple possible future rollouts. The resulting trajectories are then evaluated by the VLMs to select the optimal action. Du \textit{et al.} \cite{liao2025genie} design a neural simulator that predicts future visuals, enabling policy models to interact within a consistent environment. A sparse memory mechanism is leveraged to further enhance the consistency over the time. 

\subsubsection{Subgoals}
Pre-trained models possess a vast repository of commonsense and procedural knowledge that can be leveraged to decompose a high-level goal, often specified in natural language (e.g., "make a cup of coffee"), into a logical sequence of concrete sub-goals or skills. Bu \textit{et al.} \cite{bu2024closed} propose to promote long-horizon manipulation tasks by decomposing the goal into sub-goals and handling error accumulations by designing a real-time feedback mechanism. Yang \textit{et al.} \cite{yang2025roboenvision} leverage VLM to produce sub-goals and utilize coarse and fine video diffusion models to generate long-horizon videos. Chen \textit{et al.} \cite{chen2025robohorizon} utilizes an LLM to generate a multi-stage plan and design a LLM-based dense reward generator for sub-tasks, providing crucial guidance for long-horizon planning. 

\subsubsection{Hierarchical Structures}
Bu \textit{et al.} \cite{gumbsch2023learning}  propose hierarchical world models with Adaptive Temporal Abstractions that separate the modeling of dynamics into high-level and low-level latent states. The low-level model captures fine-grained, short-term dynamics for immediate reactions, while the high-level model abstracts over longer temporal horizons to represent extended dependencies and long-term goals. By dynamically adapting the temporal granularity of the high-level latent states, the model can efficiently plan and predict over long horizons while maintaining accurate short-term predictions through the low-level module.

\subsubsection{More Strategies.} Driess \textit{et al.} \cite{driess2023palm} provide a goal image in addition to language instructions. Du \textit{et al.} \cite{du2023video} propose to take advantage of long-horizon inference of VLMs and the low-level visual dynamic modelling ability of text-to-video models to handle long-horizon visual planning. A tree search over the space of possible video sequences to find proper long-horizon plans. Ren \textit{et al.} \cite{ren2025videoworld} lean compact representations for the visual world that preserve the detailed temporal dynamics by means of causal encoder-decoder and quantization with a discrete codebook \cite{mentzer2024finite}. 

\subsection{Spatiotemporal Consistency}
Spatiotemporal consistency plays a vital role in ensuring coherent and physically plausible predictions of future states. It guarantees that the model preserves object continuity, motion smoothness, and causal relationships across time, enabling stable video simulation and reliable dynamics forecasting.

\subsubsection{Data Perspective} In conditional video synthesis, Du \textit{et al.} \cite{du2023learning} incorporate the observed image as additional context when denoising each frame. Specifically, it adapts a temporal super-resolution diffusion architecture by tiling the conditioned visual observation across all timesteps. Each intermediate noisy frame is concatenated with the observed image throughout sampling, providing a strong spatial anchor that enforces consistent environmental states across time. Ko \textit{et al.} \cite{ko2024learning} concatenates the initial condition frame with all subsequent frames, providing a stable reference that preserves both the spatial layout and temporal evolution of the environment throughout the sequence. Zhen \textit{et al.} \cite{zhen2025tesseract} refine depth maps using normal integration to enhance spatial consistency. Optical flow is then calculated to ensure depth coherence across frames, maintaining consistent scene geometry over time.

\subsubsection{Model Perspective} Yang \textit{et al.} \cite{yang2025roboenvision} noted that in autoregressive predictions, standard spatiotemporal attention in video diffusion models degrades frame consistency due to limited long-range context. To address this, the temporal attention layers are replaced with 3D full attention layers, enabling computation of attention across all spatiotemporal tokens and better modeling of large motions. Additionally, the spatial attention layers are modified by reinjecting the VAE features of the first frame and computing cross-attention with the spatial tokens of the query features, further enhancing frame coherence.

\subsubsection{Memory Mechanism} 
Memory mechanisms preserve historical information, enabling the coherent evolution of spatial and temporal patterns over time. For example, Liao \textit{et al.} \cite{liao2025genie} design a sparse memory mechanism to provide long-term historical context, improving spatiotemporal consistency and task relevance. More information can refer to Section \ref{Memo}.

\subsection{Generalization}
Robots are expected to operate robustly in complex and novel environments, interacting with unfamiliar objects and performing tasks beyond their training distribution.  

\subsubsection{Data Scaling}
An intuitive and effective strategy to enhance generalization is to scale the diversity and volume of training data. For example, Cheang \textit{et al.} \cite{cheang2024gr} increase the number of pre-training videos from 0.8 million in \cite{wu2024unleashing} to 38 million. Assran \textit{et al.} \cite{assran2025v} expand the dataset from 2 million used by \cite{bardes2024revisiting} to 22 million videos. Wang \textit{et al.} \cite{wang2025learning} expand each of the 40 datasets by increasing trajectories from 10 up to $10^{6}$. Cheang \textit{et al.} \cite{cheang2025gr} train the model with web-scale vision-language data,  human trajectory data and robot trajectory data. Kevin \textit{et al.} \cite{intelligence2025pi_} leverage diverse mobile manipulator data, diverse multi-environment non-mobile robot data, cross-embodiment laboratory data, high-level subtask prediction, and multi-modal web data. Cheang \textit{et al.} \cite{barcellona2025dream,cheang2024gr} investigate \textbf{data augmentation} strategies to enhance generalization. In \cite{barcellona2025dream}, object rotation and roto-translation are applied. Cheang \textit{et al.} \cite{cheang2024gr} generate novel scenes by injecting objects using a diffusion model \cite{ho2020denoising} and/or altering backgrounds with the Segment Anything Model (SAM)  \cite{kirillov2023segment}. A video generation model \cite{kirillov2023segment} is subsequently employed to synthesize videos that preserve the original robot motions from the inpainted frames. Liao \textit{et al.}   \cite{liao2025genie} augment the dataset with a diverse set of failure cases, including erroneous executions, incomplete behaviors, and suboptimal control trajectories—collected from both human teleoperation and real-world robotic deployments. One problem of data scaling is that it is unlikely to collect all data for each tasks. At the same time, how to balance different data tasks is also challenging. Moreover, performance gains by scaling data is also limited for consistent performance improvements. 

\subsubsection{Use of Pretrained Models}
Many methods aim to enhance generalization by leveraging the generative capabilities of video models. For example, Zhu \textit{et al.} \cite{team2025aether} combine video generation with geometric-aware learning to improve synthetic-to-real generalization across unseen viewpoints and support multiple downstream tasks. Zhen \textit{et al.} \cite{zhen2025tesseract} fine-tune a video generation model on RGB, depth, and normal videos to encode detailed shape, configuration, and temporal dynamics, enabling generalization to unseen scenes, objects, and cross-domain scenarios. The generalization capabilities of large language models, such as video-language models \cite{wang2025founder} and vision-language models \cite{mazzaglia2024genrl}, can be leveraged to enhance world models. By extracting high-level knowledge about the environment, these models facilitate more effective low-level dynamics modeling.

\subsubsection{Instructions Decomposing} 
Another generation issue comes from unseen instructions. To handle this, Zhou \textit{et al.} \cite{zhou2024robodreamer} enhance the ability to unseen instructions by decomposing each spatial relation phrase into a set of compositional components with the pre-trained parser \cite{kitaev2019multilingual} and the rule-based approach. Detailed information can refer to Section \ref{IUF}.

\subsubsection{Invariant Representations}
Generalization can be significantly improved by learning invariant representations to superficial or task-irrelevant changes in the environment. For example, Pang \textit{et al.} \cite{pang2025reviwo} model learns to explicitly decompose visual observations into a view-invariant representation, which is used for the control policy, and a view-dependent representation. This decoupling makes the resulting policy robust to changes in camera viewpoint, a common source of failure in visuomotor control. Similarly, the Martinez \textit{et al.} \cite{martinez2025coral} framework learns a transferable communicative context between two agents, which enables zero-shot adaptation to entirely unseen sparse-reward environments by decoupling the representation learning from the control problem. Wu \textit{et al.} \cite{wu2023pre} disentangle the modeling of context and dynamics by introducing a context encoder, enabling the model to capture shared knowledge for predictions.

\subsubsection{Task-relevant Information Focused} 
Video data often contain irrelevant data to the actions such as background and robot arm, which would limited the generalization ability of the learned world models. To handle this, \cite{zhi20253dflowaction} propose to object-centric world models, which concentrated on object movements via the optical flow predictions that is independent of embodiment. Finn \textit{et al.} \cite{finn2016unsupervised} propose to explicitly model and predict motion that are relatively invariant to the object appearance, enabling long-range predictions and generalize to unseen objects.

\subsubsection{Other Strategies} 
Black \textit{et al.} \cite{black2024zero} use a pretrained image-editing model to generate subgoals from language commands and current observations, enabling low-level controllers to act and generalize to novel objects and scenarios. Self-supervised learning without task-specific rewards that can enhancing generalization abilities into different tasks \cite{sekar2020planning}. 

\subsection{Physics-informed Learning}
Existing world models struggle to generate physically consistent videos because they lack an inherent understanding of physics, often producing unrealistic dynamics and implausible event sequences. Simply scaling up training data or model size is insufficient to capture the underlying physical laws~\cite{kang2025far}. To address this challenge, several approaches have been proposed.
For example, Yang \textit{et al.}~\cite{yang2025vlipp} introduce a two-stage image-to-video generation framework that explicitly incorporates physics through vision- and language-informed physical priors. Team \textit{et al.}~\cite{team2025aether} estimate depth and camera pose directly from videos, facilitating physics-informed learning and enabling world models to infer and predict physically consistent dynamics. Peper \textit{et al.}~\cite{peper2025four} argue that advancing from physics-informed to physics-interpretable world models requires rethinking model design, and propose four guiding principles: organizing latent spaces by physical intent, encoding invariant and equivariant environmental representations, integrating multiple supervision signals, and partitioning generative outputs to improve both scalability and verifiability.

\begin{tcolorbox}[colback=blue!5!white, colframe=blue!70!white, title=Implications for Generalization and Physics-informed World Modeling]
While large-scale training improves the predictive, and generative abilities of world models, handling complex environments requires going beyond simple replication of observations. World models must capture the underlying physical and causal mechanisms of the world, enabling them to generate and predict consistent dynamics across diverse and unseen scenarios.
\end{tcolorbox}

\subsection{Memory} \label{Memo}
Memory mechanisms enable world models to store and retrieve relevant past information, supporting hidden-state disambiguation and long-horizon reasoning. For example, LeCun \textit{et al.}~\cite{lecun2022path} incorporate a memory module that maintains past, current, and predicted world states along with intrinsic costs, allowing retrieval of contextual information for reasoning and training. Huang \textit{et al.}~\cite{huang2025enerverse} propose a sparse contextual memory mechanism that preserves essential prior information throughout the generation process in a non-redundant manner, theoretically enabling the generation of sequences of arbitrary length. Zhou \textit{et al.}~\cite{zhou2025learning} employ a 3D feature-map memory to maintain temporal consistency during sequence generation.

\noindent\textbf{Memory Efficiency.} 
Standard transformer blocks apply multi-head self-attention to all tokens in the input token sequence, resulting in quadratic computation cost. Zhu \textit{et al.} \cite{zhu2025irasim} leverage the memory-efficient spatial-temporal attention mechanism to reduce the computation cost. Liao \textit{et al.} \cite{liao2025genie} randomly sampled parse memory frames from prior video history to augment temporal diversity to improve representational invariance, and use low-frame-rate video sequence for fine-tuning frames.

\subsection{Other Techniques and Challenges}
\subsubsection{Video Fidelity}
To achieve high-fidelity video generation, several methods leverage powerful generative models. For instance, Ko \textit{et al.} \cite{ko2024learning} employ an image diffusion model based on a U-Net with factorized spatiotemporal convolutions as the fundamental building block. Guo \textit{et al.} \cite{guo2025flowdreamer} utilize the pre-trained variational autoencoder from Stable Diffusion \cite{rombach2022high}. Souvcek \textit{et al.} \cite{souvcek2024genhowto} propose to make use of a variety of action and final state prompts. 

\subsubsection{Closed-loop Learning}
Closed-loop learning enables agents to actively refine their internal world models by observing and responding to real-time feedback from the environment. This continuous perception–action cycle grounds learning in physical reality, enhances generalization, and allows adaptive correction. Driess \textit{et al.} \cite{driess2023palm} update observations based on the actions executed, which are then fed into VLMs to enable the robot to correct or reorganize its plan in response to environmental changes and task progress. Bu \textit{et al.} \cite{bu2024closed} design a feedback mechanism that is based on the element-wise discrepancy measure between current and goal state embeddings. Zhi \textit{et al.} \cite{zhi20253dflowaction}, estimate the location of the moving objects, depth prediction, 3D optical flow by input into GPT-4o to verify alignment with given instructions, enabling closed-loop planning. 

\subsubsection{Reasoning}
Reasoning enables a robot to interpret sensory input and translate it into purposeful actions rather than reactive responses. Zhou \textit{et al.} \cite{zhang2025dreamvla}  enhance the reasoning and genrealization ability by incorporating context information and predicting dynamic regions, depth map, semantic knowledge by means of foundation models, e.g., DINOv2 \cite{oquab2024dinov2} and SAM \cite{kirillov2023segment}. Ye \textit{et al.} \cite{Ye2025GigaBrain} introduce an Embodied Chain-of-Thought as an intermediate reasoning representation, enabling more structured and interpretable decision-making in embodied tasks. Ye \textit{et al.} \cite{Ye2025GigaBrain} \cite{zhao2025cot} generates a sub-goal image that represents the robot’s planned state in pixel space, and then conditions its action on both the current observation and the generated subgoal image.

\subsubsection{Sim-to-real Gap}
The substantial gap between synthetic and real-world data limits the transferability and real-world performance of robotic systems. To handle this, Huang \textit{et al.} \cite{huang2025enerverse} propose combining the generative model with 4D Gaussian Splatting, forming a self-reinforcing data loop to reduce the sim-to-real gap.

\subsubsection{3D Robotics World Prediction} 
General-purpose video generation models neglect the substantial gap between their representation space and the three-dimensional, temporally interconnected robotics environment, thereby hindering accurate action policy prediction. For example, Wen \textit{et al.} \cite{wen2024vidman} focuses on 2D image prediction before action generation. To handle this, Huang \textit{et al.} \cite{huang2025enerverse}  propose Free Anchor Views, a multi-view video representation offering flexible, task-adaptive perspectives to address challenges like motion ambiguity and environmental constraints. 

\subsubsection{Fine-grained Robot-object Interaction}
Robots are expected to perform precise manipulation, which requires world models to support fine-grained robot-object interactions. To achieve this, Zhu \textit{et al.} \cite{zhu2025irasim} design a novel frame-level action-conditioning module to achieve precise action-frame alignment. He \textit{et al.} \cite{he2025pre} adopt two different pre-trained video generative models as the base models, and introduce a minimalist yet powerful add-on action-conditioned module that improves frame-level action awareness while maintaining architectural flexibility.

\subsubsection{Multi-agent Operation}
Certain tasks necessitate coordinated operation among multiple robots to achieve successful completion. To this end, Zhang \textit{et al.} \cite{zhang2025combo} factorize the joint actions of different agents as a set of text prompt and leverage composable video diffusion models to learn world dynamics and make predictions. An agent-dependent loss is imposed to let the model focus on the related pixel, where the loss coefficient matrix is based on each agent’s reachable region.

% \subsubsection{Multi-view.}
% Zhang \textit{et al.} \cite{liao2025genie} concentrate on the egocentric nature of perception in dual-arm robotic systems.  Zhang \textit{et al.} \cite{liao2025genie} extend the world model into a multi-view, language-and-image-conditioned generation framework that leverages temporally synchronised inputs from three on-board cameras: a head-mounted view and two wrist-mounted views. It concatenates all views and leverages cross-view attentions. 

\subsubsection{Error Propagation} 
In autoregressive models, subsequent actions are generated based on previous predictions, leading to error propagation over time. To handle this, Cen \textit{et al.} \cite{cen2025worldvla} propose an attention mask strategy that selectively masks prior actions during the generation of the current action. It enables both future imagination and action generation. 

% Cen \textit{et al.} \cite{cen2025worldvla} indicate that generating multiple actions in sequence leads to performance drop in autoregressive models. The primary reason for this is that pretrained multimodal language models have predominantly been exposed to images and text rather than actions, resulting in limited action generalization capabilities. 

\section{Towards Defining Core Components and Capabilities of World Models}  \label{def_wm}

From our survey of current approaches, we summarize some potential key components and capabilities that a world model should possess. Future research may identify additional dimensions necessary for comprehensive world modeling.

%-------------------------------------------------------------------------- 
\begin{figure}
\center
     \includegraphics[width=0.35\textwidth]{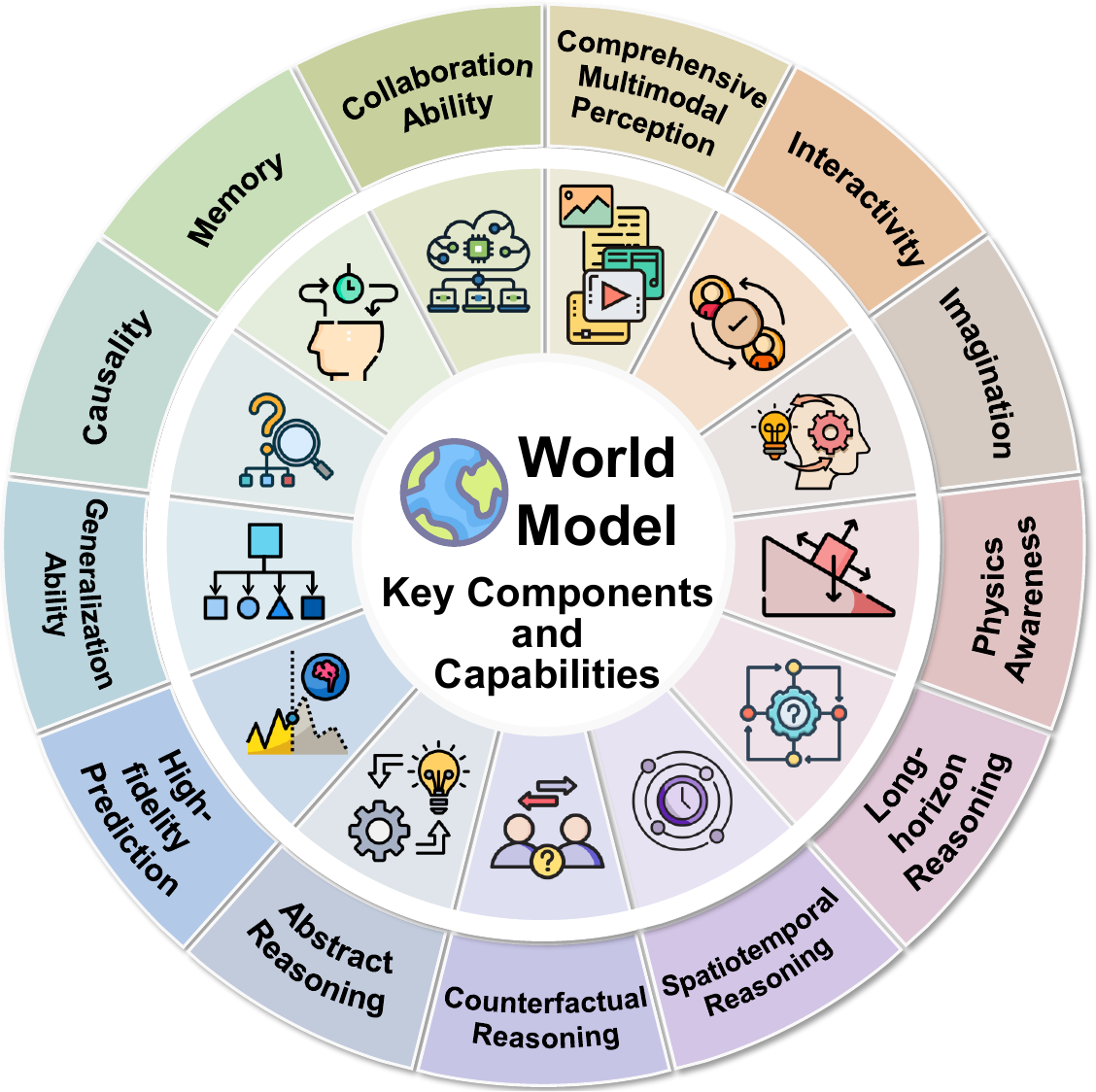}
   \caption{Potential Core Components and Capabilities of World Models.}
  \label{c61}
\end{figure}
%--------------------------------------------------------------------------

\begin{enumerate}
    \item \textbf{Comprehensive Multimodal Perception.} World models should be capable of perceiving and integrating information across all available modalities, such as vision, language, action, touch, force, and proprioception, along with the spatial and temporal structures. By jointly modeling these modalities and dimensions, they can construct a unified and dynamic understanding of the environment that facilitate decision-making and support robot training. 
    \item \textbf{Interactivity.} World models should engage dynamically with their environments, not merely by passively observing or predicting changes, but by modeling how actions influence future states. Such action-conditioned dynamics enable agents to simulate interactions, evaluate potential outcomes, and plan behaviors grounded in causal understanding of the world.
    \item \textbf{Imagination.} Imagination enables world models to simulate and evaluate possible futures, allowing agents to learn, plan, and reason without external interaction.
    \item \textbf{Long-horizon Reasoning.} It enables world models to anticipate distant consequences of actions, plan multi-step behaviors, and optimize long-term outcomes rather than short-term rewards.
    \item \textbf{Spatiotemporal Reasoning.} World models should reason about spatial and temporal relationships among entities to understand and predict dynamic changes in the environment.
    \item \textbf{Counterfactual Reasoning.} This enables world models to imagine alternative futures under different actions, allowing agents to evaluate possible outcomes and select the most effective course of action.
    \item \textbf{Abstract Reasoning.} The world is immensely complex, and world models cannot capture every detail. Therefore, they must extract and represent the underlying principles and basic mechanisms that govern the world’s dynamics.
    \item \textbf{High-fidelity Prediction.} World models should generate accurate and detailed predictions of future states or observations, maintaining spatial, temporal, and physical consistency to ensure reliable simulation and planning.
    \item \textbf{Physics Awareness.} World models should maintain consistency with physical principles, enabling them to generate dynamically plausible predictions that support safe and reliable robotic interaction.
    \item \textbf{Generalization Ability.} To operate effectively in complex real-world settings, world models must generalize beyond their training distributions, adapting to new tasks, objects, and domains.
    \item \textbf{Causality.} World models should understand relationship between actions (causes) and their effects (outcomes) in the world. This causal understanding enables agents to predict how interventions will change future states, distinguish correlation from true influence, and generalize their behavior to unseen situations by reasoning about cause–effect mechanisms rather than memorized patterns.
    \item \textbf{Memory.} It enables world models to store and recall past experiences, ensuring temporal consistency and coherent predictions. In addition, world models should be able to access and integrate external information, thereby supporting richer reasoning, long-term planning, and adaptability—analogous to the role of retrieval-augmented generation (RAG) in language models.
    \item \textbf{Collaboration Ability.} World models should support both inter-agent and intra-agent coordination by reasoning about the behaviors, goals, and intentions of others and managing cooperation among multiple effectors (e.g., multi-arm systems).
    % \item \textbf{Embodied grounding.} Facilitating transfer of learned dynamics from simulation to real-world environments.
\end{enumerate}

\begin{table*}[t]
\centering
\caption{A Summary of Representative Datasets. \underline{\textit{H}}: Hour, \underline{\textit{Manip}}: Manipulation, \underline{\textit{Env.}}: Environments, \underline{\textit{Traj.}}: Trajectories.}
\label{robo_dataset}
\vspace{0.1cm}
\resizebox{1\linewidth}{!}{
\begin{tabular}{r|c|c|c|c|c|cccccc}
\toprule
 \multicolumn{1}{c|}{\multirow{1}{*}{\textbf{Dataset}}} & \multicolumn{1}{c|}{\textbf{Env.}} & \multicolumn{1}{c|}{\textbf{Data Type}}  &\textbf{Dataset Size} &\textbf{Actor} &\textbf{Task/Skill/Content} &\textbf{Notes}\\
\hline
\hline
\rowcolor{blue!5}
Robotic Pushing~\cite{ebert2018robustness} &Laboratory &Video, Gripper Pose &59k  & Robotic Arm  &Push & - \\
RT-1~\cite{brohan2023rt} &Office, Kitchen &Video & $\sim$ 130k  &Robot  &744 Tasks, e.g., Pick, Move & - \\
\rowcolor{blue!5}
Howto100M~\cite{miech2019howto100m} &12 Env., e.g., Home, Garden &Video, Annotation &136M  &Human &23k Tasks, e.g., Cook, Mark &- \\
Ego4D \cite{grauman2022ego4d}& Dailylife, Outdoor, Indoor &Video, Annotation, Audio  &3,670 H  &Human &Diverse Tasks, e.g., flip, lift &  4D, Egocentric\\    
\rowcolor{blue!5}
Ego-Exo4D \cite{grauman2024ego} &123 Natural Scenes & Video, Audio, 3D Point Cloud, Annotation  &221.26 H Videos, 79M+ 3D Body Poses &Human &43 Tasks, 689 Keysteps &Multi-view, 4D,  Egocentric, Exocentric\\
ManiFlow-110k \cite{zhi20253dflowaction} &- & Optical Flow &110k &Robot, Human & Object Movement & 3D \\
\rowcolor{blue!5}    
Something-Something V2 \cite{goyal2017something} &Human-object Interaction & Video &108+K &Human &12 Tasks, e.g., Pour, Drop & - \\
EPIC-KITCHENS \cite{damen2018scaling} &Kitchen &Video, Action Narrations  &432 Sequences/55 H  &Human &323 Tasks, e.g., Cook, Clean &Egocentric  \\  
\rowcolor{blue!5} 
Kinetics-700 \cite{carreira2019short} &Kitchen & Video &650K &Human &700 Tasks, e.g., Pour, Walk &  \\
Bridgedata v2 \cite{walke2023bridgedata} &24 Env., e.g., Kitchens, Tabletops &video &60.1K Traj. &Robot &13 Tasks, e.g., Sweep &-  \\
\rowcolor{blue!5}   
GR-2 \cite{cheang2024gr} &Table-top & Video  &40K Traj., &Human &Diverse tasks/skills, e.g., Pick, Place &-  \\
HM3D \cite{ramakrishnan22021habitat} &Photorealistic Indoor & RGB-D Video &1K Scenes &- &- &3D  \\
\rowcolor{blue!5}  
LanguageTable \cite{lynch2023interactive} &Simulated \& Real-world Tabletop & Video &600k Traj. &Robot, end-effector &Diverse tasks/skills, e.g., Push, Separate &-  \\   
Bridge Data \cite{ebert2021bridge} &Toy \& Real Kitchens, Toy Sinks & Video  &7.2K Trajec. &Robot &71 Tasks, e.g., Flip, Put &-  \\
\rowcolor{blue!5}  
Matterport3D \cite{chang2017matterport3d} &Building-scale Scenes &RGB-D Images   &194K+ &Tripod-mounted Cameras &- & Panoramic-view\\
Multisensory-Universe \cite{hong2024multiply}  &Indoor Scenes &Images, Audio, Tactile, etc &500k &Robot & 67 Robotic  Manip. tasks &3D, Multi-view   \\ 
\rowcolor{blue!5}
Open X-Embodiment \cite{o2024open} &311 Scenes, e.g., Household &Video, Depth, 3D Information  &1M+ Traj. &Robot &527 Skills/160266 Tasks, e.g., Pick, Move &3D  \\ 
DROID \cite{khazatsky2024droid} &52 Buildings/564 Scenes &RGB Video, Depth, Proprioception, Instructions  &76k Traj., 350 H &Robot &86 Tasks &-  \\ 
\rowcolor{blue!5}
ManiSkill \cite{gu2023maniskill2} &Simulation &RGB/RGB-D Video, 3D Point Clouds, Proprioception &4M+ Frames, 2K+ Object Mdels &Robot &20  Manip. Tasks &3D  \\
RoboTurk \cite{mandlekar2019scaling} &Tabletop &RGB \& Depth Images, Robot Sensor Data &2K+ Demonstration Frames, 111.25 H &Robot Arms &Object Search, Tower Creation, Laundry Layout & 3D, Long-horizon Object  Manip. \\ 
\rowcolor{blue!5}     
AgiBot-World-Beta \cite{bu2025agibot} &106 Scenes  &Traj. &1M+ Traj., 2.9K H &Robot Arms &217 Tasks, 87 Skills & Dual-arm, Dexterous hands, Collaboration\\
3D-EIT \cite{zhen20243d} &Simulation &Images, 3D Point Cloud, Annotation &2M &Robot, Human &Robot  Manip., Human-object Interaction  & 3D\\  
\rowcolor{blue!5}
TesserAct \cite{zhen2025tesseract} &Synthetic \& Real Scenes &RGB, Depth, Normal videos &285K &Robot Arms & Robotic  Manip. &4D, Based on \cite{james2020rlbench,brohan2023rt,walke2023bridgedata,goyal2017something} \\
SynGrasp-1B \cite{deng2025graspvla} &Tabletop, Simulation &Image &1B Frames, 10K+ Objects &Robot & Grasp & -  \\
\rowcolor{blue!5}
RH20T-P \cite{chen2024rh20t}  &Tabletop, etc &Video &38k &Robot & 67 Robotic  Manip. Tasks &Primitive-level, Multi-view   \\
RoboNet \cite{dasari2020robonet} &Diverse ENV., e.g., arena &Video, Traj. &15M Video Frame, 62K Traj. &Robot Arm and Gripper & Manip. Tasks, e.g., push and pick-and-place &-  \\
\rowcolor{blue!5}
GenHowTo  \cite{souvcek2024genhowto}  &Kitchens, etc &Image Triplets &200k &Human &Manip., e.g., Cut Tasks &- \\
\toprule  
\end{tabular}}
\end{table*}
%-------------------------------------------------------------------------

\section{Dataset}  \label{datasets}
There are abundant datasets that facilitate robot learning, including general robotic manipulation datasets \cite{khazatsky2024droid,ebert2018robustness,walke2023bridgedata,lynch2023interactive,ebert2021bridge,mandlekar2019scaling,deng2025graspvla,chen2024rh20t}, dual-arm robotic manipulation datasets \cite{bu2025agibot},  human manipulation datasets \cite{miech2019howto100m,goyal2017something,souvcek2024genhowto,carreira2019short,cheang2024gr,zhen20243d}, combinations of robotic \& human manipulation \cite{zhen20243d}, egocentric datasets \cite{grauman2022ego4d,grauman2024ego,damen2018scaling}, 3D \& 4D datasets \cite{zhen2025tesseract,zhen20243d,mandlekar2019scaling, zhi20253dflowaction,grauman2022ego4d,grauman2024ego,ramakrishnan22021habitat,chang2017matterport3d,hong2024multiply,o2024open,gu2023maniskill2},  multi-view datasets \cite{hong2024multiply,grauman2024ego,chen2024rh20t} and panoramic-view datasets \cite{chang2017matterport3d}. A detailed information  can be found in Table. \ref{robo_dataset}.

Recent efforts in world models for robotic manipulation have leveraged large and diverse datasets, often combining multiple sources, to enable generalization across tasks and environments. For example, Yang \textit{et al.} \cite{yang2023learning} construct a large-scale natural dataset combining simulated executions and renderings \cite{ramakrishnan22021habitat,lynch2023interactive}, real robot data \cite{ebert2021bridge}, human activity videos \cite{grauman2022ego4d,damen2018scaling,goyal2017something}, 3D panorama scans \cite{chang2017matterport3d}, and internet text-image data (i.e., LAION-400M \cite{schuhmann2021laion}). Bruce \textit{et al.} \cite{bruce2024genie} combine the RT-1 dataset \cite{brohan2023rt} with real robot grasping data \cite{kalashnikov2018scalable}. Wu \textit{et al.} \cite{wu2024ivideogpt} train the world model based on the combination of the Open X-Embodiment (OXE) dataset \cite{o2024open} and the Something-Something v2 trajectory dataset \cite{goyal2017something}. Bruce \textit{et al.} \cite{cheang2024gr} employ a pretraining and fine-tuning strategy. In the pretraining stage, a combination of human demonstration datasets such as Howto100M \cite{miech2019howto100m}, Ego4D \cite{grauman2022ego4d}, Something-Something V2 \cite{goyal2017something}, EPIC-KITCHENS \cite{damen2018scaling}, Kinetics-700 \cite{carreira2019short}, and robot datasets \cite{brohan2023rt,walke2023bridgedata}. Fine-tuning data includes 105 table-top tasks via teleoperation covering eight skills (e.g., pick, place). Data augmentations are performed to add new objects or change backgrounds by means of a diffusion model \cite{ho2020denoising} and the Segment Anything Model (SAM) \cite{kirillov2023segment}, as well as a video generation model \cite{ma2025latte} to synthesize new videos. Du \textit{et al.} \cite{du2023learning} curate an internet-scale pretraining dataset consisting of 14 million video-text pairs, 60 million image-text pairs \cite{ho2022imagen}, LAION-400M \cite{schuhmann2021laion}, and a smaller real-world robotic dataset \cite{ebert2021bridge}. Huang \textit{et al.} \cite{huang2025enerverse} construct multi-anchor view video datasets using public sources including RT-1 \cite{brohan2023rt}, Taco-Play \cite{rosete2023latent}, ManiSkill \cite{gu2023maniskill2}, BridgeData V2 \cite{walke2023bridgedata}, LanguageTable \cite{lynch2023interactive}, and RoboTurk \cite{mandlekar2019scaling}, augmented with Isaac Sim simulations \cite{mittal2023orbit}.  \cite{zhen2025tesseract} construct a 4D embodied video dataset based on previous datasets \cite{james2020rlbench,brohan2023rt,walke2023bridgedata,goyal2017something} by measuring depth and normal information.

\section{Conclusion and Future research directions}  \label{cfrd}
\subsection{Conclusion}
This survey provides a comprehensive evaluation of current approaches to world modeling, examining their relevance for robotic manipulation, underlying architectures, functionalities, key challenges, and proposed solutions. By synthesizing these findings, we offer insights into the nature of fully realized world models and outline the efforts required to advance the field. Our goal is to provide readers with a solid foundation and guide future research directions in world modeling.

\subsection{Future research directions}

From our survey of current approaches and analysis of the core components and capabilities of world models, it is evident that present models fall short of accurately representing real-world phenomena. The limitations and the set of expected capabilities define promising directions for future research. To provide additional context, we also discuss several significant research directions.

\noindent\textbf{More Diverse Modalities.}
The real world contains diverse forms of information, and no single sensory modality can capture its full complexity. This motivates world models capable of perceiving and integrating multiple modalities, including vision, language, action, touch, force, and proprioception, along with their interactions. Early progress has been made in this direction. For example, Hong \textit{et al.} \cite{hong2024multiply} introduce the Multisensory-Universe dataset, which features interactive scenes enriched with tactile, audio, and temperature signals, generated with the assistance of ChatGPT \cite{achiam2023gpt}.

\noindent\textbf{Hierarchical World Models.}
Hierarchical systems play a critical role in building effective world models, as they allow agents to reason across multiple layers of abstraction. However, designing hierarchical models is inherently challenging: separating low-level and high-level dynamics is difficult, and coordinating interactions across layers adds further complexity. While existing studies primarily emphasize model design \cite{gumbsch2023learning,lecun2022path,wang2025dmwm,xing2025critiques}, their validation in complex real-world environments remains limited.

\noindent\textbf{Causality} is a fundamental principle for understanding and modeling the world, describing how events or factors influence outcomes and enabling reasoning about future consequences. It is the key to the world model and agents to allow them to interact with the world, which is inline with the human cognition. Richens \textit{et al.} \cite{richens2024robust} indicate that learning a causal model is the key to ensure the generalization ability to new domains. Wang \textit{et al.} \cite{wang2022causal,tomar2021model} learn a causal dynamics model by removing unnecessary dependencies for tasks, which however are constrained to specific tasks. Gupta \textit{et al.} \cite{gupta2024essential} argue that conventional theory-driven approaches to causal modeling, such as those in \cite{stuart2010matching, chernozhukov2018double}, are insufficient for world models that aim for generalizable understanding. These methods rely on predefined variables and case-specific theoretical properties. In the real world, sensory inputs are complex, often unstructured, and key theoretical properties, such as identifiability, may not hold.

\noindent\textbf{Resource-Constrained Deployment.}
Current world models, particularly those based on video generation, are computationally intensive and contain hundreds of millions of parameters, which limits their feasibility for real-world robotic deployment and on-device inference. To enable practical applications, designing lightweight and efficient world models has become increasingly important. Quantization and model compression techniques offer promising directions for reducing memory and computational costs, and have been extensively explored in related domains \cite{polino2018model,gholami2022survey,shang2023post,li2021lightweight}, providing both direct solutions and inspiration for future lightweight world model architectures.

\noindent\textbf{Fairness and Security.}
As world models become integral to embodied agents and decision-making systems, ensuring their ethical alignment and fairness is critical. Unlike conventional vision or language models, world models directly influence how autonomous agents perceive, reason, and act within real environments, which amplifies the consequences of biased or unsafe representations. To handle this, emerging research explores bias auditing, fairness-aware training, and safety-constrained learning objectives to prevent harmful behaviors and unintended policy generalization.

Furthermore, deep models are known to be vulnerable to adversarial attacks, which can compromise performance by introducing imperceptible perturbations to inputs \cite{szegedy2013intriguing,zhang2024universal}, modifying model parameters \cite{ren2023dimension,park2022blurs}, or even exploiting hardware-level weaknesses \cite{cojocar2020we,jattke2024zenhammer}.
These vulnerabilities raise serious concerns regarding the security and reliability of world models, especially when deployed in safety-critical domains.
To date, systematic studies on the robustness and security of world models remain limited, underscoring an urgent need for dedicated research into adversarial resilience, trustworthy deployment, and secure model adaptation.
 
\noindent\textbf{Evaluation Protocols.}
Current evaluation practices for world models are fragmented and only loosely aligned with their intended capabilities, often relying on task-specific or proxy metrics and partial human validation \cite{liao2025genie}. There is a pressing need for standardized benchmarks and unified evaluation frameworks that can comprehensively assess world model competence across multiple dimensions, including visual fidelity, policy success, causal consistency, physical plausibility, generalization, and long-horizon reasoning.

\noindent\textbf{Beyond Human Intelligence.}
Insights from human cognition have profoundly influenced the design of robotic and world modeling systems \cite{lecun2022path,gumbsch2023learning,bjorck2025gr00t,wang2025dmwm}. However, the completeness of the world extends beyond human cognition, which is bounded by partial observation, finite memory, limited attention, and inherent heuristic biases. World models are therefore expected to transcend human cognitive bounds, providing a deeper and more systematic understanding of complex environments.

\noindent\textbf{Structured and Abstract World Understanding.}

\begin{quote}
``\textit{Truth is ever to be found in simplicity, and not in the multiplicity and confusion of things}.''
\hfill --- Isaac Newton
\end{quote}

The ultimate goal of world models is to capture underlying regularities and structured abstractions, rather than memorizing every detail of a complex environment. This focus on essential structure is key for enabling world models to generalize across diverse environments.

% \noindent\textbf{Embodied grounding.} World models should bridge the gap between imagination and physical execution by grounding predicted plans into robot-specific control actions and updating internal representations through real-world feedback.

%As we know, the real world is immensely complex, and approximating its full dynamics through a comprehensive world model remains a virtually impossible task, at least for now.

% \clearpage
% \balance
%%%%%%%% REFERENCES
{\small
\bibliographystyle{IEEEtran}
% \nocite{*}

\bibliography{main}

@inproceedings{kang2025far,
  title={How Far Is Video Generation from World Model: A Physical Law Perspective},
  author={Kang, Bingyi and Yue, Yang and Lu, Rui and Lin, Zhijie and Zhao, Yang and Wang, Kaixin and Huang, Gao and Feng, Jiashi},
  booktitle={Forty-second International Conference on Machine Learning},
  year={2025}
}

@article{richens2024robust,
  title={Robust agents learn causal world models},
  author={Richens, Jonathan and Everitt, Tom},
  journal={arXiv preprint arXiv:2402.10877},
  year={2024}
}

@inproceedings{wang2022causal,
  title={Causal Dynamics Learning for Task-Independent State Abstraction},
  author={Wang, Zizhao and Xiao, Xuesu and Xu, Zifan and Zhu, Yuke and Stone, Peter},
  booktitle={International Conference on Machine Learning},
  pages={23151--23180},
  year={2022}
}

@inproceedings{barcellona2025dream,
  title={Dream to Manipulate: Compositional World Models Empowering Robot Imitation Learning with Imagination},
  author={Barcellona, Leonardo and Zadaianchuk, Andrii and Allegro, Davide and Papa, Samuele and Ghidoni, Stefano and Gavves, Efstratios},
  booktitle={The Thirteenth International Conference on Learning Representations},
  year={2025}
}

@article{kerbl20233d,
  title={3D Gaussian splatting for real-time radiance field rendering.},
  author={Kerbl, Bernhard and Kopanas, Georgios and Leimk{\"u}hler, Thomas and Drettakis, George},
  journal={ACM Trans. Graph.},
  volume={42},
  number={4},
  pages={1–14},
  year={2023}
}

@inproceedings{wu2023daydreamer,
  title={Daydreamer: World models for physical robot learning},
  author={Wu, Philipp and Escontrela, Alejandro and Hafner, Danijar and Abbeel, Pieter and Goldberg, Ken},
  booktitle={Conference on robot learning},
  pages={2226--2240},
  year={2023},
  organization={PMLR}
}

@article{lecun2022path,
  title={A path towards autonomous machine intelligence},
  author={LeCun, Yann},
  journal={Open Review},
  volume={62},
  number={1},
  pages={1--62},
  year={2022}
}

@article{ferraro2025focus,
  title={FOCUS: object-centric world models for robotic manipulation},
  author={Ferraro, Stefano and Mazzaglia, Pietro and Verbelen, Tim and Dhoedt, Bart},
  journal={Frontiers in Neurorobotics},
  volume={19},
  pages={1585386},
  year={2025},
  publisher={Frontiers Media SA}
}

@article{liao2025genie,
  title={Genie Envisioner: A Unified World Foundation Platform for Robotic Manipulation},
  author={Liao, Yue and Zhou, Pengfei and Huang, Siyuan and Yang, Donglin and Chen, Shengcong and Jiang, Yuxin and Hu, Yue and Cai, Jingbin and Liu, Si and Luo, Jianlan and others},
  journal={arXiv preprint arXiv:2508.05635},
  year={2025}
}

@article{agarwal2025cosmos,
  title={Cosmos world foundation model platform for physical ai},
  author={Agarwal, Niket and Ali, Arslan and Bala, Maciej and Balaji, Yogesh and Barker, Erik and Cai, Tiffany and Chattopadhyay, Prithvijit and Chen, Yongxin and Cui, Yin and Ding, Yifan and others},
  journal={arXiv preprint arXiv:2501.03575},
  year={2025}
}

@article{he2025pre,
  title={Pre-trained video generative models as world simulators},
  author={He, Haoran and Zhang, Yang and Lin, Liang and Xu, Zhongwen and Pan, Ling},
  journal={arXiv preprint arXiv:2502.07825},
  year={2025}
}

@inproceedings{black2024zero,
  title={Zero-Shot Robotic Manipulation with Pre-Trained Image-Editing Diffusion Models},
  author={Black, Kevin and Nakamoto, Mitsuhiko and Atreya, Pranav and Walke, Homer Rich and Finn, Chelsea and Kumar, Aviral and Levine, Sergey},
  booktitle={The Twelfth International Conference on Learning Representations},
  year={2024}
}

@article{zheng2024open,
  title={Open-sora: Democratizing efficient video production for all},
  author={Zheng, Zangwei and Peng, Xiangyu and Yang, Tianji and Shen, Chenhui and Li, Shenggui and Liu, Hongxin and Zhou, Yukun and Li, Tianyi and You, Yang},
  journal={arXiv preprint arXiv:2412.20404},
  year={2024}
}

@article{wu2024ivideogpt,
  title={ivideogpt: Interactive videogpts are scalable world models},
  author={Wu, Jialong and Yin, Shaofeng and Feng, Ningya and He, Xu and Li, Dong and Hao, Jianye and Long, Mingsheng},
  journal={Advances in Neural Information Processing Systems},
  volume={37},
  pages={68082--68119},
  year={2024}
}

@inproceedings{o2024open,
  title={Open x-embodiment: Robotic learning datasets and rt-x models: Open x-embodiment collaboration 0},
  author={O’Neill, Abby and Rehman, Abdul and Maddukuri, Abhiram and Gupta, Abhishek and Padalkar, Abhishek and Lee, Abraham and Pooley, Acorn and Gupta, Agrim and Mandlekar, Ajay and Jain, Ajinkya and others},
  booktitle={2024 IEEE International Conference on Robotics and Automation (ICRA)},
  pages={6892--6903},
  year={2024},
  organization={IEEE}
}

@article{ha2018world,
  title={World Models},
  author={Ha, David and Schmidhuber, J{\"u}rgen},
  journal={arXiv preprint arXiv:1803.10122},
  year={2018}
}

@article{quevedo2025evaluating,
  title={Evaluating Robot Policies in a World Model},
  author={Quevedo, Julian and Liang, Percy and Yang, Sherry},
  journal={arXiv preprint arXiv:2506.00613},
  year={2025}
}

@inproceedings{todorov2012mujoco,
  title={Mujoco: A physics engine for model-based control},
  author={Todorov, Emanuel and Erez, Tom and Tassa, Yuval},
  booktitle={2012 IEEE/RSJ international conference on intelligent robots and systems},
  pages={5026--5033},
  year={2012},
  organization={IEEE}
}

@inproceedings{erez2015simulation,
  title={Simulation tools for model-based robotics: Comparison of bullet, havok, mujoco, ode and physx},
  author={Erez, Tom and Tassa, Yuval and Todorov, Emanuel},
  booktitle={2015 IEEE international conference on robotics and automation (ICRA)},
  pages={4397--4404},
  year={2015},
  organization={IEEE}
}

@misc{tedrake2019drake,
  title={Drake: Model-based design and verification for robotics},
  author={Tedrake, Russ and others},
  year={2019}
}

@article{sunderhauf2018limits,
  title={The limits and potentials of deep learning for robotics},
  author={S{\"u}nderhauf, Niko and Brock, Oliver and Scheirer, Walter and Hadsell, Raia and Fox, Dieter and Leitner, J{\"u}rgen and Upcroft, Ben and Abbeel, Pieter and Burgard, Wolfram and Milford, Michael and others},
  journal={The International journal of robotics research},
  volume={37},
  number={4-5},
  pages={405--420},
  year={2018},
  publisher={SAGE Publications Sage UK: London, England}
}

@article{afzal2020study,
  title={A study on the challenges of using robotics simulators for testing},
  author={Afzal, Afsoon and Katz, Deborah S and Goues, Claire Le and Timperley, Christopher S},
  journal={arXiv preprint arXiv:2004.07368},
  year={2020}
}

@article{choi2021use,
  title={On the use of simulation in robotics: Opportunities, challenges, and suggestions for moving forward},
  author={Choi, HeeSun and Crump, Cindy and Duriez, Christian and Elmquist, Asher and Hager, Gregory and Han, David and Hearl, Frank and Hodgins, Jessica and Jain, Abhinandan and Leve, Frederick and others},
  journal={Proceedings of the National Academy of Sciences},
  volume={118},
  number={1},
  pages={e1907856118},
  year={2021},
  publisher={National Academy of Sciences}
}

@inproceedings{zhao2020sim,
  title={Sim-to-real transfer in deep reinforcement learning for robotics: a survey},
  author={Zhao, Wenshuai and Queralta, Jorge Pe{\~n}a and Westerlund, Tomi},
  booktitle={2020 IEEE symposium series on computational intelligence (SSCI)},
  pages={737--744},
  year={2020},
  organization={IEEE}
}

@article{dulac2019challenges,
  title={Challenges of real-world reinforcement learning},
  author={Dulac-Arnold, Gabriel and Mankowitz, Daniel and Hester, Todd},
  journal={arXiv preprint arXiv:1904.12901},
  year={2019}
}

@article{hurst2024gpt,
  title={Gpt-4o system card},
  author={Hurst, Aaron and Lerer, Adam and Goucher, Adam P and Perelman, Adam and Ramesh, Aditya and Clark, Aidan and Ostrow, AJ and Welihinda, Akila and Hayes, Alan and Radford, Alec and others},
  journal={arXiv preprint arXiv:2410.21276},
  year={2024}
}

@article{achiam2023gpt,
  title={Gpt-4 technical report},
  author={Achiam, Josh and Adler, Steven and Agarwal, Sandhini and Ahmad, Lama and Akkaya, Ilge and Aleman, Florencia Leoni and Almeida, Diogo and Altenschmidt, Janko and Altman, Sam and Anadkat, Shyamal and others},
  journal={arXiv preprint arXiv:2303.08774},
  year={2023}
}

@inproceedings{gao2024flip,
  title={FLIP: Flow-Centric Generative Planning as General-Purpose Manipulation World Model},
  author={Gao, Chongkai and Zhang, Haozhuo and Xu, Zhixuan and Zhehao, Cai and Shao, Lin},
  booktitle={The Thirteenth International Conference on Learning Representations},
  year={2024}
}

@article{schmidhuber2015learning,
  title={On learning to think: Algorithmic information theory for novel combinations of reinforcement learning controllers and recurrent neural world models},
  author={Schmidhuber, J{\"u}rgen},
  journal={arXiv preprint arXiv:1511.09249},
  year={2015}
}

@article{ali2025humanoid,
  title={Humanoid World Models: Open World Foundation Models for Humanoid Robotics},
  author={Ali, Muhammad Qasim and Sridhar, Aditya and Matiana, Shahbuland and Wong, Alex and Al-Sharman, Mohammad},
  journal={arXiv preprint arXiv:2506.01182},
  year={2025}
}

@article{guo2025flowdreamer,
  title={FlowDreamer: A RGB-D World Model with Flow-based Motion Representations for Robot Manipulation},
  author={Guo, Jun and Ma, Xiaojian and Wang, Yikai and Yang, Min and Liu, Huaping and Li, Qing},
  journal={arXiv preprint arXiv:2505.10075},
  year={2025}
}

@inproceedings{rombach2022high,
  title={High-resolution image synthesis with latent diffusion models},
  author={Rombach, Robin and Blattmann, Andreas and Lorenz, Dominik and Esser, Patrick and Ommer, Bj{\"o}rn},
  booktitle={Proceedings of the IEEE/CVF conference on computer vision and pattern recognition},
  pages={10684--10695},
  year={2022}
}

@inproceedings{hafner2021mastering,
  title={Mastering Atari with Discrete World Models},
  author={Hafner, Danijar and Lillicrap, Timothy P and Norouzi, Mohammad and Ba, Jimmy},
  booktitle={International Conference on Learning Representations},
  year={2021}
}

@article{lu2025gwm,
  title={GWM: Towards Scalable Gaussian World Models for Robotic Manipulation},
  author={Lu, Guanxing and Jia, Baoxiong and Li, Puhao and Chen, Yixin and Wang, Ziwei and Tang, Yansong and Huang, Siyuan},
  journal={arXiv preprint arXiv:2508.17600},
  year={2025}
}

@article{zhi20253dflowaction,
  title={3DFlowAction: Learning Cross-Embodiment Manipulation from 3D Flow World Model},
  author={Zhi, Hongyan and Chen, Peihao and Zhou, Siyuan and Dong, Yubo and Wu, Quanxi and Han, Lei and Tan, Mingkui},
  journal={arXiv preprint arXiv:2506.06199},
  year={2025}
}

@article{cen2025worldvla,
  title={WorldVLA: Towards Autoregressive Action World Model},
  author={Cen, Jun and Yu, Chaohui and Yuan, Hangjie and Jiang, Yuming and Huang, Siteng and Guo, Jiayan and Li, Xin and Song, Yibing and Luo, Hao and Wang, Fan and others},
  journal={arXiv preprint arXiv:2506.21539},
  year={2025}
}

@article{zhang2025dreamvla,
  title={DreamVLA: a vision-language-action model dreamed with comprehensive world knowledge},
  author={Zhang, Wenyao and Liu, Hongsi and Qi, Zekun and Wang, Yunnan and Yu, Xinqiang and Zhang, Jiazhao and Dong, Runpei and He, Jiawei and Wang, He and Zhang, Zhizheng and others},
  journal={arXiv preprint arXiv:2507.04447},
  year={2025}
}

@article{xiang2024pandora,
  title={Pandora: Towards general world model with natural language actions and video states},
  author={Xiang, Jiannan and Liu, Guangyi and Gu, Yi and Gao, Qiyue and Ning, Yuting and Zha, Yuheng and Feng, Zeyu and Tao, Tianhua and Hao, Shibo and Shi, Yemin and others},
  journal={arXiv preprint arXiv:2406.09455},
  year={2024}
}

@article{song2025physical,
  title={Physical Autoregressive Model for Robotic Manipulation without Action Pretraining},
  author={Song, Zijian and Qin, Sihan and Chen, Tianshui and Lin, Liang and Wang, Guangrun},
  journal={arXiv preprint arXiv:2508.09822},
  year={2025}
}

@inproceedings{deng2025autoregressive,
  title={Autoregressive Video Generation without Vector Quantization},
  author={Deng, Haoge and Pan, Ting and Diao, Haiwen and Luo, Zhengxiong and Cui, Yufeng and Lu, Huchuan and Shan, Shiguang and Qi, Yonggang and Wang, Xinlong},
  booktitle={The Thirteenth International Conference on Learning Representations},
  year={2025}
}

@article{javaheripi2023phi,
  title={Phi-2: The surprising power of small language models},
  author={Javaheripi, Mojan and Bubeck, S{\'e}bastien and Abdin, Marah and Aneja, Jyoti and Bubeck, Sebastien and Mendes, Caio C{\'e}sar Teodoro and Chen, Weizhu and Del Giorno, Allie and Eldan, Ronen and Gopi, Sivakanth and others},
  journal={Microsoft Research Blog},
  volume={1},
  number={3},
  pages={3},
  year={2023}
}

@article{bjorck2025gr00t,
  title={Gr00t n1: An open foundation model for generalist humanoid robots},
  author={Bjorck, Johan and Casta{\~n}eda, Fernando and Cherniadev, Nikita and Da, Xingye and Ding, Runyu and Fan, Linxi and Fang, Yu and Fox, Dieter and Hu, Fengyuan and Huang, Spencer and others},
  journal={arXiv preprint arXiv:2503.14734},
  year={2025}
}

@book{kahneman2011thinking,
  title={Thinking, fast and slow},
  author={Kahneman, Daniel},
  year={2011},
  publisher={macmillan}
}

@article{jang2025dreamgen,
  title={DreamGen: Unlocking Generalization in Robot Learning through Video World Models},
  author={Jang, Joel and Ye, Seonghyeon and Lin, Zongyu and Xiang, Jiannan and Bjorck, Johan and Fang, Yu and Hu, Fengyuan and Huang, Spencer and Kundalia, Kaushil and Lin, Yen-Chen and others},
  journal={arXiv preprint arXiv:2505.12705},
  year={2025}
}

@inproceedings{zhou2024robodreamer,
  title={RoboDreamer: Learning Compositional World Models for Robot Imagination},
  author={Zhou, Siyuan and Du, Yilun and Chen, Jiaben and Li, Yandong and Yeung, Dit-Yan and Gan, Chuang},
  booktitle={International Conference on Machine Learning},
  pages={61885--61896},
  year={2024},
  organization={PMLR}
}

@article{azzolini2025cosmos,
  title={Cosmos-reason1: From physical common sense to embodied reasoning},
  author={Azzolini, Alisson and Bai, Junjie and Brandon, Hannah and Cao, Jiaxin and Chattopadhyay, Prithvijit and Chen, Huayu and Chu, Jinju and Cui, Yin and Diamond, Jenna and Ding, Yifan and others},
  journal={arXiv preprint arXiv:2503.15558},
  year={2025}
}

@inproceedings{kitaev2019multilingual,
  title={Multilingual Constituency Parsing with Self-Attention and Pre-Training},
  author={Kitaev, Nikita and Cao, Steven and Klein, Dan},
  booktitle={Proceedings of the 57th Annual Meeting of the Association for Computational Linguistics},
  pages={3499--3505},
  year={2019}
}

@article{li2025manipdreamer,
  title={ManipDreamer: Boosting Robotic Manipulation World Model with Action Tree and Visual Guidance},
  author={Li, Ying and Wei, Xiaobao and Chi, Xiaowei and Li, Yuming and Zhao, Zhongyu and Wang, Hao and Ma, Ningning and Lu, Ming and Zhang, Shanghang},
  journal={arXiv preprint arXiv:2504.16464},
  year={2025}
}

@article{liu2025robotransfer,
  title={RoboTransfer: Geometry-Consistent Video Diffusion for Robotic Visual Policy Transfer},
  author={Liu, Liu and Wang, Xiaofeng and Zhao, Guosheng and Li, Keyu and Qin, Wenkang and Qiu, Jiaxiong and Zhu, Zheng and Huang, Guan and Su, Zhizhong},
  journal={arXiv preprint arXiv:2505.23171},
  year={2025}
}

@inproceedings{wang2025language,
  title={This\&that: Language-gesture controlled video generation for robot planning},
  author={Wang, Boyang and Sridhar, Nikhil and Feng, Chao and Van der Merwe, Mark and Fishman, Adam and Fazeli, Nima and Park, Jeong Joon},
  booktitle={2025 IEEE International Conference on Robotics and Automation (ICRA)},
  pages={12842--12849},
  year={2025},
  organization={IEEE}
}

@inproceedings{grauman2024ego,
  title={Ego-exo4d: Understanding skilled human activity from first-and third-person perspectives},
  author={Grauman, Kristen and Westbury, Andrew and Torresani, Lorenzo and Kitani, Kris and Malik, Jitendra and Afouras, Triantafyllos and Ashutosh, Kumar and Baiyya, Vijay and Bansal, Siddhant and Boote, Bikram and others},
  booktitle={Proceedings of the IEEE/CVF Conference on Computer Vision and Pattern Recognition},
  pages={19383--19400},
  year={2024}
}

@article{yang2023learning,
  title={Learning interactive real-world simulators},
  author={Yang, Mengjiao and Du, Yilun and Ghasemipour, Kamyar and Tompson, Jonathan and Schuurmans, Dale and Abbeel, Pieter},
  journal={arXiv preprint arXiv:2310.06114},
  volume={1},
  number={2},
  pages={6},
  year={2023}
}

@inproceedings{ramakrishnan22021habitat,
  title={Habitat-Matterport 3D Dataset (HM3D): 1000 Large-scale 3D Environments for Embodied AI},
  author={Ramakrishnan, Santhosh Kumar and Gokaslan, Aaron and Wijmans, Erik and Maksymets, Oleksandr and Clegg, Alexander and Turner, John M and Undersander, Eric and Galuba, Wojciech and Westbury, Andrew and Chang, Angel X and others},
  booktitle={Thirty-fifth Conference on Neural Information Processing Systems Datasets and Benchmarks Track (Round 2)},
  year={2021}
}

@article{lynch2023interactive,
  title={Interactive language: Talking to robots in real time},
  author={Lynch, Corey and Wahid, Ayzaan and Tompson, Jonathan and Ding, Tianli and Betker, James and Baruch, Robert and Armstrong, Travis and Florence, Pete},
  journal={IEEE Robotics and Automation Letters},
  year={2023},
  publisher={IEEE}
}

@article{ebert2021bridge,
  title={Bridge data: Boosting generalization of robotic skills with cross-domain datasets},
  author={Ebert, Frederik and Yang, Yanlai and Schmeckpeper, Karl and Bucher, Bernadette and Georgakis, Georgios and Daniilidis, Kostas and Finn, Chelsea and Levine, Sergey},
  journal={arXiv preprint arXiv:2109.13396},
  year={2021}
}

@inproceedings{zitkovich2023rt,
  title={Rt-2: Vision-language-action models transfer web knowledge to robotic control},
  author={Zitkovich, Brianna and Yu, Tianhe and Xu, Sichun and Xu, Peng and Xiao, Ted and Xia, Fei and Wu, Jialin and Wohlhart, Paul and Welker, Stefan and Wahid, Ayzaan and others},
  booktitle={Conference on Robot Learning},
  pages={2165--2183},
  year={2023},
  organization={PMLR}
}

@article{black2024pi_0,
  title={$\pi_0$: A Vision-Language-Action Flow Model for General Robot Control},
  author={Black, Kevin and Brown, Noah and Driess, Danny and Esmail, Adnan and Equi, Michael and Finn, Chelsea and Fusai, Niccolo and Groom, Lachy and Hausman, Karol and Ichter, Brian and others},
  journal={arXiv preprint arXiv:2410.24164},
  year={2024}
}

@article{cheang2025gr,
  title={Gr-3 technical report},
  author={Cheang, Chilam and Chen, Sijin and Cui, Zhongren and Hu, Yingdong and Huang, Liqun and Kong, Tao and Li, Hang and Li, Yifeng and Liu, Yuxiao and Ma, Xiao and others},
  journal={arXiv preprint arXiv:2507.15493},
  year={2025}
}

@article{intelligence2025pi_,
  title={$\pi_{0.5} $: a Vision-Language-Action Model with Open-World Generalization},
  author={Black, Kevin and Brown, Noah and Darpinian, James and Dhabalia, Karan and Driess, Danny and Esmail, Adnan and Equi, Michael and Finn, Chelsea and Fusai, Niccolo and others},
  journal={arXiv preprint arXiv:2504.16054},
  year={2025}
}

@inproceedings{grauman2022ego4d,
  title={Ego4d: Around the world in 3,000 hours of egocentric video},
  author={Grauman, Kristen and Westbury, Andrew and Byrne, Eugene and Chavis, Zachary and Furnari, Antonino and Girdhar, Rohit and Hamburger, Jackson and Jiang, Hao and Liu, Miao and Liu, Xingyu and others},
  booktitle={Proceedings of the IEEE/CVF conference on computer vision and pattern recognition},
  pages={18995--19012},
  year={2022}
}

@inproceedings{bruce2024genie,
  title={Genie: generative interactive environments},
  author={Bruce, Jake and Dennis, Michael and Edwards, Ashley and Parker-Holder, Jack and Shi, Yuge and Hughes, Edward and Lai, Matthew and Mavalankar, Aditi and Steigerwald, Richie and Apps, Chris and others},
  booktitle={Proceedings of the 41st International Conference on Machine Learning},
  pages={4603--4623},
  year={2024}
}

@inproceedings{wu2024unleashing,
  title={Unleashing Large-Scale Video Generative Pre-training for Visual Robot Manipulation},
  author={Wu, Hongtao and Jing, Ya and Cheang, Chilam and Chen, Guangzeng and Xu, Jiafeng and Li, Xinghang and Liu, Minghuan and Li, Hang and Kong, Tao},
  booktitle={The Twelfth International Conference on Learning Representations},
  year={2024}
}

@article{guo2024prediction,
  title={Prediction with action: Visual policy learning via joint denoising process},
  author={Guo, Yanjiang and Hu, Yucheng and Zhang, Jianke and Wang, Yen-Jen and Chen, Xiaoyu and Lu, Chaochao and Chen, Jianyu},
  journal={Advances in Neural Information Processing Systems},
  volume={37},
  pages={112386--112410},
  year={2024}
}

@article{cheang2024gr,
  title={Gr-2: A generative video-language-action model with web-scale knowledge for robot manipulation},
  author={Cheang, Chi-Lam and Chen, Guangzeng and Jing, Ya and Kong, Tao and Li, Hang and Li, Yifeng and Liu, Yuxiao and Wu, Hongtao and Xu, Jiafeng and Yang, Yichu and others},
  journal={arXiv preprint arXiv:2410.06158},
  year={2024}
}

@article{yang2025roboenvision,
  title={RoboEnvision: A Long-Horizon Video Generation Model for Multi-Task Robot Manipulation},
  author={Yang, Liudi and Bai, Yang and Eskandar, George and Shen, Fengyi and Altillawi, Mohammad and Chen, Dong and Majumder, Soumajit and Liu, Ziyuan and Kutyniok, Gitta and Valada, Abhinav},
  journal={arXiv preprint arXiv:2506.22007},
  year={2025}
}

@article{Ye2025GigaBrain,
  title={GigaBrain-0: A World Model-Powered Vision-Language-Action Model},
  author={Angen, Ye},
  journal={arXiv:2510.19430},
  year={2025}
}

@inproceedings{liu2025rdt,
  title={RDT-1B: a Diffusion Foundation Model for Bimanual Manipulation},
  author={Liu, Songming and Wu, Lingxuan and Li, Bangguo and Tan, Hengkai and Chen, Huayu and Wang, Zhengyi and Xu, Ke and Su, Hang and Zhu, Jun},
  booktitle={The Thirteenth International Conference on Learning Representations},
  year={2025}
}

@inproceedings{ko2024learning,
  title={Learning to Act from Actionless Videos through Dense Correspondences},
  author={Ko, Po-Chen and Mao, Jiayuan and Du, Yilun and Sun, Shao-Hua and Tenenbaum, Joshua B},
  booktitle={The Twelfth International Conference on Learning Representations},
  year={2024}
}

@article{dhariwal2021diffusion,
  title={Diffusion models beat gans on image synthesis},
  author={Dhariwal, Prafulla and Nichol, Alexander},
  journal={Advances in neural information processing systems},
  volume={34},
  pages={8780--8794},
  year={2021}
}

@inproceedings{radford2021learning,
  title={Learning transferable visual models from natural language supervision},
  author={Radford, Alec and Kim, Jong Wook and Hallacy, Chris and Ramesh, Aditya and Goh, Gabriel and Agarwal, Sandhini and Sastry, Girish and Askell, Amanda and Mishkin, Pamela and Clark, Jack and others},
  booktitle={International conference on machine learning},
  pages={8748--8763},
  year={2021},
  organization={PmLR}
}

@inproceedings{zhang2025combo,
  title={COMBO: Compositional World Models for Embodied Multi-Agent Cooperation},
  author={Zhang, Hongxin and Wang, Zeyuan and Lyu, Qiushi and Zhang, Zheyuan and Chen, Sunli and Shu, Tianmin and Dariush, Behzad and Lee, Kwonjoon and Du, Yilun and Gan, Chuang},
  booktitle={The Thirteenth International Conference on Learning Representations},
  year={2025}
}

@inproceedings{gao2025adaworld,
  title={AdaWorld: Learning Adaptable World Models with Latent Actions},
  author={Gao, Shenyuan and Zhou, Siyuan and Du, Yilun and Zhang, Jun and Gan, Chuang},
  booktitle={Forty-second International Conference on Machine Learning},
  year={2025}
}

@article{zhen2025tesseract,
  title={TesserAct: learning 4D embodied world models},
  author={Zhen, Haoyu and Sun, Qiao and Zhang, Hongxin and Li, Junyan and Zhou, Siyuan and Du, Yilun and Gan, Chuang},
  journal={arXiv preprint arXiv:2504.20995},
  year={2025}
}

@article{james2020rlbench,
  title={Rlbench: The robot learning benchmark \& learning environment},
  author={James, Stephen and Ma, Zicong and Arrojo, David Rovick and Davison, Andrew J},
  journal={IEEE Robotics and Automation Letters},
  volume={5},
  number={2},
  pages={3019--3026},
  year={2020},
  publisher={IEEE}
}

@article{brooks2024video,
  title={Video generation models as world simulators},
  author={Brooks, Tim and Peebles, Bill and Holmes, Connor and DePue, Will and Guo, Yufei and Jing, Li and Schnurr, David and Taylor, Joe and Luhman, Troy and Luhman, Eric and others},
  journal={OpenAI Blog},
  volume={1},
  number={8},
  pages={1},
  year={2024}
}

@article{wen2024vidman,
  title={Vidman: Exploiting implicit dynamics from video diffusion model for effective robot manipulation},
  author={Wen, Youpeng and Lin, Junfan and Zhu, Yi and Han, Jianhua and Xu, Hang and Zhao, Shen and Liang, Xiaodan},
  journal={Advances in Neural Information Processing Systems},
  volume={37},
  pages={41051--41075},
  year={2024}
}

@inproceedings{he2022masked,
  title={Masked autoencoders are scalable vision learners},
  author={He, Kaiming and Chen, Xinlei and Xie, Saining and Li, Yanghao and Doll{\'a}r, Piotr and Girshick, Ross},
  booktitle={Proceedings of the IEEE/CVF conference on computer vision and pattern recognition},
  pages={16000--16009},
  year={2022}
}

@article{alayrac2022flamingo,
  title={Flamingo: a visual language model for few-shot learning},
  author={Alayrac, Jean-Baptiste and Donahue, Jeff and Luc, Pauline and Miech, Antoine and Barr, Iain and Hasson, Yana and Lenc, Karel and Mensch, Arthur and Millican, Katherine and Reynolds, Malcolm and others},
  journal={Advances in neural information processing systems},
  volume={35},
  pages={23716--23736},
  year={2022}
}

@article{zhang2025up,
  title={UP-VLA: A Unified Understanding and Prediction Model for Embodied Agent},
  author={Zhang, Jianke and Guo, Yanjiang and Hu, Yucheng and Chen, Xiaoyu and Zhu, Xiang and Chen, Jianyu},
  journal={ICML},
  year={2025}
}

@inproceedings{zhao2025cot,
  title={Cot-vla: Visual chain-of-thought reasoning for vision-language-action models},
  author={Zhao, Qingqing and Lu, Yao and Kim, Moo Jin and Fu, Zipeng and Zhang, Zhuoyang and Wu, Yecheng and Li, Zhaoshuo and Ma, Qianli and Han, Song and Finn, Chelsea and others},
  booktitle={Proceedings of the Computer Vision and Pattern Recognition Conference},
  pages={1702--1713},
  year={2025}
}

@inproceedings{tian2025predictive,
  title={Predictive Inverse Dynamics Models are Scalable Learners for Robotic Manipulation},
  author={Tian, Yang and Yang, Sizhe and Zeng, Jia and Wang, Ping and Lin, Dahua and Dong, Hao and Pang, Jiangmiao},
  booktitle={The Thirteenth International Conference on Learning Representations},
  year={2025}
}

@article{zhu2025unified,
  title={Unified world models: Coupling video and action diffusion for pretraining on large robotic datasets},
  author={Zhu, Chuning and Yu, Raymond and Feng, Siyuan and Burchfiel, Benjamin and Shah, Paarth and Gupta, Abhishek},
  journal={arXiv preprint arXiv:2504.02792},
  year={2025}
}

@article{oquab2024dinov2,
  title={DINOv2: Learning Robust Visual Features without Supervision},
  author={Oquab, Maxime and Darcet, Timoth{\'e}e and Moutakanni, Th{\'e}o and Vo, Huy and Szafraniec, Marc and Khalidov, Vasil and Fernandez, Pierre and Haziza, Daniel and Massa, Francisco and El-Nouby, Alaaeldin and others},
  journal={Transactions on Machine Learning Research Journal},
  pages={1--31},
  year={2024}
}

@article{karaev2024cotracker3,
  title={Cotracker3: Simpler and better point tracking by pseudo-labelling real videos},
  author={Karaev, Nikita and Makarov, Iurii and Wang, Jianyuan and Neverova, Natalia and Vedaldi, Andrea and Rupprecht, Christian},
  journal={arXiv preprint arXiv:2410.11831},
  year={2024}
}

@inproceedings{karaev2024cotracker,
  title={Cotracker: It is better to track together},
  author={Karaev, Nikita and Rocco, Ignacio and Graham, Benjamin and Neverova, Natalia and Vedaldi, Andrea and Rupprecht, Christian},
  booktitle={European conference on computer vision},
  pages={18--35},
  year={2024},
  organization={Springer}
}

@inproceedings{yang2024depth,
  title={Depth anything: Unleashing the power of large-scale unlabeled data},
  author={Yang, Lihe and Kang, Bingyi and Huang, Zilong and Xu, Xiaogang and Feng, Jiashi and Zhao, Hengshuang},
  booktitle={Proceedings of the IEEE/CVF conference on computer vision and pattern recognition},
  pages={10371--10381},
  year={2024}
}

@article{khazatsky2024droid,
  title={Droid: A large-scale in-the-wild robot manipulation dataset},
  author={Khazatsky, Alexander and Pertsch, Karl and Nair, Suraj and Balakrishna, Ashwin and Dasari, Sudeep and Karamcheti, Siddharth and Nasiriany, Soroush and Srirama, Mohan Kumar and Chen, Lawrence Yunliang and Ellis, Kirsty and others},
  journal={arXiv preprint arXiv:2403.12945},
  year={2024}
}

@inproceedings{huang2024embodied,
  title={An embodied generalist agent in 3D world},
  author={Huang, Jiangyong and Yong, Silong and Ma, Xiaojian and Linghu, Xiongkun and Li, Puhao and Wang, Yan and Li, Qing and Zhu, Song-Chun and Jia, Baoxiong and Huang, Siyuan},
  booktitle={Proceedings of the 41st International Conference on Machine Learning},
  pages={20413--20451},
  year={2024}
}

@inproceedings{hong2024multiply,
  title={Multiply: A multisensory object-centric embodied large language model in 3d world},
  author={Hong, Yining and Zheng, Zishuo and Chen, Peihao and Wang, Yian and Li, Junyan and Gan, Chuang},
  booktitle={Proceedings of the IEEE/CVF Conference on Computer Vision and Pattern Recognition},
  pages={26406--26416},
  year={2024}
}

@inproceedings{kim2025openvla,
  title={OpenVLA: An Open-Source Vision-Language-Action Model},
  author={Kim, Moo Jin and Pertsch, Karl and Karamcheti, Siddharth and Xiao, Ted and Balakrishna, Ashwin and Nair, Suraj and Rafailov, Rafael and Foster, Ethan P and Sanketi, Pannag R and Vuong, Quan and others},
  booktitle={Conference on Robot Learning},
  pages={2679--2713},
  year={2025},
  organization={PMLR}
}

@article{minsky1975framework,
  title={A framework for representing knowledge},
  author={MINSKY, M},
  journal={The psychology of computer vision},
  pages={211--277},
  year={1975},
  publisher={McGraw-Hill}
}

@inproceedings{zhen20243d,
  title={3D-VLA: a 3D vision-language-action generative world model},
  author={Zhen, Haoyu and Qiu, Xiaowen and Chen, Peihao and Yang, Jincheng and Yan, Xin and Du, Yilun and Hong, Yining and Gan, Chuang},
  booktitle={Proceedings of the 41st International Conference on Machine Learning},
  pages={61229--61245},
  year={2024}
}

@article{hong20233d,
  title={3d-llm: Injecting the 3d world into large language models},
  author={Hong, Yining and Zhen, Haoyu and Chen, Peihao and Zheng, Shuhong and Du, Yilun and Chen, Zhenfang and Gan, Chuang},
  journal={Advances in Neural Information Processing Systems},
  volume={36},
  pages={20482--20494},
  year={2023}
}

@inproceedings{team2025aether,
  title={Aether: Geometric-aware unified world modeling},
  author={Zhu, Haoyi and Wang, Yifan and Zhou, Jianjun and Chang, Wenzheng and Zhou, Yang and Li, Zizun and Chen, Junyi and Shen, Chunhua and Pang, Jiangmiao and others},
  booktitle={ICCV},
  year={2025}
}

@article{kingma2013auto,
  title={Auto-encoding variational bayes},
  author={Kingma, Diederik P and Welling, Max},
  journal={arXiv preprint arXiv:1312.6114},
  year={2013}
}

@inproceedings{sudhakar2024controlling,
  title={Controlling the world by sleight of hand},
  author={Sudhakar, Sruthi and Liu, Ruoshi and Hoorick, Basile Van and Vondrick, Carl and Zemel, Richard},
  booktitle={European Conference on Computer Vision},
  pages={414--430},
  year={2024},
  organization={Springer}
}

@article{du2023video,
  title={Video Language Planning},
  author={Du, Yilun and Yang, Mengjiao and Florence, Pete and Xia, Fei and Wahid, Ayzaan and Ichter, Brian and Sermanet, Pierre and Yu, Tianhe and Abbeel, Pieter and Tenenbaum, Joshua B and others},
  journal={arXiv preprint arXiv:2310.10625},
  year={2023}
}

@article{bu2024closed,
  title={Closed-loop visuomotor control with generative expectation for robotic manipulation},
  author={Bu, Qingwen and Zeng, Jia and Chen, Li and Yang, Yanchao and Zhou, Guyue and Yan, Junchi and Luo, Ping and Cui, Heming and Ma, Yi and Li, Hongyang},
  journal={Advances in Neural Information Processing Systems},
  volume={37},
  pages={139002--139029},
  year={2024}
}

@inproceedings{souvcek2024genhowto,
  title={Genhowto: Learning to generate actions and state transformations from instructional videos},
  author={Sou{\v{c}}ek, Tom{\'a}{\v{s}} and Damen, Dima and Wray, Michael and Laptev, Ivan and Sivic, Josef},
  booktitle={Proceedings of the IEEE/CVF Conference on Computer Vision and Pattern Recognition},
  pages={6561--6571},
  year={2024}
}

@article{ho2022imagen,
  title={Imagen video: High definition video generation with diffusion models},
  author={Ho, Jonathan and Chan, William and Saharia, Chitwan and Whang, Jay and Gao, Ruiqi and Gritsenko, Alexey and Kingma, Diederik P and Poole, Ben and Norouzi, Mohammad and Fleet, David J and others},
  journal={arXiv preprint arXiv:2210.02303},
  year={2022}
}

@inproceedings{driess2023palm,
  title={PaLM-E: an embodied multimodal language model},
  author={Driess, Danny and Xia, Fei and Sajjadi, Mehdi SM and Lynch, Corey and Chowdhery, Aakanksha and Ichter, Brian and Wahid, Ayzaan and Tompson, Jonathan and Vuong, Quan and Yu, Tianhe and others},
  booktitle={Proceedings of the 40th International Conference on Machine Learning},
  pages={8469--8488},
  year={2023}
}

@article{ahn2022can,
  title={Do as i can, not as i say: Grounding language in robotic affordances},
  author={Ahn, Michael and Brohan, Anthony and Brown, Noah and Chebotar, Yevgen and Cortes, Omar and David, Byron and Finn, Chelsea and Fu, Chuyuan and Gopalakrishnan, Keerthana and Hausman, Karol and others},
  journal={arXiv preprint arXiv:2204.01691},
  year={2022}
}

@article{du2023learning,
  title={Learning universal policies via text-guided video generation},
  author={Du, Yilun and Yang, Sherry and Dai, Bo and Dai, Hanjun and Nachum, Ofir and Tenenbaum, Josh and Schuurmans, Dale and Abbeel, Pieter},
  journal={Advances in neural information processing systems},
  volume={36},
  pages={9156--9172},
  year={2023}
}

@misc{team2023internlm,
  title={Internlm: A multilingual language model with progressively enhanced capabilities},
  author={Team, InternLM},
  year={2023}
}

@article{chen2025egoagent,
  title={EgoAgent: A Joint Predictive Agent Model in Egocentric Worlds},
  author={Chen, Lu and Wang, Yizhou and Tang, Shixiang and Ma, Qianhong and He, Tong and Ouyang, Wanli and Zhou, Xiaowei and Bao, Hujun and Peng, Sida},
  journal={arXiv preprint arXiv:2502.05857},
  year={2025}
}

@article{baker2022video,
  title={Video pretraining (vpt): Learning to act by watching unlabeled online videos},
  author={Baker, Bowen and Akkaya, Ilge and Zhokov, Peter and Huizinga, Joost and Tang, Jie and Ecoffet, Adrien and Houghton, Brandon and Sampedro, Raul and Clune, Jeff},
  journal={Advances in Neural Information Processing Systems},
  volume={35},
  pages={24639--24654},
  year={2022}
}

@article{zhou2025learning,
  title={Learning 3D Persistent Embodied World Models},
  author={Zhou, Siyuan and Du, Yilun and Yang, Yuncong and Han, Lei and Chen, Peihao and Yeung, Dit-Yan and Gan, Chuang},
  journal={arXiv preprint arXiv:2505.05495},
  year={2025}
}

@inproceedings{ye2025latent,
  title={Latent Action Pretraining from Videos},
  author={Ye, Seonghyeon and Jang, Joel and Jeon, Byeongguk and Joo, Se June and Yang, Jianwei and Peng, Baolin and Mandlekar, Ajay and Tan, Reuben and Chao, Yu-Wei and Lin, Bill Yuchen and others},
  booktitle={The Thirteenth International Conference on Learning Representations},
  year={2025}
}

@inproceedings{ren2025videoworld,
  title={Videoworld: Exploring knowledge learning from unlabeled videos},
  author={Ren, Zhongwei and Wei, Yunchao and Guo, Xun and Zhao, Yao and Kang, Bingyi and Feng, Jiashi and Jin, Xiaojie},
  booktitle={Proceedings of the Computer Vision and Pattern Recognition Conference},
  pages={29029--29039},
  year={2025}
}

@inproceedings{mentzer2024finite,
  title={Finite Scalar Quantization: VQ-VAE Made Simple},
  author={Mentzer, Fabian and Minnen, David and Agustsson, Eirikur and Tschannen, Michael},
  booktitle={The Twelfth International Conference on Learning Representations},
  year={2025}
}

@inproceedings{zheng2025universal,
  title={Universal actions for enhanced embodied foundation models},
  author={Zheng, Jinliang and Li, Jianxiong and Liu, Dongxiu and Zheng, Yinan and Wang, Zhihao and Ou, Zhonghong and Liu, Yu and Liu, Jingjing and Zhang, Ya-Qin and Zhan, Xianyuan},
  booktitle={Proceedings of the Computer Vision and Pattern Recognition Conference},
  pages={22508--22519},
  year={2025}
}

@article{huang2025enerverse,
  title={Enerverse: Envisioning embodied future space for robotics manipulation},
  author={Huang, Siyuan and Chen, Liliang and Zhou, Pengfei and Chen, Shengcong and Jiang, Zhengkai and Hu, Yue and Liao, Yue and Gao, Peng and Li, Hongsheng and Yao, Maoqing and others},
  journal={arXiv preprint arXiv:2501.01895},
  year={2025}
}

@inproceedings{villar2025playslot,
  title={PlaySlot: Learning Inverse Latent Dynamics for Controllable Object-Centric Video Prediction and Planning},
  author={Villar-Corrales, Angel and Behnke, Sven},
  booktitle={Forty-second International Conference on Machine Learning},
  year={2025}
}

@article{li2025worldeval,
  title={WorldEval: World Model as Real-World Robot Policies Evaluator},
  author={Li, Yaxuan and Zhu, Yichen and Wen, Junjie and Shen, Chaomin and Xu, Yi},
  journal={arXiv preprint arXiv:2505.19017},
  year={2025}
}

@article{deng2025graspvla,
  title={Graspvla: a grasping foundation model pre-trained on billion-scale synthetic action data},
  author={Deng, Shengliang and Yan, Mi and Wei, Songlin and Ma, Haixin and Yang, Yuxin and Chen, Jiayi and Zhang, Zhiqi and Yang, Taoyu and Zhang, Xuheng and Cui, Heming and others},
  journal={arXiv preprint arXiv:2505.03233},
  year={2025}
}

@article{wang2025learning,
  title={Learning Real-World Action-Video Dynamics with Heterogeneous Masked Autoregression},
  author={Wang, Lirui and Zhao, Kevin and Liu, Chaoqi and Chen, Xinlei},
  journal={arXiv preprint arXiv:2502.04296},
  year={2025}
}

@inproceedings{doshi2025scaling,
  title={Scaling Cross-Embodied Learning: One Policy for Manipulation, Navigation, Locomotion and Aviation},
  author={Doshi, Ria and Walke, Homer Rich and Mees, Oier and Dasari, Sudeep and Levine, Sergey},
  booktitle={Conference on Robot Learning},
  pages={496--512},
  year={2025},
  organization={PMLR}
}

@article{team2024octo,
  title={Octo: An open-source generalist robot policy},
  author={Team, Octo Model and Ghosh, Dibya and Walke, Homer and Pertsch, Karl and Black, Kevin and Mees, Oier and Dasari, Sudeep and Hejna, Joey and Kreiman, Tobias and Xu, Charles and others},
  journal={arXiv preprint arXiv:2405.12213},
  year={2024}
}

@article{wang2024scaling,
  title={Scaling proprioceptive-visual learning with heterogeneous pre-trained transformers},
  author={Wang, Lirui and Chen, Xinlei and Zhao, Jialiang and He, Kaiming},
  journal={Advances in neural information processing systems},
  volume={37},
  pages={124420--124450},
  year={2024}
}

@inproceedings{zhu2025irasim,
  title={Irasim: Learning interactive real-robot action simulators},
  author={Zhu, Fangqi and Wu, Hongtao and Guo, Song and Liu, Yuxiao and Cheang, Chilam and Kong, Tao},
  booktitle={ICCV},
  year={2025}
}

@article{ho2022video,
  title={Video diffusion models},
  author={Ho, Jonathan and Salimans, Tim and Gritsenko, Alexey and Chan, William and Norouzi, Mohammad and Fleet, David J},
  journal={Advances in neural information processing systems},
  volume={35},
  pages={8633--8646},
  year={2022}
}

@article{zhao2024vlmpc,
  title={Vlmpc: Vision-language model predictive control for robotic manipulation},
  author={Zhao, Wentao and Chen, Jiaming and Meng, Ziyu and Mao, Donghui and Song, Ran and Zhang, Wei},
  journal={arXiv preprint arXiv:2407.09829},
  year={2024}
}

@inproceedings{ebert2018robustness,
  title={Robustness via retrying: Closed-loop robotic manipulation with self-supervised learning},
  author={Ebert, Frederik and Dasari, Sudeep and Lee, Alex X and Levine, Sergey and Finn, Chelsea},
  booktitle={Conference on robot learning},
  pages={983--993},
  year={2018},
  organization={PMLR}
}

@inproceedings{nair2022learning,
  title={Learning language-conditioned robot behavior from offline data and crowd-sourced annotation},
  author={Nair, Suraj and Mitchell, Eric and Chen, Kevin and Savarese, Silvio and Finn, Chelsea and others},
  booktitle={Conference on Robot Learning},
  pages={1303--1315},
  year={2022},
  organization={PMLR}
}

@article{hu2023look,
  title={Look before you leap: Unveiling the power of gpt-4v in robotic vision-language planning},
  author={Hu, Yingdong and Lin, Fanqi and Zhang, Tong and Yi, Li and Gao, Yang},
  journal={arXiv preprint arXiv:2311.17842},
  year={2023}
}

@inproceedings{dasari2020robonet,
  title={RoboNet: Large-Scale Multi-Robot Learning},
  author={Dasari, Sudeep and Ebert, Frederik and Tian, Stephen and Nair, Suraj and Bucher, Bernadette and Schmeckpeper, Karl and Singh, Siddharth and Levine, Sergey and Finn, Chelsea},
  booktitle={Conference on Robot Learning},
  pages={885--897},
  year={2020}
}

@inproceedings{dosovitskiy2017carla,
  title={CARLA: An open urban driving simulator},
  author={Dosovitskiy, Alexey and Ros, German and Codevilla, Felipe and Lopez, Antonio and Koltun, Vladlen},
  booktitle={Conference on robot learning},
  pages={1--16},
  year={2017}
}

@inproceedings{deitke2020robothor,
  title={Robothor: An open simulation-to-real embodied ai platform},
  author={Deitke, Matt and Han, Winson and Herrasti, Alvaro and Kembhavi, Aniruddha and Kolve, Eric and Mottaghi, Roozbeh and Salvador, Jordi and Schwenk, Dustin and VanderBilt, Eli and Wallingford, Matthew and others},
  booktitle={Proceedings of the IEEE/CVF conference on computer vision and pattern recognition},
  pages={3164--3174},
  year={2020}
}

@inproceedings{chen2024rh20t,
  title={RH20T-P: A Primitive-Level Robotic Manipulation Dataset Towards Composable Generalization Agents in Real-world Scenarios},
  author={Chen, Zeren and Shi, Zhelun and Lu, Xiaoya and He, Lehan and Qian, Sucheng and Yin, Zhenfei and Ouyang, Wanli and Shao, Jing and Qiao, Yu and Lu, Cewu and others},
  booktitle={NeurIPS 2024 Workshop on Open-World Agents},
  year={2024}
}

@article{wang2025dmwm,
  title={DMWM: Dual-Mind World Model with Long-Term Imagination},
  author={Wang, Lingyi and Shelim, Rashed and Saad, Walid and Ramakrishnan, Naren},
  journal={arXiv preprint arXiv:2502.07591},
  year={2025}
}

@article{hafner2023mastering,
  title={Mastering diverse domains through world models},
  author={Hafner, Danijar and Pasukonis, Jurgis and Ba, Jimmy and Lillicrap, Timothy},
  journal={arXiv preprint arXiv:2301.04104},
  year={2023}
}

@article{yang2025vlipp,
  title={VLIPP: Towards Physically Plausible Video Generation with Vision and Language Informed Physical Prior},
  author={Yang, Xindi and Li, Baolu and Zhang, Yiming and Yin, Zhenfei and Bai, Lei and Ma, Liqian and Wang, Zhiyong and Cai, Jianfei and Wong, Tien-Tsin and Lu, Huchuan and others},
  journal={arXiv preprint arXiv:2503.23368},
  year={2025}
}

@article{escontrela2023video,
  title={Video prediction models as rewards for reinforcement learning},
  author={Escontrela, Alejandro and Adeniji, Ademi and Yan, Wilson and Jain, Ajay and Peng, Xue Bin and Goldberg, Ken and Lee, Youngwoon and Hafner, Danijar and Abbeel, Pieter},
  journal={Advances in Neural Information Processing Systems},
  volume={36},
  pages={68760--68783},
  year={2023}
}

@article{liang2025video,
  title={Video Generators are Robot Policies},
  author={Liang, Junbang and Tokmakov, Pavel and Liu, Ruoshi and Sudhakar, Sruthi and Shah, Paarth and Ambrus, Rares and Vondrick, Carl},
  journal={arXiv preprint arXiv:2508.00795},
  year={2025}
}

@incollection{gholami2022survey,
  title={A survey of quantization methods for efficient neural network inference},
  author={Gholami, Amir and Kim, Sehoon and Dong, Zhen and Yao, Zhewei and Mahoney, Michael W and Keutzer, Kurt},
  booktitle={Low-power computer vision},
  pages={291--326},
  year={2022},
  publisher={Chapman and Hall/CRC}
}

@inproceedings{richens2025general,
  title={General agents need world models},
  author={Richens, Jonathan and Everitt, Tom and Abel, David},
  booktitle={Forty-second International Conference on Machine Learning},
  year={2025}
}

@inproceedings{shang2023post,
  title={Post-training quantization on diffusion models},
  author={Shang, Yuzhang and Yuan, Zhihang and Xie, Bin and Wu, Bingzhe and Yan, Yan},
  booktitle={Proceedings of the IEEE/CVF conference on computer vision and pattern recognition},
  pages={1972--1981},
  year={2023}
}

@inproceedings{cojocar2020we,
  title={Are we susceptible to rowhammer? an end-to-end methodology for cloud providers},
  author={Cojocar, Lucian and Kim, Jeremie and Patel, Minesh and Tsai, Lillian and Saroiu, Stefan and Wolman, Alec and Mutlu, Onur},
  booktitle={SP},
  pages={712--728},
  year={2020}
}

@inproceedings{jattke2024zenhammer,
  title={$\{$ZenHammer$\}$: Rowhammer Attacks on $\{$AMD$\}$ Zen-based Platforms},
  author={Jattke, Patrick and Wipfli, Max and Solt, Flavien and Marazzi, Michele and B{\"o}lcskei, Matej and Razavi, Kaveh},
  booktitle={USENIX Security},
  pages={1615--1633},
  year={2024}
}

@inproceedings{ren2023dimension,
  title={Dimension-independent certified neural network watermarks via mollifier smoothing},
  author={Ren, Jiaxiang and Zhou, Yang and Jin, Jiayin and Lyu, Lingjuan and Yan, Da},
  booktitle={ICML},
  pages={28976--29008},
  year={2023},
  organization={PMLR}
}

@inproceedings{park2022blurs,
  title={Blurs behave like ensembles: Spatial smoothings to improve accuracy, uncertainty, and robustness},
  author={Park, Namuk and Kim, Songkuk},
  booktitle={ICML},
  pages={17390--17419},
  year={2022}
}

@article{szegedy2013intriguing,
  title={Intriguing properties of neural networks},
  author={Szegedy, Christian and Zaremba, Wojciech and Sutskever, Ilya and Bruna, Joan and Erhan, Dumitru and Goodfellow, Ian and Fergus, Rob},
  journal={arXiv preprint arXiv:1312.6199},
  year={2013}
}

@inproceedings{zhang2024universal,
  title={Universal adversarial perturbations for vision-language pre-trained models},
  author={Zhang, Peng-Fei and Huang, Zi and Bai, Guangdong},
  booktitle={Proceedings of the 47th International ACM SIGIR Conference on Research and Development in Information Retrieval},
  pages={862--871},
  year={2024}
}

@inproceedings{li2021lightweight,
  title={Lightweight self-attentive sequential recommendation},
  author={Li, Yang and Chen, Tong and Zhang, Peng-Fei and Yin, Hongzhi},
  booktitle={Proceedings of the 30th ACM international conference on information \& knowledge management},
  pages={967--977},
  year={2021}
}

@inproceedings{polino2018model,
  title={Model compression via distillation and quantization},
  author={Polino, Antonio and Pascanu, Razvan and Alistarh, Dan},
  booktitle={International Conference on Learning Representations},
  year={2018}
}

@article{song2025hume,
  title={Hume: Introducing System-2 Thinking in Visual-Language-Action Model},
  author={Song, Haoming and Qu, Delin and Yao, Yuanqi and Chen, Qizhi and Lv, Qi and Tang, Yiwen and Shi, Modi and Ren, Guanghui and Yao, Maoqing and Zhao, Bin and others},
  journal={arXiv preprint arXiv:2505.21432},
  year={2025}
}

@article{team2023gemini,
  title={Gemini: a family of highly capable multimodal models},
  author={Team, Gemini and Anil, Rohan and Borgeaud, Sebastian and Alayrac, Jean-Baptiste and Yu, Jiahui and Soricut, Radu and Schalkwyk, Johan and Dai, Andrew M and Hauth, Anja and Millican, Katie and others},
  journal={arXiv preprint arXiv:2312.11805},
  year={2023}
}

@article{team2025gemini,
  title={Gemini robotics: Bringing ai into the physical world},
  author={Team, Gemini Robotics and Abeyruwan, Saminda and Ainslie, Joshua and Alayrac, Jean-Baptiste and Arenas, Montserrat Gonzalez and Armstrong, Travis and Balakrishna, Ashwin and Baruch, Robert and Bauza, Maria and Blokzijl, Michiel and others},
  journal={arXiv preprint arXiv:2503.20020},
  year={2025}
}

@inproceedings{wang2025founder,
  title={Founder: Grounding foundation models in world models for open-ended embodied decision making},
  author={Wang, Yucen and Yu, Rui and Wan, Shenghua and Gan, Le and Zhan, De-Chuan},
  booktitle={Forty-second International Conference on Machine Learning},
  year={2025}
}

@article{mazzaglia2024genrl,
  title={GenRL: Multimodal-foundation world models for generalization in embodied agents},
  author={Mazzaglia, Pietro and Verbelen, Tim and Dhoedt, Bart and Courville, Aaron and Rajeswar, Sai},
  journal={Advances in neural information processing systems},
  volume={37},
  pages={27529--27555},
  year={2024}
}

@inproceedings{sekar2020planning,
  title={Planning to explore via self-supervised world models},
  author={Sekar, Ramanan and Rybkin, Oleh and Daniilidis, Kostas and Abbeel, Pieter and Hafner, Danijar and Pathak, Deepak},
  booktitle={International conference on machine learning},
  pages={8583--8592},
  year={2020}
}

@article{gupta2024essential,
  title={The essential role of causality in foundation world models for embodied AI},
  author={Gupta, Tarun and Gong, Wenbo and Ma, Chao and Pawlowski, Nick and Hilmkil, Agrin and Scetbon, Meyer and Rigter, Marc and Famoti, Ade and Llorens, Ashley Juan and Gao, Jianfeng and others},
  journal={arXiv preprint arXiv:2402.06665},
  year={2024}
}

@article{tomar2021model,
  title={Model-invariant state abstractions for model-based reinforcement learning},
  author={Tomar, Manan and Zhang, Amy and Calandra, Roberto and Taylor, Matthew E and Pineau, Joelle},
  journal={arXiv preprint arXiv:2102.09850},
  year={2021}
}

@article{stuart2010matching,
  title={Matching methods for causal inference: A review and a look forward},
  author={Stuart, Elizabeth A},
  journal={Statistical science: a review journal of the Institute of Mathematical Statistics},
  volume={25},
  number={1},
  pages={1},
  year={2010}
}

@article{chernozhukov2018double,
  title={Double/debiased machine learning for treatment and structural parameters},
  author={Chernozhukov, Victor and Chetverikov, Denis and Demirer, Mert and Duflo, Esther and Hansen, Christian and Newey, Whitney and Robins, James},
  journal={The Econometrics Journal},
  pages={C1--C68},
  year={2018},
  publisher={JSTOR}
}

@article{peper2025four,
  title={Four Principles for Physically Interpretable World Models},
  author={Peper, Jordan and Mao, Zhenjiang and Geng, Yuang and Pan, Siyuan and Ruchkin, Ivan},
  journal={arXiv preprint arXiv:2503.02143},
  year={2025}
}

@article{xing2025critiques,
  title={Critiques of World Models},
  author={Xing, Eric and Deng, Mingkai and Hou, Jinyu and Hu, Zhiting},
  journal={arXiv preprint arXiv:2507.05169},
  year={2025}
}

@inproceedings{rigter2025avid,
  title={AVID: Adapting Video Diffusion Models to World Models},
  author={Rigter, Marc and Gupta, Tarun and Hilmkil, Agrin and Ma, Chao},
  booktitle={Reinforcement Learning Conference},
  year={2025}
}

@inproceedings{finn2016unsupervised,
  title={Unsupervised learning for physical interaction through video prediction},
  author={Finn, Chelsea and Goodfellow, Ian and Levine, Sergey},
  booktitle={Proceedings of the 30th International Conference on Neural Information Processing Systems},
  pages={64--72},
  year={2016}
}

@article{bu2025agibot,
  title={Agibot world colosseo: A large-scale manipulation platform for scalable and intelligent embodied systems},
  author={Bu, Qingwen and Cai, Jisong and Chen, Li and Cui, Xiuqi and Ding, Yan and Feng, Siyuan and Gao, Shenyuan and He, Xindong and Hu, Xuan and Huang, Xu and others},
  journal={arXiv preprint arXiv:2503.06669},
  year={2025}
}

@article{wan2025wan,
  title={Wan: Open and advanced large-scale video generative models},
  author={Wan, Team and Wang, Ang and Ai, Baole and Wen, Bin and Mao, Chaojie and Xie, Chen-Wei and Chen, Di and Yu, Feiwu and Zhao, Haiming and Yang, Jianxiao and others},
  journal={arXiv preprint arXiv:2503.20314},
  year={2025}
}

@inproceedings{walke2023bridgedata,
  title={Bridgedata v2: A dataset for robot learning at scale},
  author={Walke, Homer Rich and Black, Kevin and Zhao, Tony Z and Vuong, Quan and Zheng, Chongyi and Hansen-Estruch, Philippe and He, Andre Wang and Myers, Vivek and Kim, Moo Jin and Du, Max and others},
  booktitle={Conference on Robot Learning},
  pages={1723--1736},
  year={2023}
}

@inproceedings{rosete2023latent,
  title={Latent plans for task-agnostic offline reinforcement learning},
  author={Rosete-Beas, Erick and Mees, Oier and Kalweit, Gabriel and Boedecker, Joschka and Burgard, Wolfram},
  booktitle={Conference on Robot Learning},
  pages={1838--1849},
  year={2023}
}

@inproceedings{gu2023maniskill2,
  title={ManiSkill2: A Unified Benchmark for Generalizable Manipulation Skills},
  author={Gu, Jiayuan and Xiang, Fanbo and Li, Xuanlin and Ling, Zhan and Liu, Xiqiang and Mu, Tongzhou and Tang, Yihe and Tao, Stone and Wei, Xinyue and Yao, Yunchao and others},
  booktitle={The Eleventh International Conference on Learning Representations},
  year={2023},
}

@inproceedings{mandlekar2019scaling,
  title={Scaling robot supervision to hundreds of hours with roboturk: Robotic manipulation dataset through human reasoning and dexterity},
  author={Mandlekar, Ajay and Booher, Jonathan and Spero, Max and Tung, Albert and Gupta, Anchit and Zhu, Yuke and Garg, Animesh and Savarese, Silvio and Fei-Fei, Li},
  booktitle={2019 IEEE/RSJ International Conference on Intelligent Robots and Systems (IROS)},
  pages={1048--1055},
  year={2019},
  organization={IEEE}
}

@article{mittal2023orbit,
  title={Orbit: A unified simulation framework for interactive robot learning environments},
  author={Mittal, Mayank and Yu, Calvin and Yu, Qinxi and Liu, Jingzhou and Rudin, Nikita and Hoeller, David and Yuan, Jia Lin and Singh, Ritvik and Guo, Yunrong and Mazhar, Hammad and others},
  journal={IEEE Robotics and Automation Letters},
  volume={8},
  number={6},
  pages={3740--3747},
  year={2023},
  publisher={IEEE}
}

@article{brohan2023rt,
  title={RT-1: Robotics Transformer for Real-World Control at Scale},
  author={Brohan, Anthony and Brown, Noah and Carbajal, Justice and Chebotar, Yevgen and Dabis, Joseph and Finn, Chelsea and Gopalakrishnan, Keerthana and Hausman, Karol and Herzog, Alexander and Hsu, Jasmine and others},
  journal={Robotics: Science and Systems},
  year={2023},
  publisher={Robotics: Science and Systems Foundation}
}

@inproceedings{kalashnikov2018scalable,
  title={Scalable deep reinforcement learning for vision-based robotic manipulation},
  author={Kalashnikov, Dmitry and Irpan, Alex and Pastor, Peter and Ibarz, Julian and Herzog, Alexander and Jang, Eric and Quillen, Deirdre and Holly, Ethan and Kalakrishnan, Mrinal and Vanhoucke, Vincent and others},
  booktitle={Conference on robot learning},
  pages={651--673},
  year={2018}
}

@inproceedings{miech2019howto100m,
  title={Howto100m: Learning a text-video embedding by watching hundred million narrated video clips},
  author={Miech, Antoine and Zhukov, Dimitri and Alayrac, Jean-Baptiste and Tapaswi, Makarand and Laptev, Ivan and Sivic, Josef},
  booktitle={Proceedings of the IEEE/CVF international conference on computer vision},
  pages={2630--2640},
  year={2019}
}

@inproceedings{kirillov2023segment,
  title={Segment anything},
  author={Kirillov, Alexander and Mintun, Eric and Ravi, Nikhila and Mao, Hanzi and Rolland, Chloe and Gustafson, Laura and Xiao, Tete and Whitehead, Spencer and Berg, Alexander C and Lo, Wan-Yen and others},
  booktitle={Proceedings of the IEEE/CVF international conference on computer vision},
  pages={4015--4026},
  year={2023}
}

@article{ho2020denoising,
  title={Denoising diffusion probabilistic models},
  author={Ho, Jonathan and Jain, Ajay and Abbeel, Pieter},
  journal={Advances in neural information processing systems},
  volume={33},
  pages={6840--6851},
  year={2020}
}

@article{ma2025latte,
  title={Latte: Latent diffusion transformer for video generation},
  author={Ma, Xin and Wang, Yaohui and Chen, Xinyuan and Jia, Gengyun and Liu, Ziwei and Li, Yuan-Fang and Chen, Cunjian and Qiao, Yu},
  journal={Transactions on Machine Learning Research},
  year={2025}
}

@inproceedings{chang2017matterport3d,
  title={Matterport3D: Learning from RGB-D Data in Indoor Environments},
  author={Chang, Angel and Dai, Angela and Funkhouser, Thomas and Halber, Maciej and Niebner, Matthias and Savva, Manolis and Song, Shuran and Zeng, Andy and Zhang, Yinda},
  booktitle={2017 International Conference on 3D Vision (3DV)},
  pages={667--676},
  year={2017},
  organization={IEEE Computer Society}
}

@article{schuhmann2021laion,
  title={Laion-400m: Open dataset of clip-filtered 400 million image-text pairs},
  author={Schuhmann, Christoph and Vencu, Richard and Beaumont, Romain and Kaczmarczyk, Robert and Mullis, Clayton and Katta, Aarush and Coombes, Theo and Jitsev, Jenia and Komatsuzaki, Aran},
  journal={arXiv preprint arXiv:2111.02114},
  year={2021}
}

@inproceedings{goyal2017something,
  title={The" something something" video database for learning and evaluating visual common sense},
  author={Goyal, Raghav and Ebrahimi Kahou, Samira and Michalski, Vincent and Materzynska, Joanna and Westphal, Susanne and Kim, Heuna and Haenel, Valentin and Fruend, Ingo and Yianilos, Peter and Mueller-Freitag, Moritz and others},
  booktitle={Proceedings of the IEEE international conference on computer vision},
  pages={5842--5850},
  year={2017}
}

@inproceedings{damen2018scaling,
  title={Scaling egocentric vision: The epic-kitchens dataset},
  author={Damen, Dima and Doughty, Hazel and Farinella, Giovanni Maria and Fidler, Sanja and Furnari, Antonino and Kazakos, Evangelos and Moltisanti, Davide and Munro, Jonathan and Perrett, Toby and Price, Will and others},
  booktitle={Proceedings of the European conference on computer vision (ECCV)},
  pages={720--736},
  year={2018}
}

@article{carreira2019short,
  title={A short note on the kinetics-700 human action dataset},
  author={Carreira, Joao and Noland, Eric and Hillier, Chloe and Zisserman, Andrew},
  journal={arXiv preprint arXiv:1907.06987},
  year={2019}
}

@inproceedings{finn2017deep,
  title={Deep visual foresight for planning robotic motion},
  author={Finn, Chelsea and Levine, Sergey},
  booktitle={2017 IEEE international conference on robotics and automation},
  pages={2786--2793},
  year={2017},
  organization={IEEE}
}

@article{ebert2018visual,
  title={Visual foresight: Model-based deep reinforcement learning for vision-based robotic control},
  author={Ebert, Frederik and Finn, Chelsea and Dasari, Sudeep and Xie, Annie and Lee, Alex and Levine, Sergey},
  journal={arXiv preprint arXiv:1812.00568},
  year={2018}
}

@article{zhou2025vision,
  title={Vision-Language-Action Model with Open-World Embodied Reasoning from Pretrained Knowledge},
  author={Zhou, Zhongyi and Zhu, Yichen and Wen, Junjie and Shen, Chaomin and Xu, Yi},
  journal={arXiv preprint arXiv:2505.21906},
  year={2025}
}

@inproceedings{hafner2019learning,
  title={Learning latent dynamics for planning from pixels},
  author={Hafner, Danijar and Lillicrap, Timothy and Fischer, Ian and Villegas, Ruben and Ha, David and Lee, Honglak and Davidson, James},
  booktitle={International conference on machine learning},
  pages={2555--2565},
  year={2019}
}

@inproceedings{gumbsch2023learning,
  title={Learning hierarchical world models with adaptive temporal abstractions from discrete latent dynamics},
  author={Gumbsch, Christian and Sajid, Noor and Martius, Georg and Butz, Martin V},
  booktitle={The Twelfth International Conference on Learning Representations},
  year={2024},
}

@inproceedings{hafnerdream,
  title={Dream to Control: Learning Behaviors by Latent Imagination},
  author={Hafner, Danijar and Lillicrap, Timothy and Ba, Jimmy and Norouzi, Mohammad},
  booktitle={International Conference on Learning Representations},
  year={2019}
}

@article{hafner2025mastering,
  title={Mastering diverse control tasks through world models},
  author={Hafner, Danijar and Pasukonis, Jurgis and Ba, Jimmy and Lillicrap, Timothy},
  journal={Nature},
  pages={1--7},
  year={2025},
  publisher={Nature Publishing Group UK London}
}

@inproceedings{hansen2022temporal,
  title={Temporal Difference Learning for Model Predictive Control},
  author={Hansen, Nicklas A and Su, Hao and Wang, Xiaolong},
  booktitle={International Conference on Machine Learning},
  pages={8387--8406},
  year={2022}

}

@inproceedings{hansen2024td,
  title={TD-MPC2: Scalable, Robust World Models for Continuous Control},
  author={Hansen, Nicklas and Su, Hao and Wang, Xiaolong},
  booktitle={The Twelfth International Conference on Learning Representations},
  year={2024},
}

@article{schrittwieser2020mastering,
  title={Mastering atari, go, chess and shogi by planning with a learned model},
  author={Schrittwieser, Julian and Antonoglou, Ioannis and Hubert, Thomas and Simonyan, Karen and Sifre, Laurent and Schmitt, Simon and Guez, Arthur and Lockhart, Edward and Hassabis, Demis and Graepel, Thore and others},
  journal={Nature},
  volume={588},
  number={7839},
  pages={604--609},
  year={2020},
  publisher={Nature Publishing Group UK London}
}

@article{silver2018general,
  title={A general reinforcement learning algorithm that masters chess, shogi, and Go through self-play},
  author={Silver, David and Hubert, Thomas and Schrittwieser, Julian and Antonoglou, Ioannis and Lai, Matthew and Guez, Arthur and Lanctot, Marc and Sifre, Laurent and Kumaran, Dharshan and Graepel, Thore and others},
  journal={Science},
  volume={362},
  number={6419},
  pages={1140--1144},
  year={2018},
  publisher={American Association for the Advancement of Science}
}

@article{ye2021mastering,
  title={Mastering atari games with limited data},
  author={Ye, Weirui and Liu, Shaohuai and Kurutach, Thanard and Abbeel, Pieter and Gao, Yang},
  journal={Advances in neural information processing systems},
  volume={34},
  pages={25476--25488},
  year={2021}
}

@article{wang2024efficientzero,
  title={Efficientzero v2: Mastering discrete and continuous control with limited data},
  author={Wang, Shengjie and Liu, Shaohuai and Ye, Weirui and You, Jiacheng and Gao, Yang},
  journal={arXiv preprint arXiv:2403.00564},
  year={2024}
}

@article{wu2023pre,
  title={Pre-training contextualized world models with in-the-wild videos for reinforcement learning},
  author={Wu, Jialong and Ma, Haoyu and Deng, Chaoyi and Long, Mingsheng},
  journal={Advances in Neural Information Processing Systems},
  volume={36},
  pages={39719--39743},
  year={2023}
}

@article{long2025survey,
  title={A Survey: Learning Embodied Intelligence from Physical Simulators and World Models},
  author={Long, Xiaoxiao and Zhao, Qingrui and Zhang, Kaiwen and Zhang, Zihao and Wang, Dingrui and Liu, Yumeng and Shu, Zhengjie and Lu, Yi and Wang, Shouzheng and Wei, Xinzhe and others},
  journal={arXiv preprint arXiv:2507.00917},
  year={2025}
}

@article{lin2025exploring,
  title={Exploring the evolution of physics cognition in video generation: A survey},
  author={Lin, Minghui and Wang, Xiang and Wang, Yishan and Wang, Shu and Dai, Fengqi and Ding, Pengxiang and Wang, Cunxiang and Zuo, Zhengrong and Sang, Nong and Huang, Siteng and others},
  journal={arXiv preprint arXiv:2503.21765},
  year={2025}
}

@article{kong20253d,
  title={3D and 4D world modeling: A survey},
  author={Kong, Lingdong and Yang, Wesley and Mei, Jianbiao and Liu, Youquan and Liang, Ao and Zhu, Dekai and Lu, Dongyue and Yin, Wei and Hu, Xiaotao and Jia, Mingkai and others},
  journal={arXiv preprint arXiv:2509.07996},
  year={2025}
}

@article{ai2025review,
  title={A review of learning-based dynamics models for robotic manipulation},
  author={Ai, Bo and Tian, Stephen and Shi, Haochen and Wang, Yixuan and Pfaff, Tobias and Tan, Cheston and Christensen, Henrik I and Su, Hao and Wu, Jiajun and Li, Yunzhu},
  journal={Science Robotics},
  volume={10},
  number={106},
  pages={eadt1497},
  year={2025},
  publisher={American Association for the Advancement of Science}
}

@article{yu2025survey,
  title={A survey of interactive generative video},
  author={Yu, Jiwen and Qin, Yiran and Che, Haoxuan and Liu, Quande and Wang, Xintao and Wan, Pengfei and Zhang, Di and Gai, Kun and Chen, Hao and Liu, Xihui},
  journal={arXiv preprint arXiv:2504.21853},
  year={2025}
}

@article{zhu2024sora,
  title={Is sora a world simulator? a comprehensive survey on general world models and beyond},
  author={Zhu, Zheng and Wang, Xiaofeng and Zhao, Wangbo and Min, Chen and Deng, Nianchen and Dou, Min and Wang, Yuqi and Shi, Botian and Wang, Kai and Zhang, Chi and others},
  journal={arXiv preprint arXiv:2405.03520},
  year={2024}
}

@article{liang2025large,
  title={Large Model Empowered Embodied AI: A Survey on Decision-Making and Embodied Learning},
  author={Liang, Wenlong and Zhou, Rui and Ma, Yang and Zhang, Bing and Li, Songlin and Liao, Yijia and Kuang, Ping},
  journal={arXiv preprint arXiv:2508.10399},
  year={2025}
}

@article{ding2025understanding,
  title={Understanding world or predicting future? a comprehensive survey of world models},
  author={Ding, Jingtao and Zhang, Yunke and Shang, Yu and Zhang, Yuheng and Zong, Zefang and Feng, Jie and Yuan, Yuan and Su, Hongyuan and Li, Nian and Sukiennik, Nicholas and others},
  journal={ACM Computing Surveys},
  volume={58},
  number={3},
  pages={1--38},
  year={2025},
  publisher={ACM New York, NY}
}

@online{nvidia_wm2025,
  title        = {World models},
  url          = {https://www.nvidia.com/en-au/glossary/world-models},
  author       = {NVIDIA},
  year={2025},
}

@article{bardes2024revisiting,
  title={Revisiting feature prediction for learning visual representations from video},
  author={Bardes, Adrien and Garrido, Quentin and Ponce, Jean and Chen, Xinlei and Rabbat, Michael and LeCun, Yann and Assran, Mahmoud and Ballas, Nicolas},
  journal={arXiv preprint arXiv:2404.08471},
  year={2024}
}

@article{assran2025v,
  title={V-jepa 2: Self-supervised video models enable understanding, prediction and planning},
  author={Assran, Mido and Bardes, Adrien and Fan, David and Garrido, Quentin and Howes, Russell and Muckley, Matthew and Rizvi, Ammar and Roberts, Claire and Sinha, Koustuv and Zholus, Artem and others},
  journal={arXiv preprint arXiv:2506.09985},
  year={2025}
}

@inproceedings{pang2025reviwo,
  title={Learning View-invariant World Models for Visual Robotic Manipulation},
  author={Pang, Jing-Cheng and Tang, Nan and Li, Kaiyuan and Tang, Yuting and Cai, Xin-Qiang and Zhang, Zhen-Yu and Niu, Gang and Sugiyama, Masashi and Yu, Yang},
  booktitle={The Thirteenth International Conference on Learning Representations},
  year={2025}
}

@article{martinez2025coral,
  title={In-Context Reinforcement Learning via Communicative World Models},
  author={Martinez-Lopez, Fernando and Li, Tao and Lu, Yingdong and Chen, Juntao},
  journal={arXiv preprint arXiv:2508.06659},
  year={2025}
}

@article{chen2025robohorizon,
  title={Robohorizon: An llm-assisted multi-view world model for long-horizon robotic manipulation},
  author={Chen, Zixuan and Huo, Jing and Chen, Yangtao and Gao, Yang},
  journal={arXiv preprint arXiv:2501.06605},
  year={2025}
}

@article{chen2025vlwm,
  title={Planning with Reasoning using Vision Language World Model},
  author={Chen, Delong and Moutakanni, Theo and Chung, Willy and Bang, Yejin and Ji, Ziwei and Bolourchi, Allen and Fung, Pascale},
  journal={arXiv preprint arXiv:2509.02722},
  year={2025}
}

@article{xiao2025world,
  title={World-Env: Leveraging World Model as a Virtual Environment for VLA Post-Training},
  author={Xiao, Junjin and Yang, Yandan and Chang, Xinyuan and Chen, Ronghan and Xiong, Feng and Xu, Mu and Zheng, Wei-Shi and Zhang, Qing},
  journal={arXiv preprint arXiv:2509.24948},
  year={2025}
}
}
\end{sloppypar}
\end{document}